\documentclass[sn-apa,iicol]{sn-jnl}

\usepackage{graphicx}
\usepackage{amsmath}
\usepackage{amssymb}
\usepackage{booktabs}
\usepackage{comment}
\usepackage{multirow}
\usepackage{mathtools}
\usepackage{epsfig}
\usepackage[misc]{ifsym}

\usepackage{makecell}
\usepackage{pifont}
\usepackage{bbding}
\usepackage{wrapfig}
\usepackage{verbatim}
\usepackage{relsize}
\usepackage{colortbl}
\usepackage{bm}
\usepackage{bbm}
\usepackage{breakurl}
\usepackage{color}
\usepackage{diagbox}
\usepackage[capitalize]{cleveref}
\usepackage{fancyvrb,cprotect}

\usepackage{graphicx}
\usepackage{amsmath}
\usepackage{amssymb}
\usepackage{booktabs}

\usepackage{bigstrut}
\usepackage{multirow}
\usepackage{multicol}
\usepackage{hhline}
\usepackage{lipsum}
\usepackage{caption}
\usepackage{subcaption}
\usepackage{url}

\usepackage{color,soul}
\usepackage{xcolor}

\newcommand{\wh}[1]{\textcolor{black}{#1}}

\DeclareUnicodeCharacter{2212}{-}

\begin{document}


\title{Towards Data-Centric Face Anti-Spoofing: Improving Cross-domain Generalization via Physics-based Data Synthesis}

\author[1]{\fnm{Rizhao} \sur{Cai$\dagger$}\email{rzcai@ntu.edu.sg}}

\author[1]{\fnm{Cecelia} \sur{Soh$\dagger$}\email{cecelia.sohyp@ntu.edu.sg}}

\author[3]{\fnm{Zitong} \sur{Yu}\email{zitong.yu@ieee.org}}

\author[4]{\fnm{Haoliang} \sur{Li}\email{haoliang.li@cityu.edu.hk
}}

\author[2]{\fnm{Wenhan} \sur{Yang\Letter}\email{yangwh@pcl.ac.cn}}

\author[1]{{\fnm{Alex C.} \sur{Kot}}\email{eackot@ntu.edu.sg}}

\affil[1]{\orgdiv{ROSE Lab}, \orgname{Nanyang Technological University}, \orgaddress{\country{Singapore}}}

\affil[2]{\orgdiv{Pengcheng Laboratory}, \orgaddress{\city{Shenzhen}, \country{China}}}

\affil[3]{\orgname{Great Bay University}, \orgaddress{\country{China}}}

\affil[4]{\orgname{City University of Hong Kong},  \orgaddress{\city{Hong Kong}, \country{China}}}

\affil[]{$\dagger$ Equal Contribution  \Letter Corresponding Author}

\abstract{Face Anti-Spoofing (FAS) research is challenged by the cross-domain problem, where there is a domain gap between the training and testing data. While recent FAS works are mainly model-centric, focusing on developing domain generalization algorithms for improving cross-domain performance, data-centric research for face anti-spoofing, improving generalization from data quality and quantity, is largely ignored. Therefore, our work starts with data-centric FAS by conducting a comprehensive investigation from the data perspective for improving cross-domain generalization of FAS models. More specifically, at first, based on physical procedures of capturing and recapturing, we propose task-specific FAS data augmentation (FAS-Aug), which increases data diversity by synthesizing data of artifacts, such as printing noise, color distortion, moiré pattern, \textit{etc}. Our experiments show that using our FAS augmentation can surpass traditional image augmentation in training FAS models to achieve better cross-domain performance. Nevertheless, we observe that models may rely on the augmented artifacts, which are not environment-invariant, and using FAS-Aug may have a negative effect. As such, we propose Spoofing Attack Risk Equalization (SARE) to prevent models from relying on certain types of artifacts and improve the generalization performance. Last but not least, our proposed FAS-Aug and SARE with recent Vision Transformer backbones can achieve state-of-the-art performance on the FAS cross-domain generalization protocols. The implementation is available at \url{https://github.com/RizhaoCai/FAS_Aug}.}

\maketitle

\section{Introduction} \label{sec:intro}
Biometric authentication systems based on Face Recognition (FR) bring great convenience to practical applications, but FR systems are vulnerable to face spoofing attacks.
Attackers could present spoofing faces of printed photos, digital replay, or even 3D masks to the camera to spoof the system. 
Face Anti-Spoofing (FAS) aims to protect FR systems by detecting spoofing attacks.

\par 
\begin{figure}
     \centering
     \small
     \includegraphics[width=0.45\textwidth]{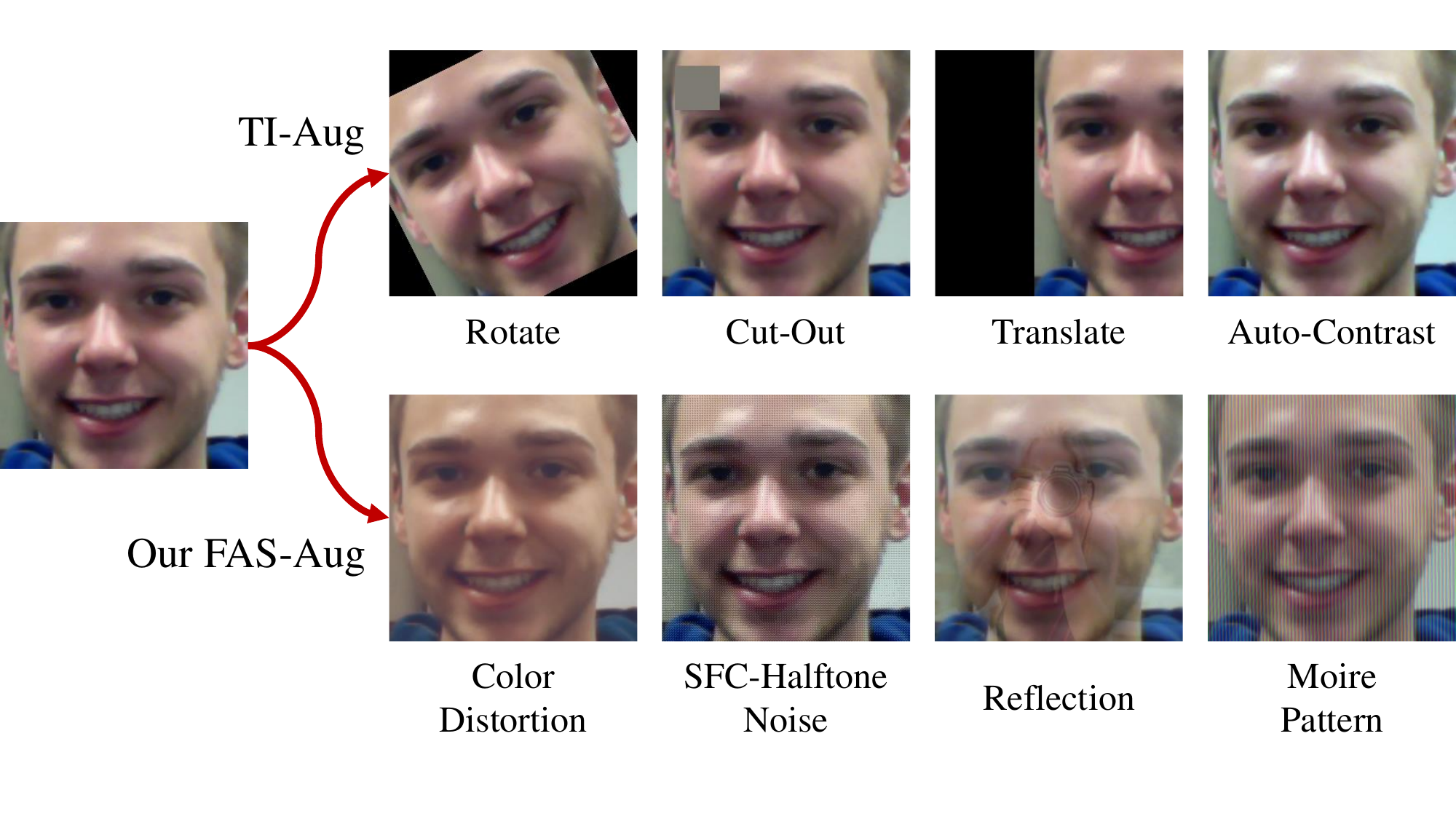}
     \caption{Compare our Traditional Augmentation (TI-Aug) and our proposed Face Anti-Spoofing Augmentation (FAS-Aug). The top row shows the TI-Aug results of `Rotate', `Cut-out', `Translate', and `Auto-Contrast'. 
     TI-Aug mainly \wh{includes} the geometric transformation, which does not provide spoofing-specific diversity. Our proposed FAS-Aug can synthesize face spoofing artifacts (bottom row), such as Color distortion, Printing halftone noise, Reflection, moiré patterns, \textit{etc.}}
     \label{fig:teaser}
\end{figure}
\textcolor{black}{Recent FAS methods are mainly leveraging deep neural networks to learn discriminative features \citep{yu2022deep}. However, these methods are still challenged by domain shift between the source training and the target testing data \citep{FAS-3DCNN-TIFS-2018,yu2022deep}, where data might be captured under different capturing environments by various devices.}
If source data domains for training do not cover the environments (\textit{e.g.} illuminations, cameras, attacks) of the target testing data, \textit{i.e.} there being the domain shift, the model may perform inconsistently from the source training domain to target testing domains, leading to poor cross-domain performance.

\par
\textcolor{black}{To deal with the domain shift problem, recent methods are model-centric, focusing on developing task-aware model architectures \citep{NASFAS-TPAMI-2020} or model learning (optimization) algorithms 
 such as adversarial learning  \citep{MADDG-CVPR-2019, SSDG-CVPR-2020}, and meta-learning \citep{MetaTeacher-TPAMI-2021,metapattern}, to improve cross-domain generalization performance given the fixed training data. 
 On the other hand, the recent success of the large multi-modal language models \citep{cai2023benchlmm}, \textit{e.g.} ChatGPT and GPT4, raises the attention on data-centric research \citep{cai2023benchlmm}. The data-centric research \citep{data-ai} focuses on improving an Artificial Intelligence (AI) system's performance by engineering the data.} \textcolor{black}{The data-centric AI has three general goals: training data development (e.g. data augmentation), inference data development, and data maintenance. Although some previous works also involve contents overlapping the three goals, such as data augmentation \citep{DCCDN_IJCAI_2021}, the contents are usually ad-hoc and auxiliary to the proposed algorithms. A holistic view from the data-centric aspect is ignored and not comprehensively discussed in previous works.}


\textcolor{black}{Since FAS data is not fixed and would evolve \citep{perez2020learning}, data-centric research for the FAS task is necessary. Therefore, we explore the data-centric Face Anti-Spoofing in this work. In the context of FAS, the goals of inference data development include intra-domain evaluation and out-of-distribution evaluation. Nevertheless, intra-domain performance is already saturated and out-of-distribution evaluation benchmarks have been established \citep{MADDG-CVPR-2019}.  In the goal of training, data development, data collection, data labeling, data preparation, and data augmentation are the subjects to study. Given that dozens of labeled datasets have been collected and released \citep{CASIA-SURF, DB-CASIAFASD, OULU_NPU_2017, FAS-Auxiliary-CVPR-2018}, we seek further to improve a FAS system performance via data augmentation.}

Data augmentation is not new and it is an effective way to improve performance. However, augmentation methods specific to the FAS task have not been comprehensively studied. Traditional Image data Augmentation (TI-Aug), such as rotation, cutout, \textit{etc.}, is useful in general computer vision applications \citep{adv_auto}, but TI-Aug is not specially designed for FAS and cannot help increase the data diversity from the perspective of spoofing artifacts. To fill in the gap, we design FAS-Aug, a bag of task-aware augmentation methods for the Face Anti-Spoofing task, which can provide synthesized spoofing artifacts based on the simulations of the physical capturing and recapturing processes. In detail, the FAS-Aug can simulate the behavior of printed photo and replay video attacks and synthesize diverse artifacts such as printing noises, color distortion, moiré patterns, reflection patterns, \textit{etc}. Fig.~\ref{fig:teaser} compares our FAS-Aug and TI-Aug, and experimental results in Section 4 validate the effectiveness of our FAS-Aug.

\textcolor{black}{In terms of the third goal of data-centric research, data maintenance refers to the process of maintaining the quality and reliability of data \citep{data-ai}. We also explore along with this direction and obtain an interesting observation that there are more spoofing face examples than real-face examples in public FAS datasets \citep{DB-CASIAFASD, LBP-FAS-BIOSIG-2012, FAS-IDA-TIFS-2015, OULU_NPU_2017,FAS-UnsupervisedDA-TIFS-2018, FAS-Auxiliary-CVPR-2018}.
However, in real applications, the obtained real-face examples could be more than the spoofing ones, as real-face images or videos can be easily downloaded from the internet but the manufacturing and collection of spoofing attack examples take extra costs. Furthermore, real-face examples could be collected during the use of a face recognition system \citep{li2022one}. Thus, an interesting problem is raised whether the increased number of real-face examples in the training stage is reliable in bringing in the performance leap.}
However, the above situation is seldom explored by previous works.
To explore this situation, we include extra real-face examples from ROSE-YOUTU dataset \citep{FAS-UnsupervisedDA-TIFS-2018} and SiW dataset \citep{FAS-Auxiliary-CVPR-2018} in the experiments of the MICO protocol \citep{MADDG-CVPR-2019,metapattern}. 
We find by empirical experiments that directly including more real-face examples in training does not necessarily improve cross-domain performance.   
Nevertheless, using our FAS-Aug brings better performance than traditional augmentation in this situation. \textcolor{black}{The study above reveals a data maintenance problem in Face Anti-Spoofing that simply increasing the amount of training data is not reliable in improving performance, and our FAS-Aug can benefit the data maintenance and provide a more reliable solution when using new collected real face data}.

Moreover, we are aware that FAS-Aug is a double-sided sword. 
\textcolor{black}{Despite the diverse spoofing artifacts brought by our proposed FAS-Aug, our generated spoofing artifacts, like many real-world artifacts are non-invariant. For example, a replay attack could contain spoofing artifacts of moiré patterns \citep{moire-analysis-TIFS-2015}, and moiré patterns rarely appear on printed photos, which usually have artifacts of printing noise. The moiré patterns and printing noise only appear on specific types of attack samples and thus non-invariant.
If a model overfits the non-invariant moiré patterns for anti-spoofing classification, the model could not be well generalized to print attacks.
}%
To enjoy the FAS-Aug without worrying about non-invariant artifacts, we further propose Spoofing Attack Risk Equalization (SARE) to balance the risks of different types of spoofing attacks to prevent models from overfitting to certain types of artifacts. 
Our SARE can be implemented by minimizing the domain-level empirical risks of spoofing attacks. 
\begin{figure}
    \centering
    \includegraphics[width=\linewidth]{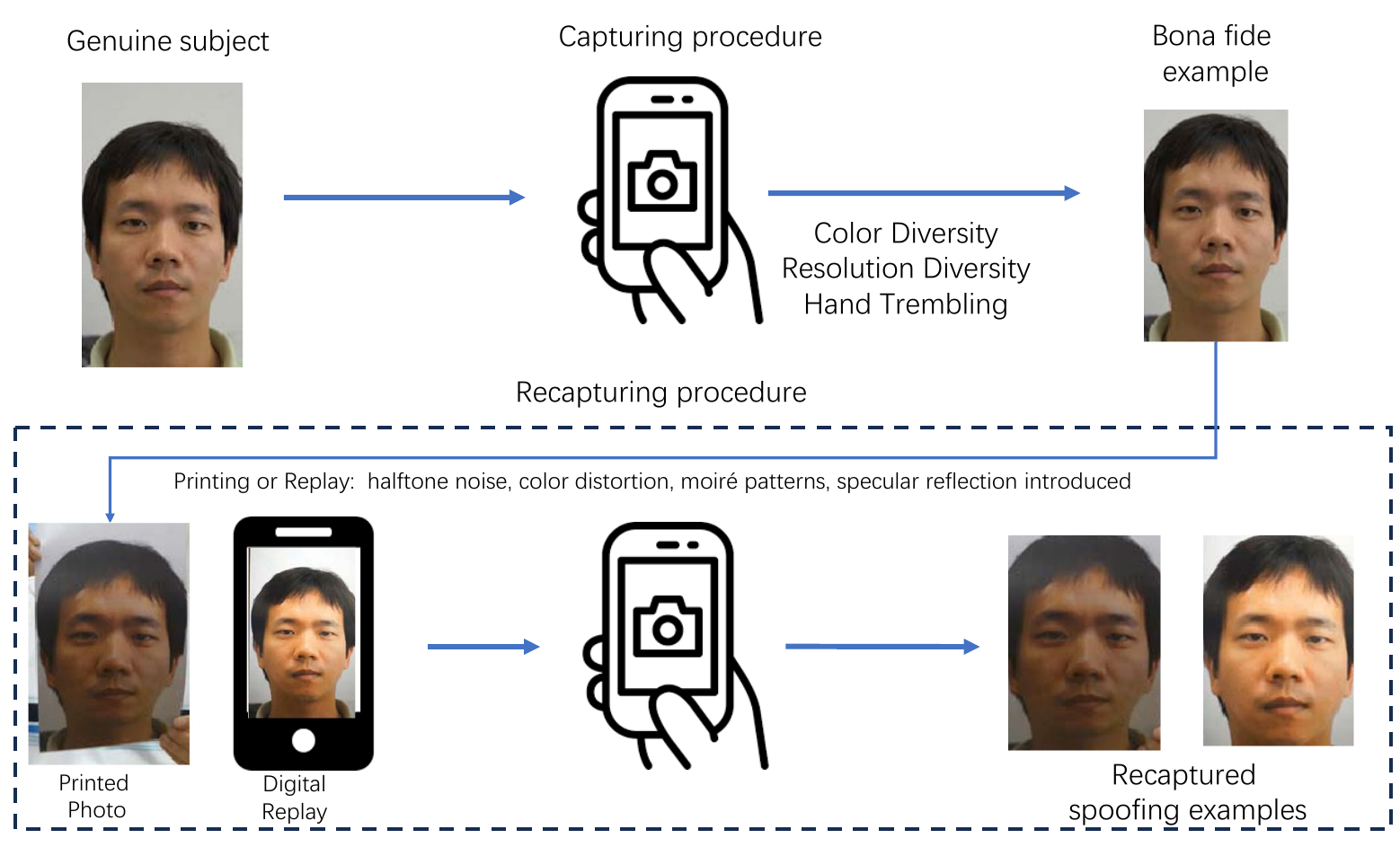}
    \caption{Illustrations of the capturing procedure and the recapturing procedure. The collection of bona fide examples goes through only the capturing process. While recaptured spoofing examples usually go through both the capturing and recapturing procedures. In the recapturing procedure, artifacts, such as Halftone, Color Distortion, \textit{etc.} are introduced into the collected images. }
    \label{fig:procedures}
\end{figure}

\par

\textcolor{black}{We highlight that data-centric research for Face Anti-Spoofing does not diminish the value of model-centric research.  Our work supplements the missing part of previous FAS research from the data aspect.} We summarize our contributions to this work as follows:
\begin{itemize}
     \item We conduct pioneering data-centric research. In terms of data training development, we propose FAS-Aug to increase FAS-specific data diversity based on \wh{the} physical procedure of capturing and recapturing for data \wh{synthesis}. It does not require a neural network for data synthesis generation. As such, our FAS-Aug is implemented as a \textit{plug-and-play} format as \textit{torchvision} image transform, which can be used with other methods and benefit the entire FAS community. 
     \textcolor{black}{The source code is available on GitHub}\footnote{\url{https://github.com/RizhaoCai/FAS_Aug}}.  \\
     \item In terms of data maintenance, we raise the ignored case in most previous works where there are more real-face examples than spoofing ones. Our proposed FAS-Aug can augment more spoofing examples to collaborate with more real face samples and bring better cross-domain generalization performance.
    \item We propose Spoofing Attack Risk Equalization (SARE) to learn more generalized features by preventing the model from relying on non-invariant artifacts of the augmented data.\\
    \item Our proposed FAS-Aug and SARE can help the vision transformer network achieve state-of-the-art performance on leave-one-out cross-domain generalization protocols.
\end{itemize}

\section{Related works}
\textbf{Face anti-spoofing}
Printed photo attack or video attack examples usually \wh{undergo} multiple capturing and recapturing processes, during which the artifacts may appear, such as blurring \citep{TIFS-2019-MotionBlur}, moiré patterns \citep{moire-analysis-TIFS-2015}, image quality distortion \citep{LI-QUALITY, IQA-ICPR-2014}, printing noise \citep{IQA-ICPR-2014}, color distortion \citep{FAS-ColorTexture-TIFS-2016}, \textit{etc}. Traditional FAS methods are mainly based on handcraft features for analysis \citep{LBP-TOP-EJIVP-2014, MotionLBP-ICB-2013,deep-forest-lbp,FAS-ColorTexture-TIFS-2016}.
 Recent FAS methods extract representative features based on deep learning, such as reinforcement learning \citep{CAI-2020-DRL}, pixel-wise supervision \citep{FAS-Auxiliary-CVPR-2018, Ternary-TIFS-2018, yu2021revisiting}, and central difference convolution \citep{CDCN-CVPR-2020}, which have achieved saturated performance in the intra-domain testing scenarios. More recent FAS works are mainly focusing on the cross-domain (cross-dataset) scenario, where the training and testing data are drawn from different distributions \citep{FAS-3DCNN-TIFS-2018}. To learn domain-invariant features, more advanced techniques are utilized, such as meta-learning \citep{metapattern, RFMetaFAS-AAAI-2020, NASFAS-TPAMI-2020,MetaTeacher-TPAMI-2021}, adversarial learning \citep{MADDG-CVPR-2019, SSDG-CVPR-2020}, disentanglement learning \citep{wang2020cross,wu2021dual,9779478}, \textit{etc.} Face Anti-Spoofing algorithms in other scenarios, such as domain adaptation \citep{liu2022source, fas-adapter-eccv2022, wang2021self, FAS-UnsupervisedDA-TIFS-2018, wang2020unsupervised,s-adapter}, continual learning~\citep{Cai_2023_ICCV}, multi-modal learning~\citep{Lin_2024_CVPR}, multi-task learning~\citep{yu2024benchmarking}, have also been studied. By contrast, the data side for FAS is relatively less explored.
%

\textbf{Data augmentation}
Data augmentation is a widespread strategy for various computer vision tasks\citep{chen2023distortion,muller2021trivialaugment,AMCMM2022}, which \wh{increases} the diversity and quantity of training data to improve model performance by geometric transformations, such as flipping, shifting color space, cropping, translation, rotation and so on~\citep{shorten2019survey}.
These operations were also widely used in contrastive learning \citep{SupCon}, self-supervised learning \citep{he2020momentum}, and Auto Augmentation Strategy \citep{cubuk2019autoaugment,li2020differentiable,cubuk2020randaugment,lingchen2020uniformaugment,muller2021trivialaugment}, but the traditional augmentations are not specifically designed for FAS.
While \citep{DCCDN_IJCAI_2021,zhang2021structure} have explored patch-based augmentation for FAS, spoofing artifacts are not synthesized. Wang \textit{et al.} \citep{FAS-CVPR2019-STASN} propose data synthesis for the FAS problem. However, Wang \textit{et al.} \citep{FAS-CVPR2019-STASN} merely design reflection artifacts synthesis. Our work even \wh{considers} more diverse artifacts, such as printing noise, moiré patterns, \wh{and} color distortion. Different from Wang \textit{et al.} \citep{FAS-CVPR2019-STASN} \wh{where} their data synthesis is based on extra collected data from \wh{the} internet and the cross-dataset testing is limited, we apply our FAS-Aug on the source training data only, without extra collected data. Moreover, we validate the \wh{effectiveness} of our proposed FAS-Aug extensively in the multi-domain generalization protocols.
\textcolor{black}{Another work that is closely related to our work is \citep{TIFS-negative}, which synthesized fake face examples as negative data via augmentation. There are two key differences between our works and \citep{TIFS-negative}. 1) First, the negative data augmentation can only synthesize fake faces, while our FAS-Aug can augment real and fake face examples. 2) The usage of negative data in \citep{TIFS-negative} relies on the proposed NDA-based generalization method, while our proposed FAS-Aug is versatile as it can collaborate with other state-of-the-art methods.
Moreover, the negative data augmentation utilizes color jitter and color mask, which is included in the TI-Aug that we used for comparison. }

\begin{figure*}
     \centering
     \begin{subfigure}[b]{0.35\textwidth}
         \centering
         \includegraphics[width=\textwidth]{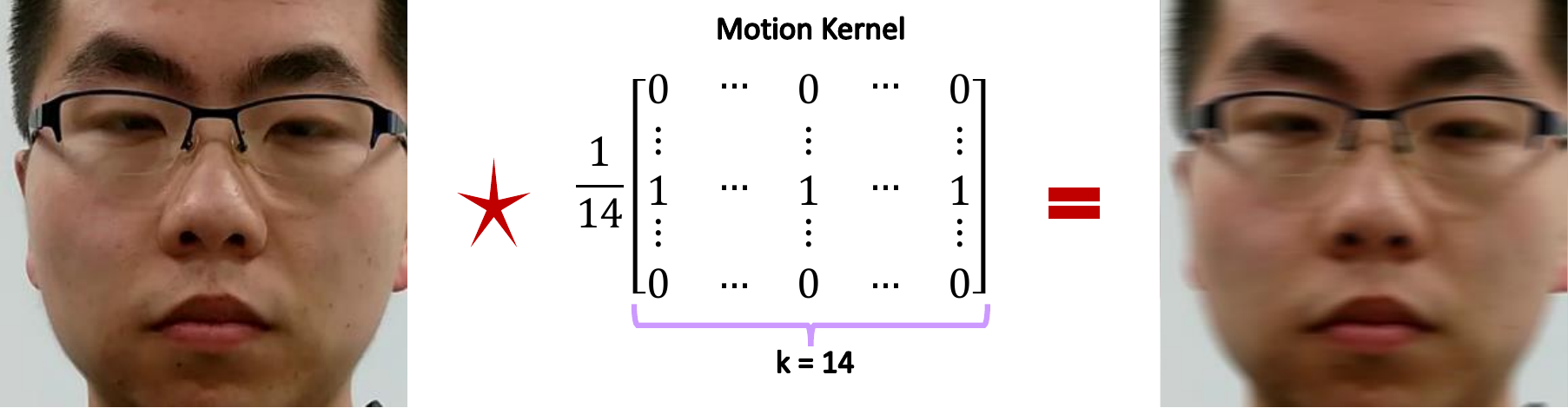}
         \subcaption{Hand trembling simulation}
         \label{fig:motion blur}
     \end{subfigure}
     \hspace{1cm}     
     \begin{subfigure}[b]{0.35\textwidth}
         \centering
         \includegraphics[width=\textwidth]{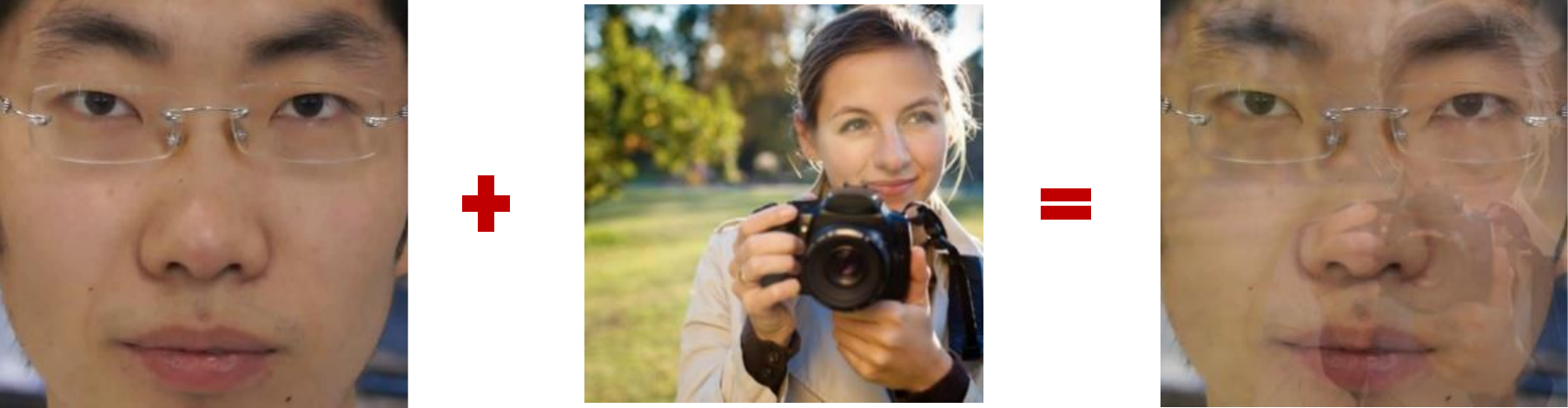}
         \subcaption{Specular reflection artifacts}
         \label{fig:reflection}
     \end{subfigure}
     \hfill
     \begin{subfigure}[b]{0.35\textwidth}
         \centering
         \includegraphics[width=\textwidth]{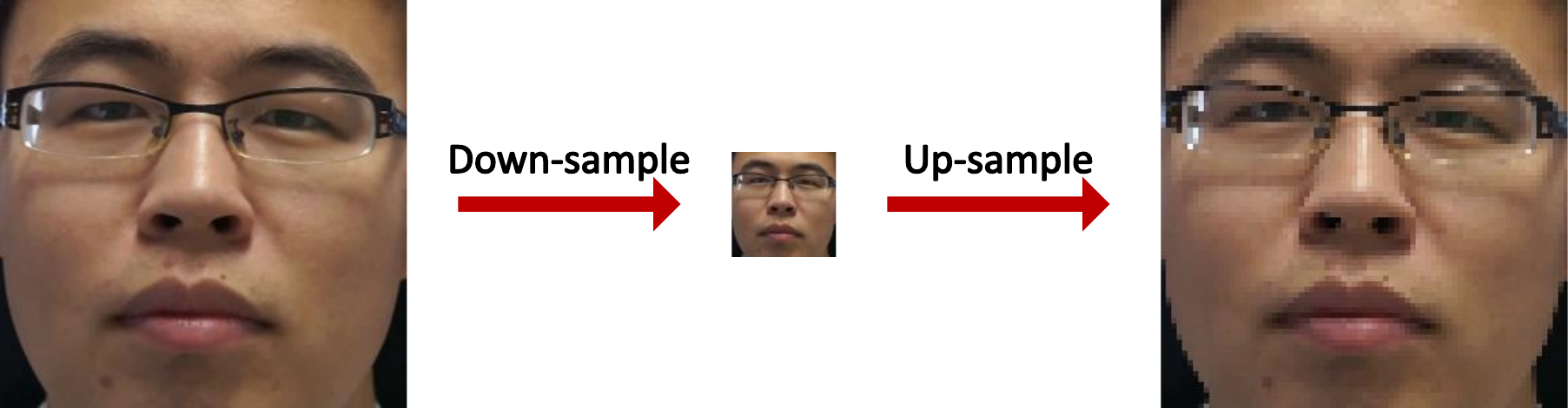}
         \subcaption{Low-resolution simulation}
         \label{fig:quality lose}
     \end{subfigure}      
     \hspace{1cm}
     \begin{subfigure}[b]{0.35\textwidth}
         \centering         
         \includegraphics[width=\textwidth]{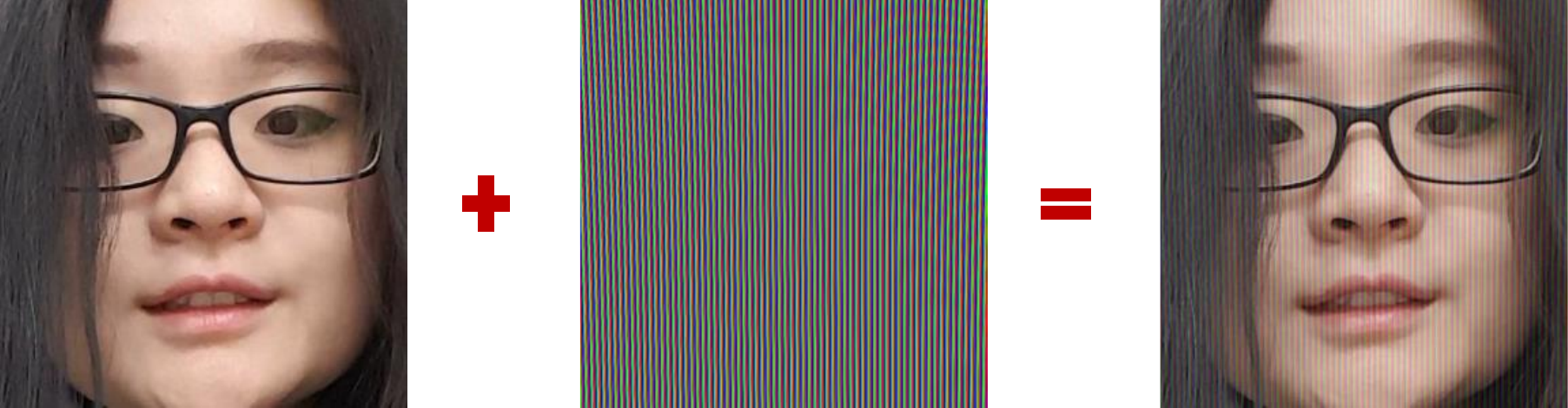}
         \subcaption{moiré pattern artifacts}
         \label{fig:moiré pattern}
     \end{subfigure}      
     \hfill
     \begin{subfigure}[b]{0.35\textwidth}
         \centering
         \includegraphics[width=\textwidth]{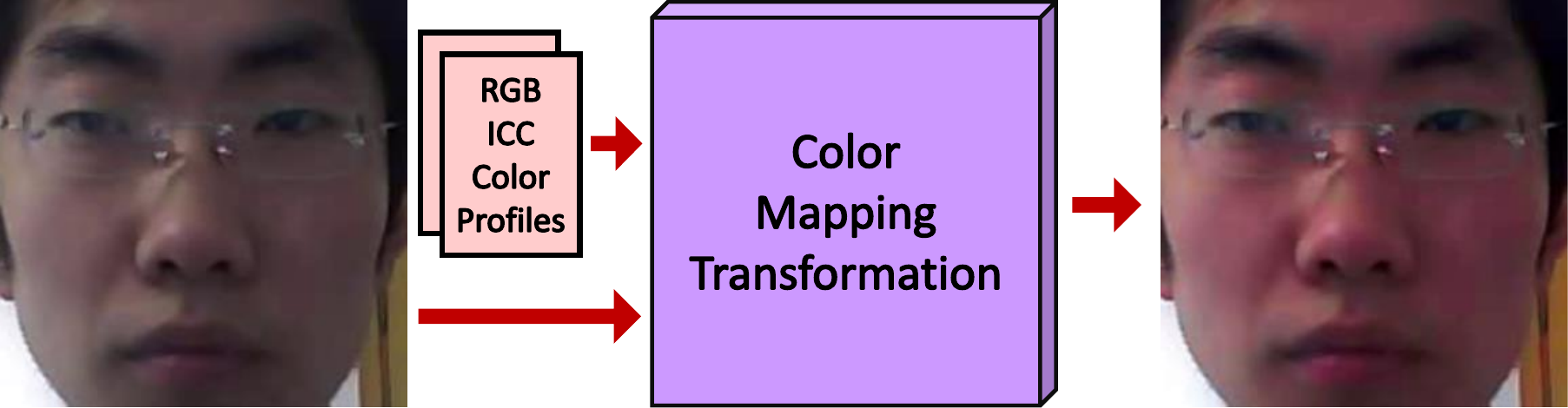}
         \subcaption{Color diversity simulation}
         \label{fig:rgb}
     \end{subfigure}
     \hspace{1cm}
     \begin{subfigure}[b]{0.35\textwidth}
         \centering     
         \includegraphics[width=\textwidth]{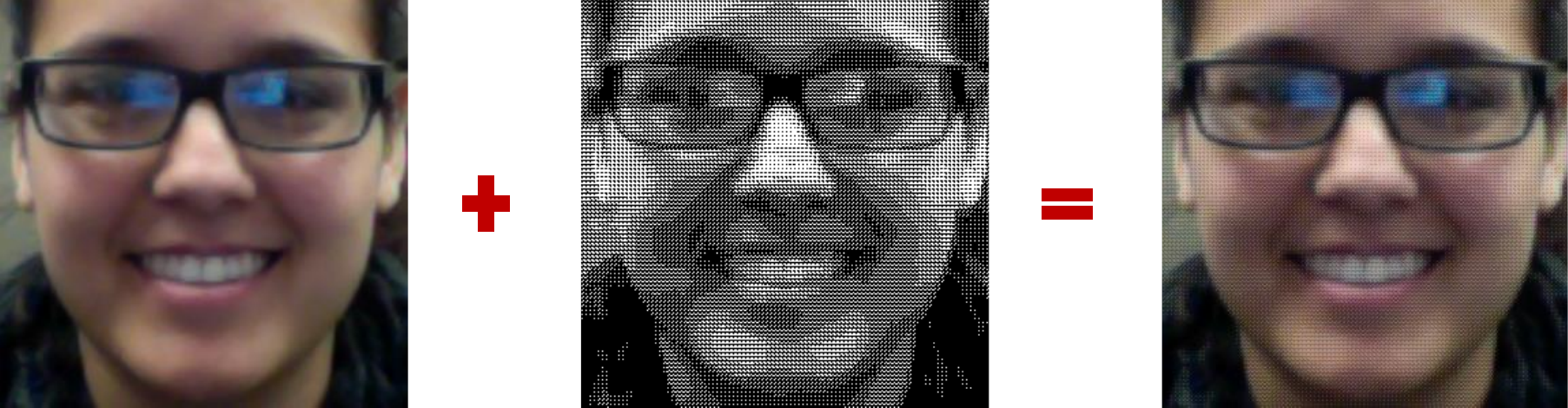}
         \subcaption{SFC-Halftone artifacts}
         \label{fig:halftone}
     \end{subfigure}      
     \hfill
     \begin{subfigure}[b]{0.35\textwidth}
         \centering      
         \includegraphics[width=\textwidth]{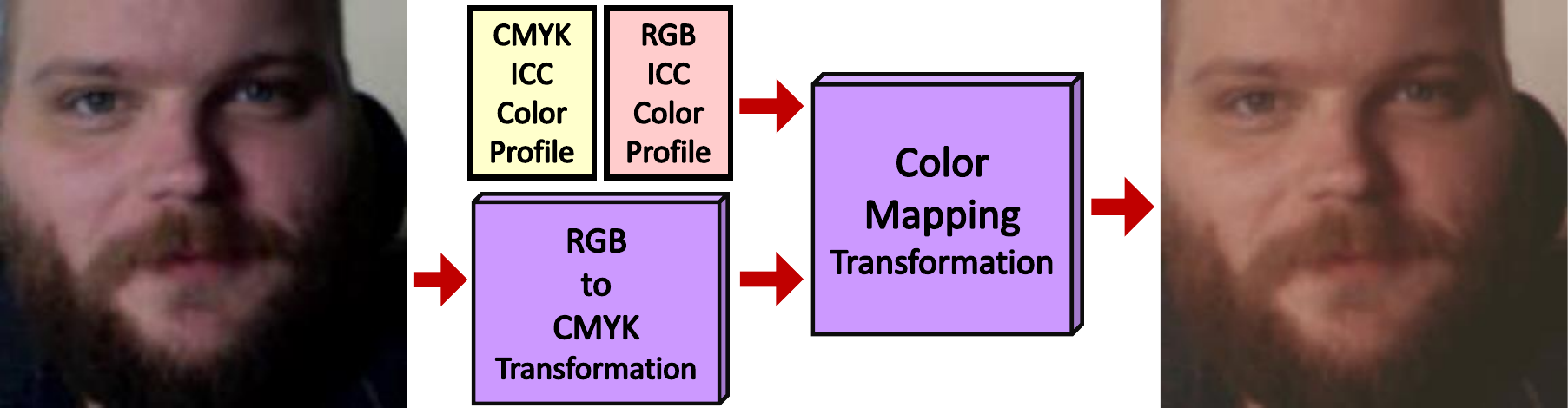}
         \subcaption{Color distortion simulation}
         \label{fig:cmyk}  
     \end{subfigure}   
     \hspace{1cm}
     \begin{subfigure}[b]{0.35\textwidth}
         \centering
         \includegraphics[width=\textwidth]{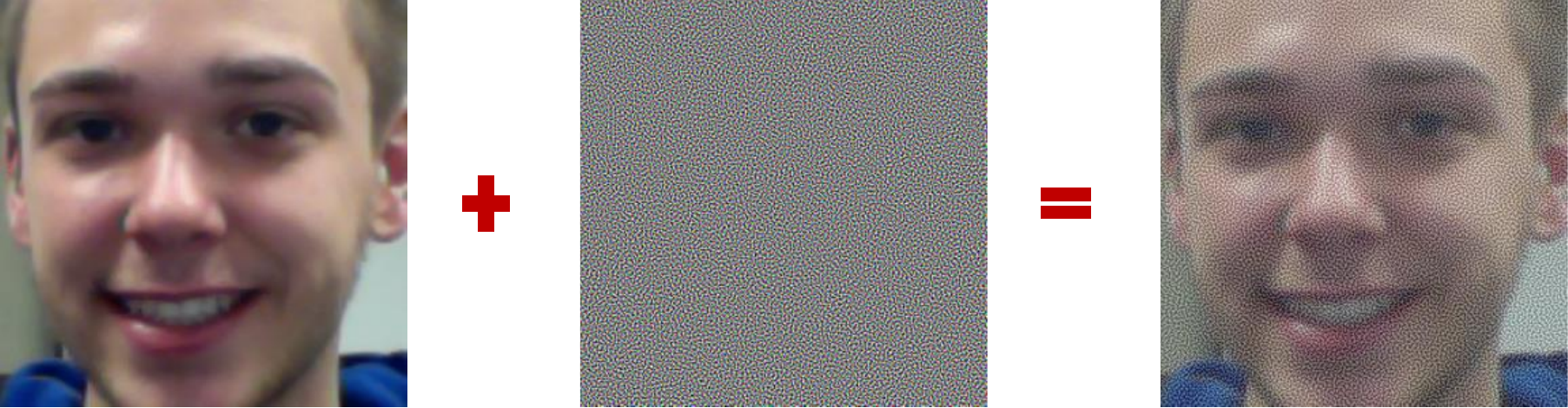}
         \subcaption{BN-Halftone artifacts}
         \label{fig:blue noise}
     \end{subfigure}      
    \caption{Illustrations of the data synthesis in our FAS-Aug. (a), (c) and (e) are examples of general capturing procedure simulation. (b) and (d) are examples of replay attack simulation. (f), (g) and (h) are examples of printed photo simulations. (a) is created by doing convolution with a kernel of 14$\times$14 size, while (c) is formed by down-sampling and followed by up-sampling. The color mapping transformations in (e) and (g) are done with the aid of two ICC color profiles\citep{international2004role}. (b), (d), (f) and (h) are synthesized according to Eq~\ref{eq:reflection}.}
    \label{fig:FAS_aug sample}      
\end{figure*}
\section{Methodology}
In this section, we first describe our Face Anti-Spoofing Augmentation (FAS-Aug), which is based on physics-based simulation.
Then, to get rid of the dependence on non-invariant spoofing artifacts, we further illustrate the optimization method based on our proposed SARE: Spoofing Attack Risk Equalization.

\subsection{Face Anti-Spoofing Augmentation}
Our proposed Face Anti-Spoofing Augmentation is achieved by synthesizing data by simulating the physical procedures of capturing and recapturing the face data, \textcolor{black}{which are illustrated in Fig.~\ref{fig:procedures}.}
\textcolor{black}{As shown in Fig.~\ref{fig:procedures}, the collection of bona fide (real face) examples goes through only the capturing process once. While recaptured spoofing examples usually go through both the capturing and recapturing procedures once or multiple times.
}
\textcolor{black}{Therefore, we are motivated to simulate the physical capturing procedure, during which the camera difference and hand movement create imaging differences of color, quality, and resolution.}

\textcolor{black}{Also, we simulate the recapturing procedure, which includes manufacturing and recapturing the printed photo or digital replay examples. During the recapturing procedure, potential spoofing artifacts are introduced into the synthesized images, such as half-tone noises, color distortion, specular reflection, and moiré pattern. Given the introduced spoofing artifacts, the synthesized images by recapturing simulation are annotated as \textit{Spoof} used in our model training.}

\begin{figure*}[tbp]
     \centering
     \includegraphics[width=0.9\textwidth]{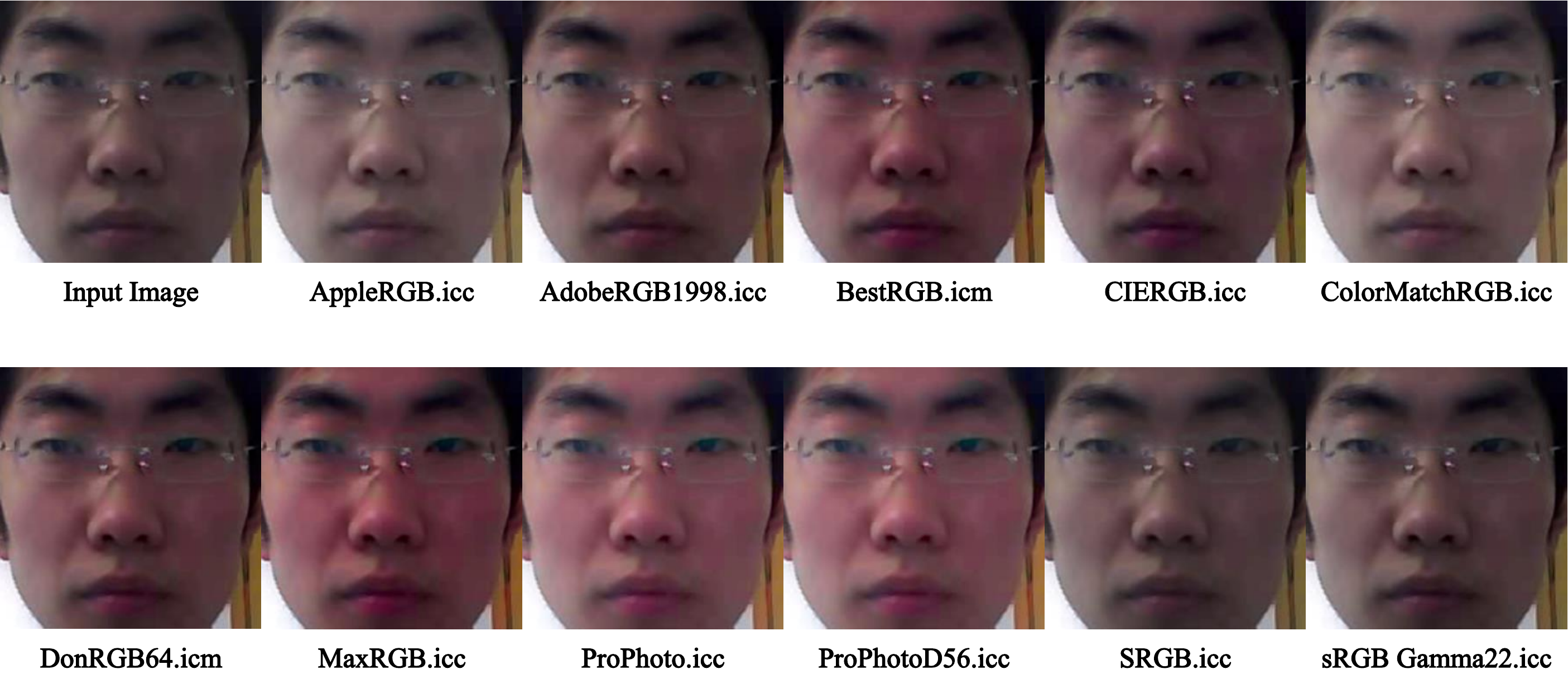}
     \caption{
     The examples of a face image applied the camera color diversity simulation augmentation with different input RGB color profiles but using a constant output profile `sRGB.icc'.
     }
     \label{fig:rgb_examples}
\end{figure*}

\subsubsection{General Capturing Procedure Simulation} \textcolor{black}{As shown in the top of Fig.~\ref{fig:procedures}, the holding hand may move and lead to trembling when capturing images. Also, different cameras would produce images with different color mapping and resolutions.  }
Therefore, the general capturing simulation can further be divided into camera diversity simulation and hand trembling simulation.

\noindent\textbf{Camera diversity simulation} Due to the varying camera modal and settings, using different cameras to capture the same view will create different colors or resolution results among cameras. 
In general, the color of the captured image is controlled by the color correction settings, such as contrast, saturation, \wh{Gamma}, gain, \textit{etc.}\citep{troscianko2015image}, whereas the resolution is mainly related to the pixel pitch of cameras \citep{xiao2009mobile}. \textcolor{black}{Therefore, we propose two augmentation operations to simulate the color diversity and resolution diversity.
}%

When simulating the color diversity,
since that lookup table for color correction is unavailable, to simulate the color diversity of cameras, we conduct color gamut mapping on images based on ICC transformation \citep{morris2005colour,green2013gamut} with open-source ICC color profiles. The color gamut mapping is based on the observation that when an image is displayed on different devices, the color would be reproduced differently.
Given an input image $\mathcal{I}$, the ICC transformation $\mathcal{T}$ can map $\mathcal{I}$ from one ICC color profile to another, and the transformation can be expressed as:
\begin{equation} \label{eq:icc}
\hat{\mathcal{I}} = \mathcal{T}(\mathcal{I}, P_i, P_o),
\end{equation}
where $\mathcal{I}$ is the face image, $\hat{\mathcal{I}}$ is the output transformed image, and $P_i$ and $P_o$ are the input and output ICC color profiles respectively. The visual expression of the process is shown in \textcolor{black}{Fig.}~\ref{fig:rgb}.
In our experiments, $P_i$ and $P_o$ are uniformly sampled from 11 open-source RGB ICC color profiles, which we collected from \citep{profile_a,profiles_b,profiles_c}.
The ICC profile transformation can be easily implemented by Python. As illustrated in Algorithm \ref{alg:code_rgb}, we use the function, \textit{profileToProfile()}, from the PIL.ImageCms module, to perform the ICC transformation $\mathcal{T}$. The input of the function consists of a face image, \textit{img}, the path of an input profile, $p_{src}$, and the path of an output profile, $p_{dst}$.
Fig.~\ref{fig:rgb_examples} shows examples of a face image applied $\mathcal{T}$ with the 11 different input profiles, and a constant output profile `sRGB.icc'. 
\begin{algorithm}
\caption{Pseudo code of conducting the camera color diversity simulation using Pillow API.}
\label{alg:code_rgb}
\small
/* $p_{img}$: path of the input image*/\par
/* $p_{src}$: path of an input ICC color profile*/\par
/* $p_{dst}$: path of an output ICC color profile*/\par
img = PIL.Image.open($p_{img}$)\par
res = PIL.ImageCms.profileToProfile(img, $p_{src}$, $p_{dst}$)\par
\end{algorithm}

Furthermore, we simulate the imaging with low-resolution cameras for resolution diversity, \textcolor{black}{and high-resolution is not achievable as the up-sampling is an ill-posed problem.}
To simulate the low resolution, we randomly down-sample the image to $s$ times the input size, where $s\in [\frac{1}{6},1]$ and $s$ is also the interpolation factor to control the magnitude of this augmentation operation. 
After that, we up-sample the image to the initial size using the nearest neighbor interpolation. 
Fig.~\ref{fig:quality lose} shows the low camera resolution simulation procedure with $s=\frac{1}{3}$.

\begin{figure}[tbp]
     \centering
     \includegraphics[width=0.3\textwidth]{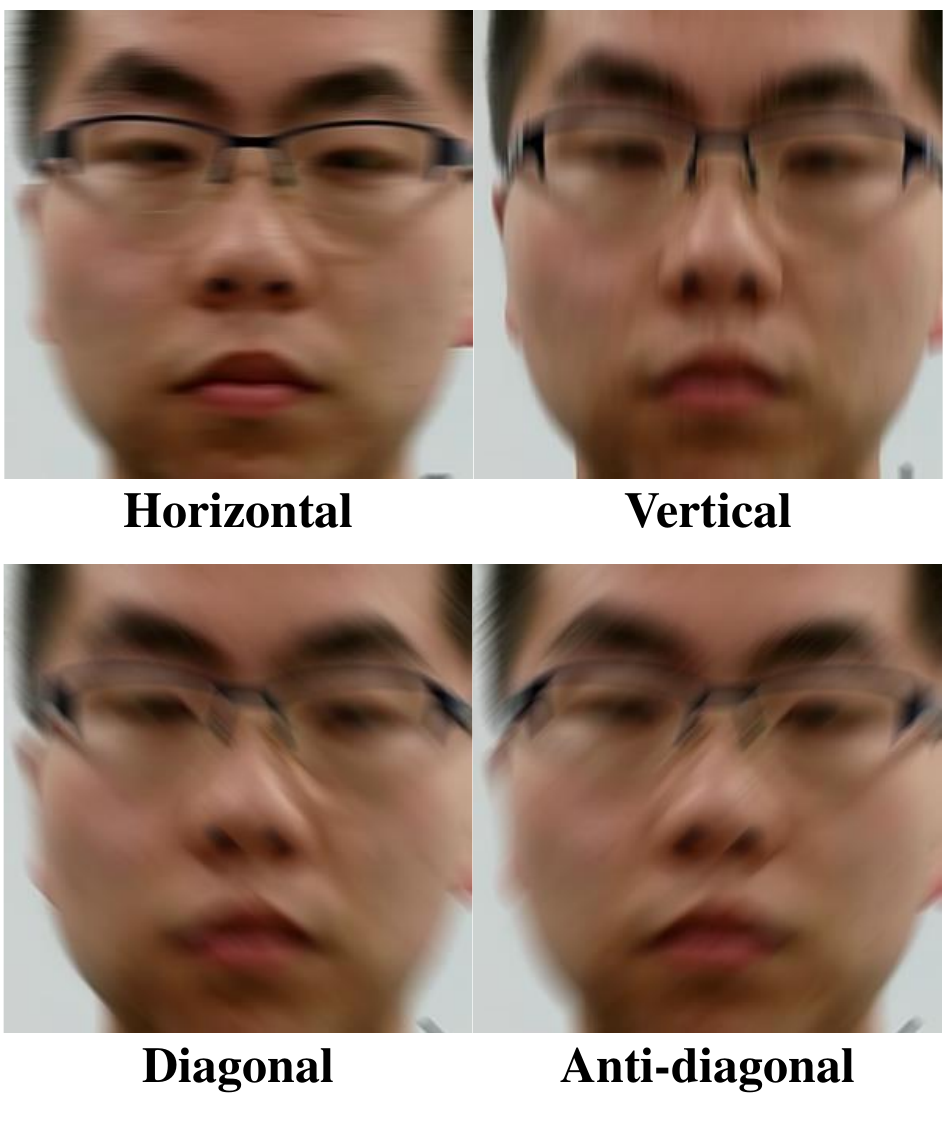}
     \caption{
     The examples of a face image applied the different hand trembling direction simulation augmentation.
     }
     \label{fig:mb_direction}
\end{figure}
\begin{figure*}[htbp]
     \centering
     \includegraphics[width=\textwidth]{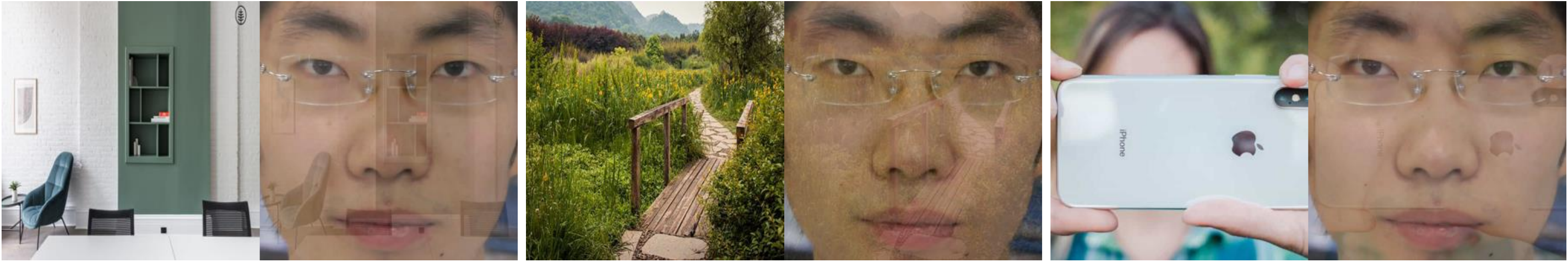}
     \caption{
     The examples of a face image applied the specular reflection artifact simulation augmentation with different kinds of background images.
     }
     \label{fig:reflection_exp}
\end{figure*}

\noindent\textbf{Hand trembling simulation} As shown in Fig.~\ref{fig:procedures}, in the process of hand-held capturing, holding cameras with unsteady hands could lead to motion blur effects on real or spoofing face images.
As such, our FAS-Aug devises an augmentation operation to simulate the motion blur effect.
Following \citep{joshi2015opencv}, given an original image $\mathcal{I}$ and a motion kernel $K$, the blurred image $\hat{\mathcal{I}}$ is synthesized as: 
\begin{equation} 
\hat{\mathcal{I}} = \frac{ 1 }{k} K * \mathcal{I},
\end{equation}
where $*$ is the convolution operator, $K \in\mathbb{R}^{k\times k}$.
The kernel size $k$ proportionally controls the movement magnitude, and the direction of the blurry effect is defined as follows.
For the horizontal blurry effect, $K_{i,j} = 1$, $K_{m\not=i,n\not=j} = 0$, $i\in [1,k]$ and $j=\lfloor\frac{ k+1 }{2}\rfloor$. For the vertical blurry effect, $K_{i,j} = 1$, $K_{m\not=i,n\not=j} = 0$, $i=\lfloor\frac{ k+1 }{2}\rfloor$ and $j\in [1,k]$. For the diagonal blurry effect, $K_{i,j} = 1$ and $K_{m\not=i,n\not=j} = 0$, $i\in [1,k]$ and $j=i$. For the anti-diagonal blurry effect, $K_{i,j} = 1$, $K_{m\not=i,n\not=j} = 0$, $i\in [1,k]$ and $j=k+1-i$. Fig.~\ref{fig:mb_direction} illustrates the results obtained from applying various directions of blurry effects to a facial image.
In our experiments, we randomly use $k \in [3,16]$, and Fig.~\ref{fig:motion blur} depicts an example of \wh{a} face image \wh{applying} the horizontal blurry effect with $k = 14$.

\subsubsection{Replay Attack Recapture Simulation}
The replay attack is conducted by presenting the video replayed on a digital display to a camera. 
During the attack, some spoofing artifacts are possibly recorded by the camera from the digital display, such as the specular reflections and the moiré patterns. 
Therefore, we design two augmentation operations by simulating the reflection and moiré pattern artifacts. \textcolor{black}{Given the introduced artifacts, face images processed by these two operations are annotated as \textit{Spoof}.}

\noindent\textbf{Specular simulation} \textcolor{black}{We first highlight that the reflection artifact in this context does not involve the skin reflection, which is reflected from human skin and is not about spoofing. Instead, we consider merely the specular reflection usually captured from a screen of a digital display. } 
When a photo is taken from a highly reflective screen, the background reflected by the screen can be captured. Such artifacts are known as specular reflection~\citep{eck2021introduction}. 
As such, the specular reflection is also possibly included in the replay attack samples~\citep{Pinto}. 
Following the reflection removal research~\citep{wan2017benchmarking_iccv, wan2017benchmarking_pami}, we \wh{synthesize} reflection artifacts by a convex combination of two images:
\begin{equation} \label{eq:reflection}
\hat{\mathcal{I}} = (1.0 - \gamma)\cdot \mathcal{I} + \gamma \cdot\mathcal{B} ,
\end{equation}
where $\gamma \in [0.03,0.2]$ is the scaling factor to control the magnitude of the reflection, $\mathcal{I}$ is the input face image, \wh{and} $\mathcal{B}$ is a background image. Fig.~\ref{fig:reflection} shows the reflection artifacts simulation on a face image with a background image. We collected 90 background images from the internet, including indoor and outdoor scenes and pictures of a person carrying a capture device. During training, an image $\mathcal{B}$ is randomly sampled from the 90 background images and randomly cropped to the size of the input face image $\mathcal{I}$. Some examples of a face input image fusing with different $\mathcal{B}$ are also shown in Fig.~\ref{fig:reflection_exp}. All the used background images can be found in the supplementary material.

\begin{figure}[htbp]
     \centering
     \small
     \includegraphics[width=0.5\textwidth]{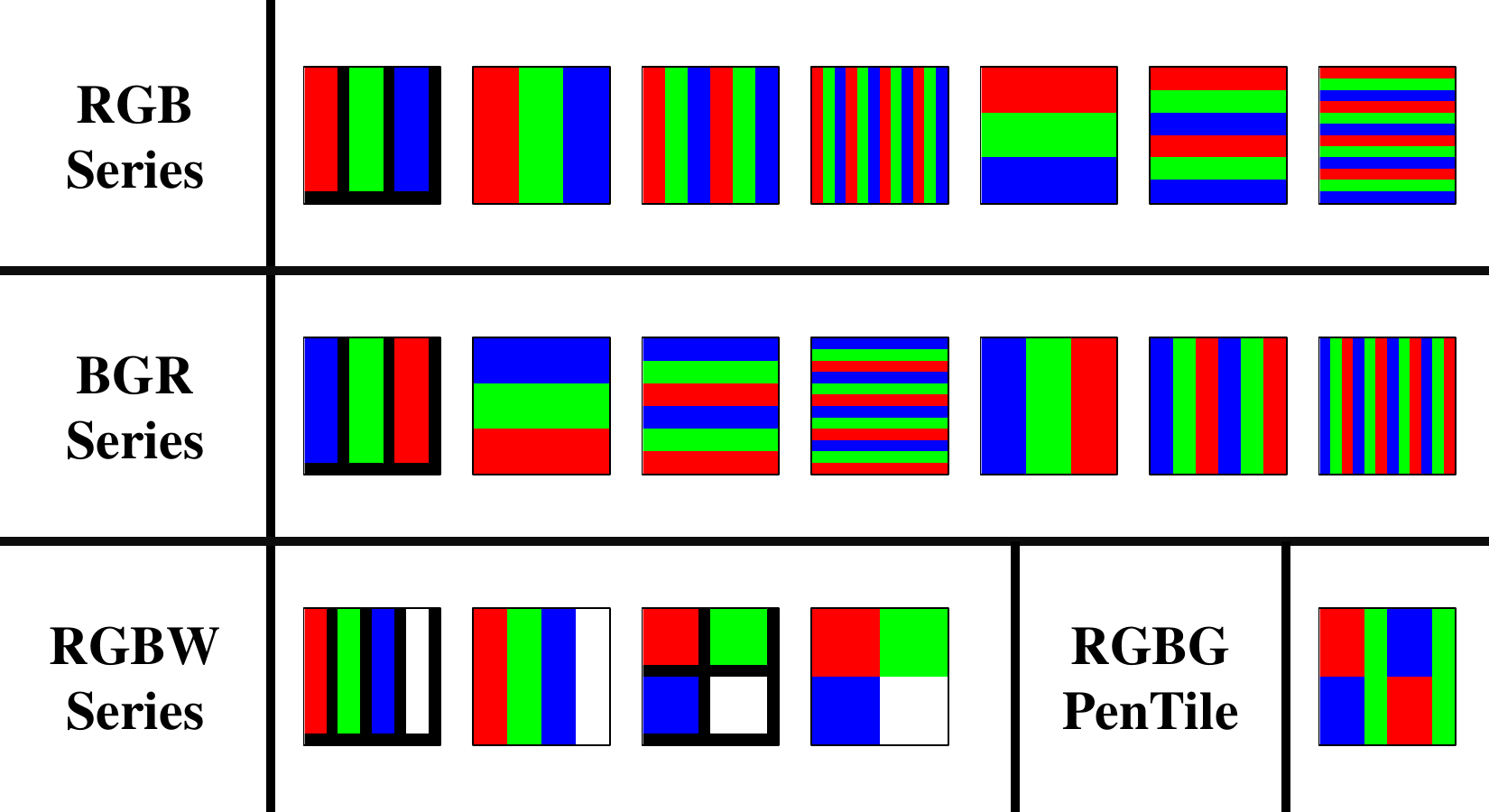}
     \caption{All subpixel layouts used for moiré pattern texture synthesis.}
     \label{fig:subpixels}
\end{figure}

\textbf{Moiré pattern artifacts simulation.} In addition, the moiré pattern is an artifact that could be captured from a digital display in replay attack samples~\citep{moire-analysis-TIFS-2015}, \wh{resulting} from the misaligned pixel or subpixel grid between the digital display and the camera \citep{garcia2015face,yuan2019aim}.
An electronic visual display is made up of tiny pixels. Each pixel has smaller subpixels that aid in displaying an image\citep{fang2013subpixel,spindler2006system,xiong2009performance,elliott200213,chae2017subpixel}. For example, RGB/BGR stripes are the traditional display subpixel layout, but, there are others subpixel layouts used in nowadays display due to the consideration of energy efficiencies and display quality, such as RGBW\citep{spindler2006system,xiong2009performance}, RGBG Pentile\citep{chae2017subpixel}, and others\citep{fang2013subpixel,elliott200213}. 
We follow the pipeline suggested in \citep{yuan2019aim} to synthesize samples with moiré patterns.
We first collect 19 various subpixel layouts with \wh{the size of $12\times 12$}, as shown in Fig.~\ref{fig:subpixels}. 
\begin{figure*}[htbp]
     \centering
     \includegraphics[width=\textwidth]{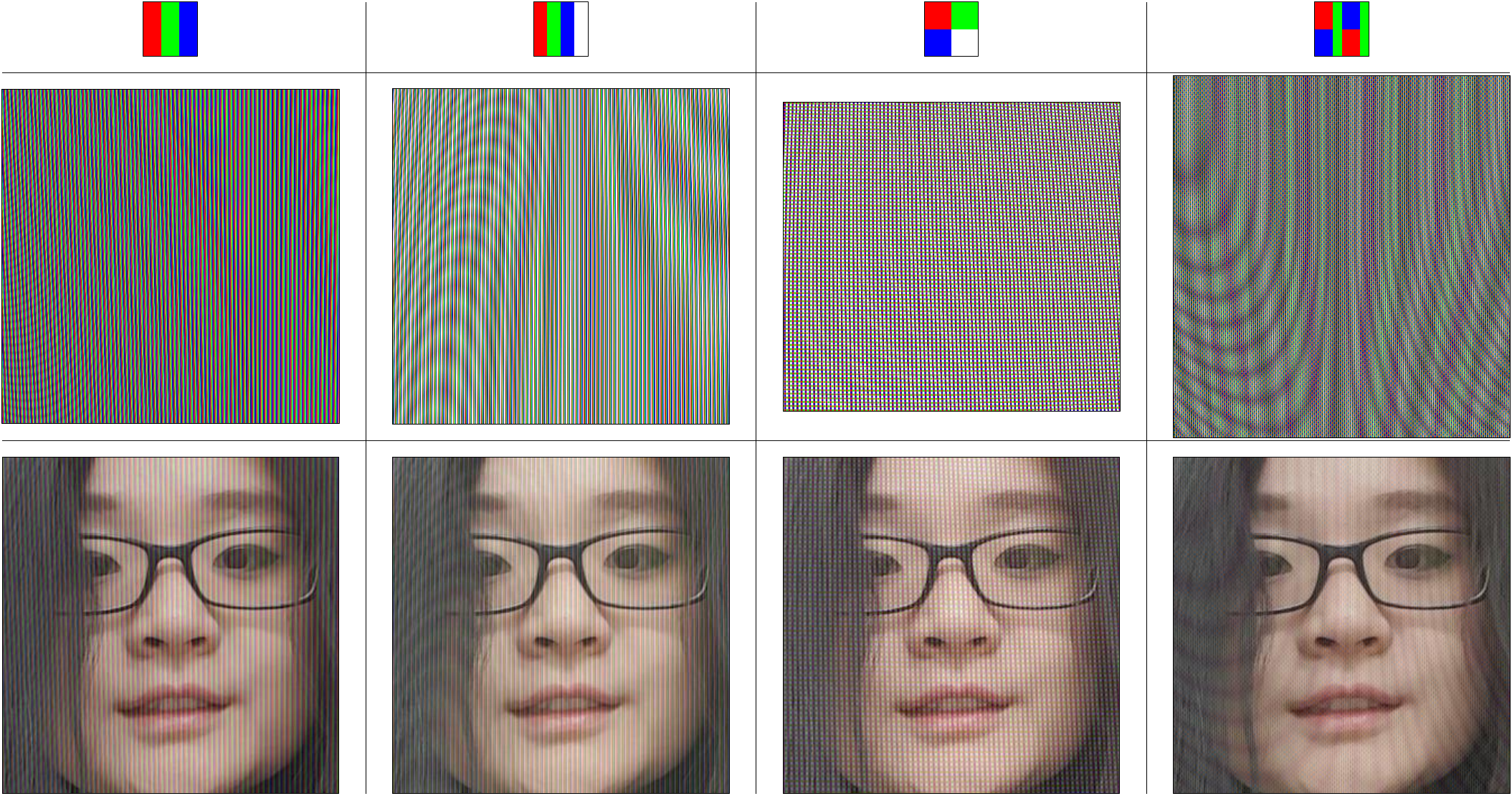}
     \caption{
     The examples of moiré pattern textures (middle row) that are generated from different subpixel layouts (top row), and the examples of applying them to a face image (bottom row). 
     }
     \label{fig:mp_exp}
\end{figure*}
Then, we repeat each subpixel along rows and columns until the size of $1024\times 1024$. 
To simulate the moiré pattern generated by varying relative positions and orientations between the digital display and camera, we perform the random projective transformation within a radius of 0.1 times of the texture's size for each corner.
The random projective transformation is iteratively performed 10 times for each subpixel texture, and eventually, we generate 190 various moiré pattern textures. 
During the augmentation, we randomly select a moiré pattern texture from a uniform distribution, \wh{and} randomly crop the texture to the size of \wh{the} face image.
Finally, we synthesize \wh{the} replay attack image with \wh{the} moiré pattern based on \wh{Eq.~\eqref{eq:reflection}}, where $\mathcal{B}$ becomes the moiré pattern texture and the magnitude controller $\gamma \in [0.01,0.3]$. 
Fig.~\ref{fig:moiré pattern} depicts an example of a face image with the moiré pattern texture formed by the BGR subpixel layout with $\gamma=0.17$. More examples of a face image applying different moiré pattern textures are shown in Fig.~\ref{fig:mp_exp}.

\subsubsection{Printed Photo Attack Recapture Simulation}
The printed photo attack is to present a printed face photo to a camera, and \wh{the} photo can be printed by different printers, which would introduce different \wh{printing} artifacts.
As such, based on \wh{the} printing process, we propose \wh{the} augmentation by simulating different printing artifacts, such as halftone noise, and color distortion.
The above augmentation operations also induce an image's label to be set as \textit{`Spoof'}.

\noindent\textbf{Halftone artifacts simulation} 
Halftoning is a general technique used in the printing process. 
This technique \wh{creates} a gradient-like effect by using dots with different sizes or spacing to approximate continuous-tone imagery \citep{pappas1994digital,dht_2018modern}. 
To simulate the halftone artifacts, we design two augmentation operations motivated by two different halftoning techniques, which are space-filling curve (SFC) dithering\citep{ht_1998space_filling} and blue noise (BN) dithering \citep{ulichney1988dithering}. 
The examples of the two halftone artifacts simulation are shown in Fig.~\ref{fig:halftone} and Fig.~\ref{fig:blue noise} respectively.

To simulate \wh{the} SFC-Halftone artifacts, we first convert the input image to a \wh{gray-scale} and down-sample the image with a scale of $\frac{1}{3}$. 
\begin{figure}[htbp]
     \centering
     \small
     \includegraphics[width=0.45\textwidth]{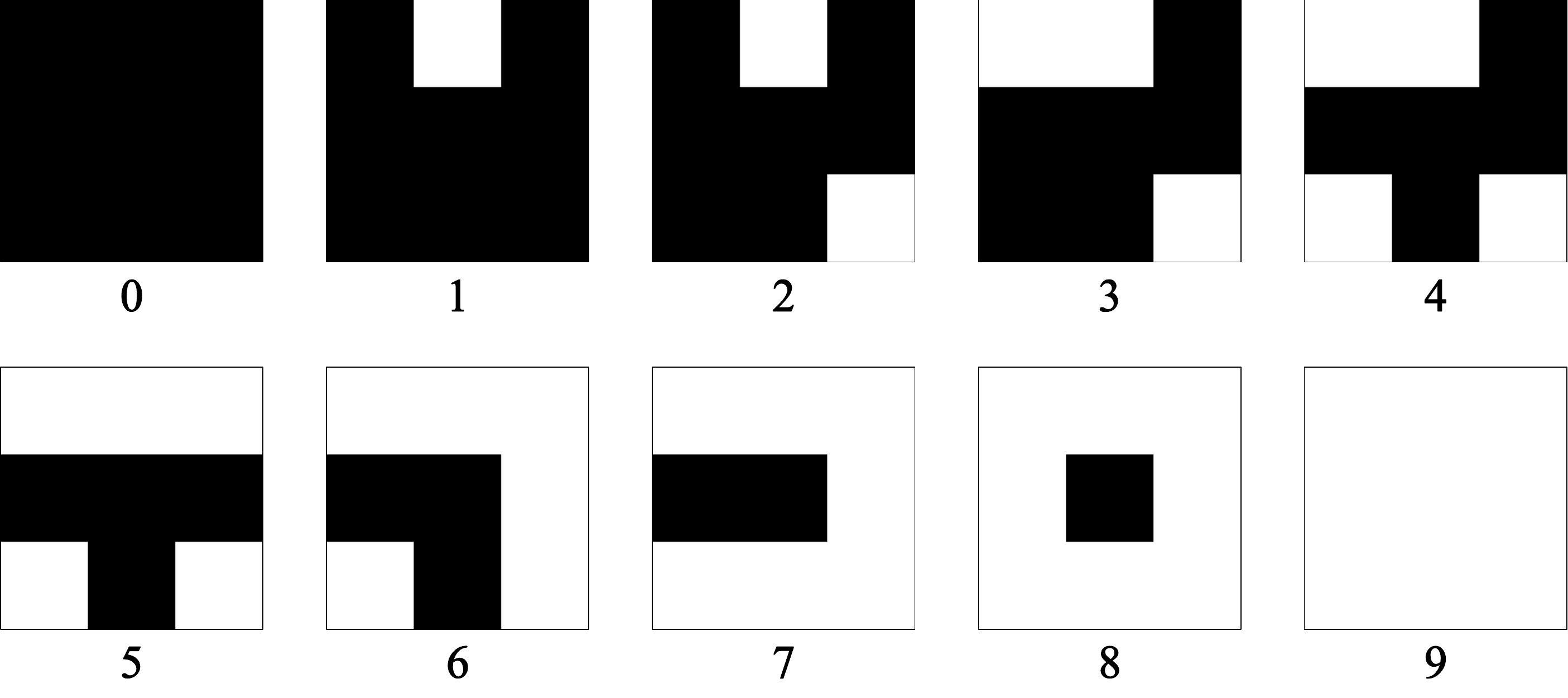}
     \caption{SFC-Halftone dot configuration.}
     \label{fig:halftone_dot}
\end{figure}
After that, we quantize the gray value into 10 levels using the straightforward algorithm suggested in \citep{gabriel_carvalho_2021}, and each level has a corresponding $3\times 3$ dot cluster configuration, as shown in Fig.~\ref{fig:halftone_dot}. Such that, we generate the $3\times 3$ dot cluster for each pixel in the downscaled image based on its gray value, and at the same time, the size is returned to its original value. 

\begin{figure*}[htbp]
     \centering
     \includegraphics[width=\textwidth]{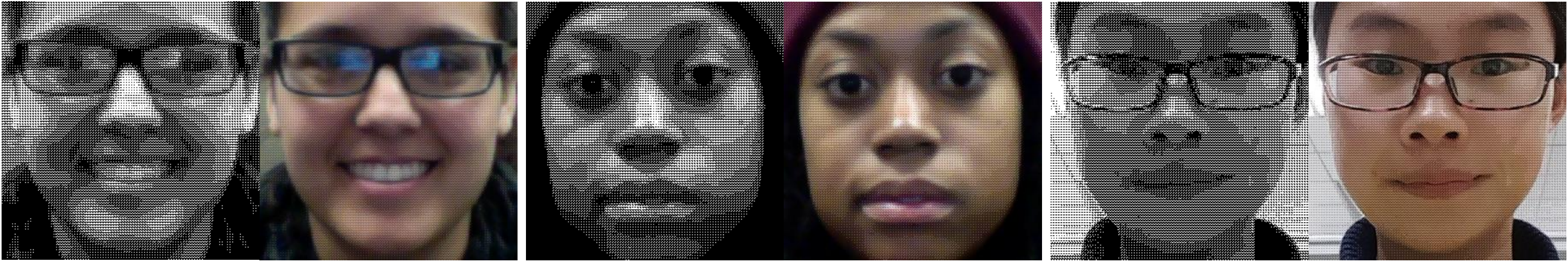}
     \caption{
     The face images applied the SFC-Halftone artifact simulation augmentation. The left shows the clustered-dot halftone image and the right is the resultant image.
     }
     \label{fig:sfc_halftone_exp}
\end{figure*}
\begin{figure*}[htbp]
     \centering
     \includegraphics[width=\textwidth]{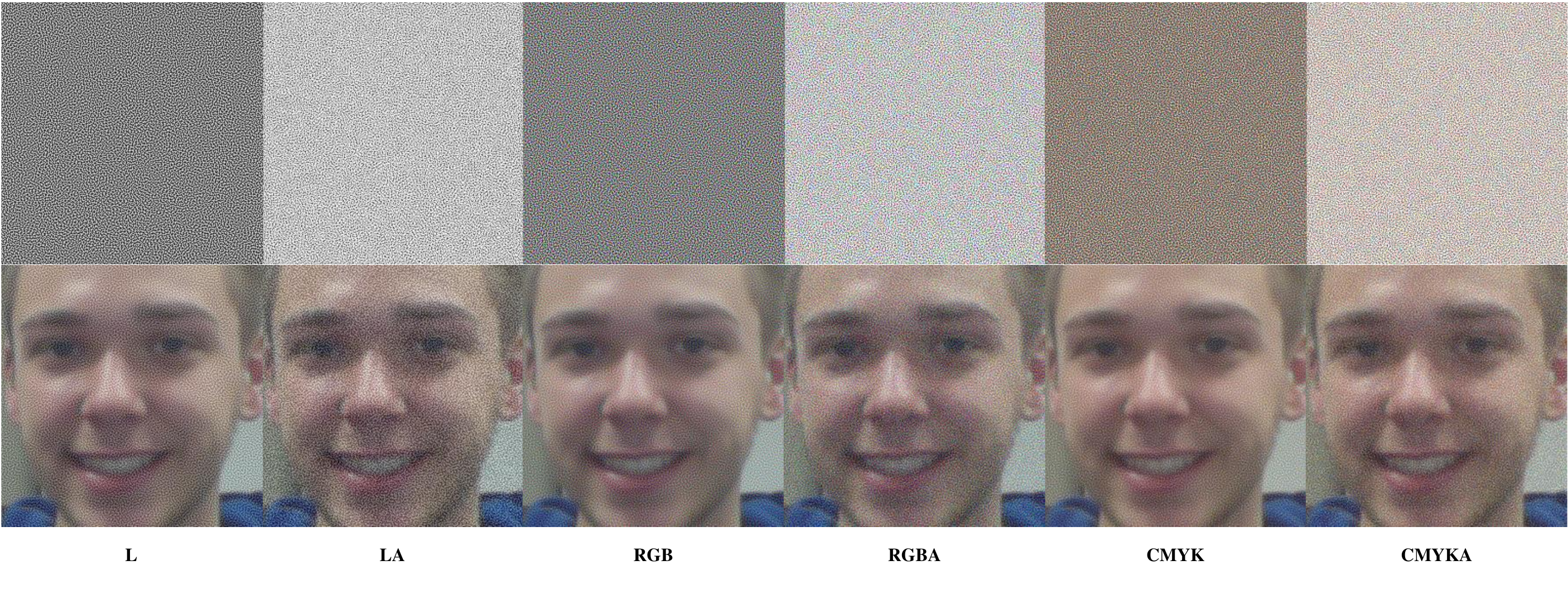}
     \caption{The examples of a face image applied different kinds of blue noise texture.}
     \label{fig:blueNoise_exp}
\end{figure*}
Lastly, we add the halftone artifact to the input face image $\mathcal{I}$ based on \wh{Eq.~\eqref{eq:reflection}}, where $\mathcal{B}$ is the corresponding clustered-dot halftone image and the magnitude controller $\gamma\in [0.01,0.2]$. The examples of clustered-dot halftone images and resultant images are demonstrated in Fig.~\ref{fig:sfc_halftone_exp}.

Besides, to simulate the BN-Halftone artifacts, we follow the blue noise texture synthesis algorithm \citep{peters_2016} to generate six types of blue noise textures with \wh{the} size of 256$\times$256 and each type \wh{is} of eight instances. The types of blue noise textures include gray-scale (L), gray-scale with transparency (LA), RGB, RGB with transparency (RGBA), CMYK and CMYK with transparency (CMYKA). The algorithm uses the void-and-cluster method \citep{ulichney1993void} to generate the blue noise texture $T$, which can simply be expressed as:
\begin{equation} \label{eq:bn}
T = G(H,W,C),
\end{equation}
where $G$ is the blue noise texture synthesis function \citep{peters_2016}, and $H$, $W$ and $C$ respectively denote the height, width, and number of channels of blue noise texture $T$. $C$ is set as 1, 2, 3, and 4 for generating the blue noise texture in the color mode of L, LA, RGB and RGBA, respectively. For generating the blue noise texture in CMYK color space, we convert the RGB blue noise textures to CMYK. For generating the blue noise texture in CMYKA color mode, we similarly convert the RGB blue noise textures to CMYK color space and then use the above algorithm with $C=1$ to generate an extra channel for the transparency channel.

\begin{figure*}[h]
     \centering
     \includegraphics[width=0.9\textwidth]{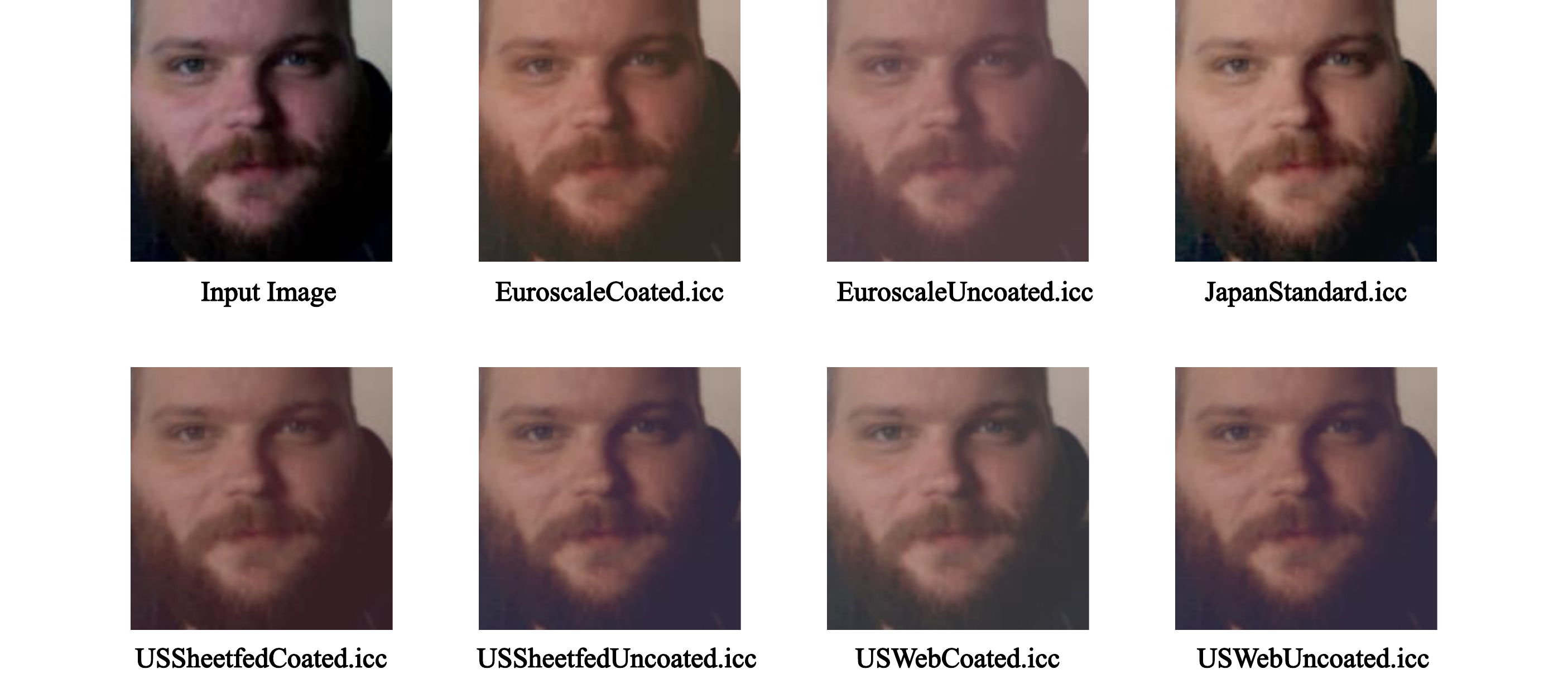}
     \caption{
     The examples of a face image applied the printer color distortion simulation augmentation with different input CMYK color profiles but using a constant output profile, `sRGB.icc'.
     }
     \label{fig:cmyk_exp}
\end{figure*}

After that, we randomly select a blue noise texture and resize it to the size of the face image. 
Then, we fuse the noise texture with the input image $\mathcal{I}$ by \wh{Eq.\eqref{eq:reflection}}, where $\mathcal{B}$ is the blue noise texture and the magnitude controller $\gamma \in [0.01,0.4]$. The examples of applying each type of blue noise texture are shown in Fig.~\ref{fig:blueNoise_exp}. 

\noindent \textbf{Printer color distortion simulation} When an image is printed, its color is distorted after it is mapped to the CYMK color gamut \citep{FAS-ColorTexture-TIFS-2016}, and the color distortion of a printed face photo can be used for FAS analysis \citep{FAS-ColorTexture-TIFS-2016}.
As such, we propose the augmentation by simulating the color distortion effect. We first convert the color mode of \wh{the} input image from RGB to CMYK by the traditional color space conversion equation\citep{cmyk_conversion1998colour}. Given an RGB image $\mathcal{I}(R,G,B)$, the CMYK image $\mathcal{I}(C,M,Y,K)$ is generated as:
\begin{equation}
\mathcal{I}_{i,j}(K) = 1-max(\mathcal{I}_{i,j}(R),\mathcal{I}_{i,j}(G),\mathcal{I}_{i,j}(B)) ,
\end{equation}
\begin{equation}
\mathcal{I}_{i,j}(C,M,Y) = \frac{1-\mathcal{I}_{i,j}(R,G,B)-\mathcal{I}_{i,j}(K)}{1-\mathcal{I}_{i,j}(K)},
\end{equation}
where $(i,j)$ denote the coordinates of pixels of the image.
After that, we convert back the CMYK color mode to RGB color mode by utilizing the ICC transformation \wh{Eq.~\eqref{eq:icc}} to map the CMYK ICC color profile to RGB ICC color profile so that the distorted color can be retained in RGB space.
The Algorithm \ref{alg:code_cmyk} shows how we apply the color distortion simulation augmentation on an input image, \textit{img}. We first use the function \textit{Image.convert()} to perform the traditional color space conversion. After that, similar to the camera color diversity simulation, we use the function, \textit{profiletoprofile()}, from the PIL.ImageCms module, to perform the Eq.~\eqref{eq:icc}. $p_{src}$ is the path of a CMYK ICC color profile that uniformly sampled from 7 open-source CMYK ICC color profiles which we collected from \citep{profile_a}, whereas $p_{dst}$ is the path of an RGB ICC color profile that uniformly sampled from 11 open-source RGB ICC color profiles \citep{profile_a,profiles_b,profiles_c}.
Fig.~\ref{fig:cmyk_exp} shows examples of a face image applied by color distortion augmentation with the 7 different CMYK input profiles and a constant RGB output profile, `sRGB.icc'. In the end, the visual expression of printer color distortion simulation is concluded in Fig.~\ref{fig:cmyk}.
\begin{algorithm}
\caption{Pseudo code of conducting the printer color distortion simulation using Pillow API}\label{alg:code_cmyk}
\small
$\/* p_{img}$: path of the input image*\/\par
$\/* p_{src}$: path of a CMYK ICC color profile*/\par
$/* p_{dst}$: path of a RGB ICC color profile*/\par
img $=$ PIL.Image.open($p_{img}$)\par
img $=$ img.convert(`CMYK')\par
res $=$ PIL.ImageCms.profileToProfile(img, $p_{src}$, $p_{dst}$)\par
\end{algorithm}

\subsection{SARE: Spoofing Attack Risk Equalization}
While the FAS-Aug can bring more diverse data with synthesized artifacts, the augmented spoofing data may make the model overfit to non-invariant artifacts, which could lead to poorer generalization performance.
The non-invariant artifacts in the FAS problem mean that some artifacts can be hints of a spoofing attack, but the absence of those artifacts does not indicate the input is a real face. 
A model relying on non-invariant artifacts could perform poorly in cross-domain testing. 
For example, if a model relies on the moiré pattern for classification when the input is a printed photo attack without the moiré pattern, the model would make the mistake of accepting it as `bona fide'.

To tackle this problem, we propose SARE: Spoofing Attack Risk Equalization. 
The insight behind SARE is that there are various spoofing artifacts, and some prominent artifacts could dominate the optimization, making models overfit and rely on such artifacts for classification. \textcolor{black}{In other words, each dataset might exhibit unique artifacts and thus have unique data distributions due to different capturing conditions. Models might overfit biased distributions dominated by specific artifacts, hence poor generalization performance.}
Therefore, we equalize the risks of different types of spoofing attacks to prevent models from relying on certain types of non-invariant artifacts.
To achieve SARE, we provide a neat implementation in that we minimize the domain-level empirical risks of spoofing attacks.

\begin{equation}
    \mathcal{L}_{SARE} =  Var\{R_1, R_2, ... R_m\}, 
\end{equation}
where $Var$ \wh{represents} the variance calculation, $R_i$ \wh{represents} the empirical risk of spoofing examples from domain $i$, and $m$ means the number of spoofing attack domains. \textcolor{black}{The loss function $\mathcal{L}_{SARE}$ is defined as variance to mitigate the overfitting. During the optimization process, individual $R_i$ value may be large yet $\mathcal{L}_{SARE}$ may remain minimal if the disparities among $R_1, R_2, \ldots, R_m$ are negligible. Conversely, if a particular $R_i$ is small while another $R_j$ is significantly larger, the optimization may become disproportionately influenced by $R_i$, thereby increasing the likelihood of overfitting. By minimizing $\mathcal{L}_{SARE}$, $R_i$ and $R_j$ are balanced to prevent predominating and skewing the model's performance from $R_i$. }

\textcolor{black}{We point out that, $R_i$ does not mean the risk of specific artifacts (moiré patterns) but the risk of attack samples of a domain. This is because defining domain by dataset labels is common in FAS, and defining $R_i$ as the attack risks of domain/dataset $i$ also implies \textbf{distributions} of artifacts in this domain. We do not define domain by the type of artifacts because artifacts and their distributions vary greatly across datasets but there is NO artifact type label, and thus defining $R_i$ according to artifact or augmentation type is difficult. Besides, performing FAS-Aug on examples of domain $i$ can lead to new examples from different distributions, thus creating a new domain $D_j$ and $R_j$. We balance the attack risks of different domains $R_i, R_j ...$ to avoid overfitting to any specific distributions and artifact types.}

\textcolor{black}{In detail, at every batch, we sample a batch of real\&spoof data $B_i$ from a source domain dataset $D_i$. Then, we forward only spoof data to the model and calculate the empirical risk by cross-entropy loss as  $R_i$. Besides, we conduct FAS-Aug on $B_i$ based on the sampled policy, the augmented data is denoted as $B_j$ ($j\neq i$). The FAS-Aug makes the data distribution of $B_j$ different from $B_i$, and thus $R_j$ can represent the attack risks of another domain $j$.}

Moreover, as all real-face examples can be assumed from one domain \citep{SSDG-CVPR-2020}, we align the features of real-face examples only to further improve discriminability between real and spoofing faces. We use the Supervised Contrastive Loss $\mathcal{L}_{Con}$ \citep{SupCon} to \wh{regularize} real-face examples only, which is shown to be \wh{better} than triplet loss as there has no hard-mining problem \citep{SupCon}. We do not align $\mathcal{L}_{Con}$ on the spoofing faces as the spoofing artifacts are diverse and aligning features of different artifacts is non-trivial.

 \begin{table*}[htbp]
  \centering
  \small
  \caption{The list of all our proposed FAS-Aug operations. \textcolor{black}{If the operation is a spoofing-specific recapturing process (Yes), the data augmented with such an operation should be labeled as `Spoof'}  }
  \resizebox{0.95\textwidth}{!}{
    \begin{tabular}{rlcc}
    \hline
    \multicolumn{1}{l}{Operation Name} & \multicolumn{1}{c}{Description} & Magnitude & \textcolor{black}{\textbf{Spoofing-specific recapturing process}} \\
    \hline
    \multicolumn{1}{l}{ColorDiversity} & Change the image's color by mapping the RGB color profile from one  & 11 RGB color profiles & \textcolor{black}{No} \\
      & to another, in which both color profiles are randomly selected. &   &  \\
    \hline
    \multicolumn{1}{l}{LowResolution} & Desample the image size with a scale of $s$, then resize to the initial & $s\in[0.01, 1]$ & \textcolor{black}{No} \\
      &  value. &   &  \\
    \hline      
    \multicolumn{1}{l}{HandTrembling} & Add the motion blur effect with a certain direction on the image by & $\textit{k}\in[1, 16]$ & \textcolor{black}{No} \\
      & doing convolution with a kernel of size \textit{k}. The direction includes &   &  \\
      & horizontal, vertical, diagonal and anti-diagonal, which is randomly  &   &  \\      
      & selected. &   &  \\            
    \hline    
    \hline  
    \multicolumn{1}{l}
    {Specular Reflection} & Add another background image to the image with a ratio $\gamma$. The  & $\gamma\in[0.03, 0.2]$ & \textcolor{black}{Yes} \\
      & background image is randomly sampled from 90 instances. &   &  \\
    \hline      
    \multicolumn{1}{l}{moiré pattern} & Add a moiré pattern texture to the image with a ratio $\gamma$. The moiré   & $\gamma\in[0.01, 0.3]$ & \textcolor{black}{Yes} \\
      & pattern texture is randomly sampled from 190 instances. &   &  \\
    \hline      
    \multicolumn{1}{l}{SFCHalftone} & Transform the image to an SFC-Halftone image and add the SFC-Halftone  & $\gamma\in[0.01, 0.2]$ & \textcolor{black}{Yes} \\
      & image to the initial image with a ratio $\gamma$. &   &  \\
    \hline      
    \multicolumn{1}{l}{BNHalftone} & Add a blue noise texture to the image with a ratio $\gamma$. The blue noise  & $\gamma\in[0.01, 0.4]$ & \textcolor{black}{Yes} \\
      & texture is randomly sampled from 48 instances. &   &  \\
    \hline      
    \multicolumn{1}{l}{Color Distortion} & Change the image's color space to CMYK, and perform the mapping of  & 7 CMYK color profiles  & \textcolor{black}{Yes} \\
      &  CMYK to RGB color profiles to retain the CMYK color space in RGB  & 11 RGB color profiles &  \\
      & mode, in which both color profiles are randomly selected. &   &  \\
    \hline
    \end{tabular}%
    }
  \label{tab:FAS_Aug operation}%
\end{table*}%

\begin{figure*}[htbp]
     \centering
     \includegraphics[width=.8\textwidth]{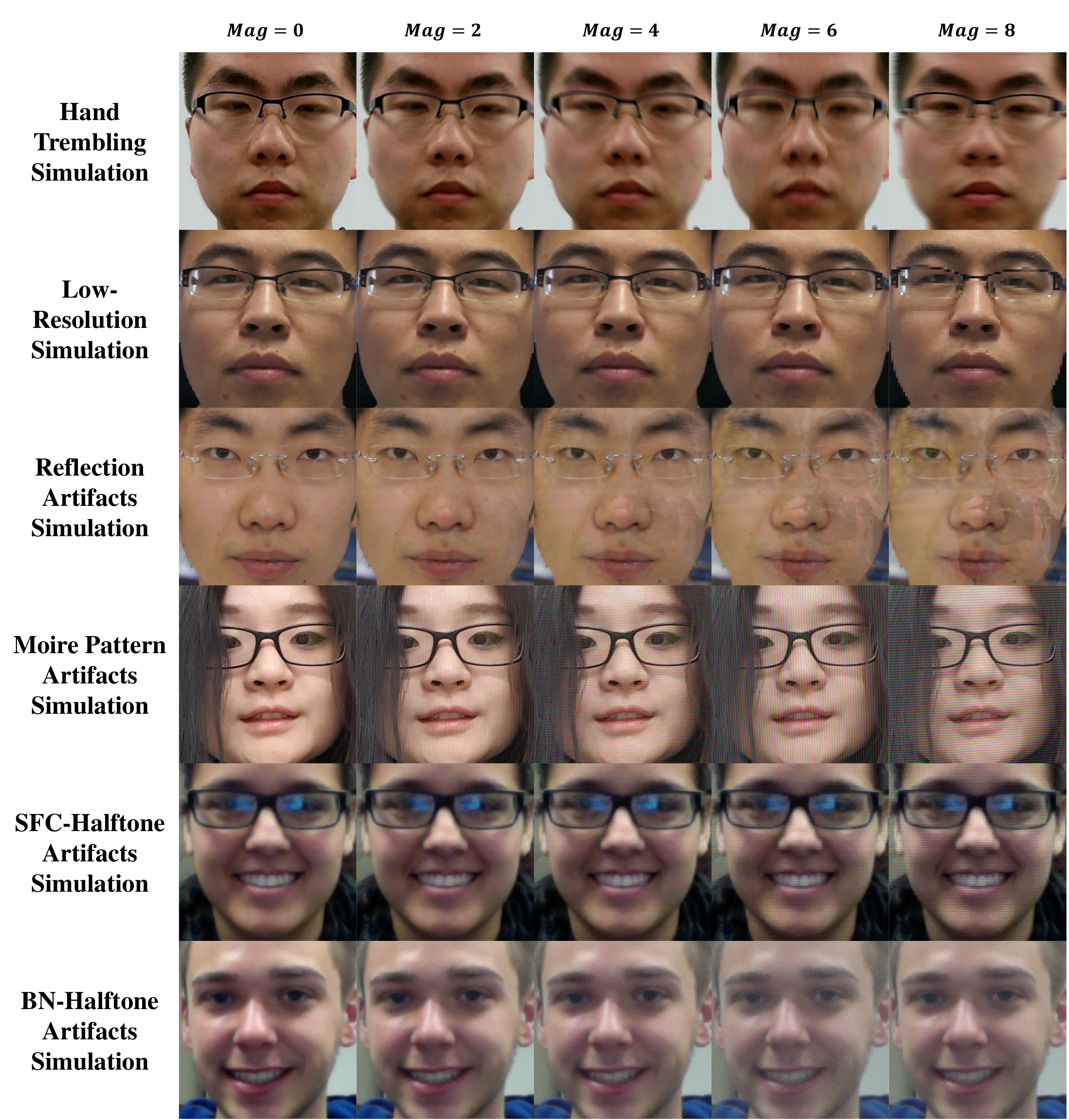}
     \caption{The examples of a face image applied the simulation augmentation with different magnitudes. }
     \label{fig:magnitude}
\end{figure*}
\begin{table*}[htbp]
  \centering
   \caption{Comparing results of TI-Aug (\&TA) and our FAS-Aug (\&FA) with different backbones. On the right side are the average results of HTER and AUC results from the left. } \label{tab:compare-TA-FA}
 \resizebox{0.95\textwidth}{!}{
    \begin{tabular}{l|cccccccc|cc}
    \hline
    \multicolumn{1}{c}{\multirow{2}[2]{*}{Method}} & \multicolumn{2}{|c}{C\&I\&O to M} & \multicolumn{2}{c}{O\&M\&I to C} & \multicolumn{2}{c}{O\&C\&M to I} & \multicolumn{2}{c}{I\&C\&M to O  } & \multicolumn{2}{c}{Average} \bigstrut[t]\\
    \cline{2-11}
          & HTER(\%) & AUC(\%) & HTER(\%) & AUC(\%) & HTER(\%) & AUC(\%) & HTER(\%) & AUC(\%) & \multicolumn{1}{c}{HTER(\%)} & \multicolumn{1}{c}{AUC(\%)} \bigstrut[b]\\
    \hline
    ResNet-18\&TA & 16.12 & 89.51 & 18.00 & 89.93 & 22.84 & 84.53 & 21.43 & 85.13 & 19.60 & 87.27 \bigstrut[t]\\
    ResNet-18\&FA(ours) & 10.34 & 94.23 & 21.50 & 87.40 & 21.94 & 84.01 & 15.13 & 92.29 & 17.23 & 89.48 \\
    ResNet-18\&FA\&CS(ours) & 8.04 & 96.21 & 19.50 & 87.46 & 20.67 & 86.74 & 12.10 & 94.32 & \textbf{15.08} & \textbf{91.18} \bigstrut[b]\\
    \hline
    ViT-Adapter\&TA & 15.56 & 91.71 & 19.11 & 87.87 & 16.47 & 82.34 & 17.73 & 89.88 & 17.22 & 87.95 \bigstrut[t]\\
    ViT-Adapter\&FA(ours) & 10.57 & 94.70 & 17.17 & 91.20 & 16.70 & 85.82 & 16.39 & 91.15 & 15.21 & 90.72 \\
    ViT-Adapter\&FA\&CS(ours) & 6.82 & 97.54 & 17.61 & 89.89 & 12.85 & 94.16 & 15.05 & 92.30 & \textbf{13.08} & \textbf{93.47} \bigstrut[b]\\
    \hline
    ViT-Convpass\&TA & 13.33 & 94.67 & 7.61  & 97.20 & 14.44 & 93.43 & 10.74 & 95.82 & 11.53 & 95.28 \bigstrut[t]\\
    ViT-Convpass\&FA(ours) & 5.29 & 97.41 & 7.89  & 96.54 & 14.22 & 93.85 & 8.31  & 97.41 & 8.93 & 96.30 \\
    ViT-Convpass\&FA\&CS(ours) & 4.62 & 98.92 & 7.28 & 97.02 & 10.89 & 97.05 & 6.77 & 98.25 & \textbf{7.39}  & \textbf{97.81} \bigstrut[b]\\
    \hline
    \end{tabular}
    }
\end{table*}%
As such, we derive the overall optimization loss function as \begin{equation}
    \mathcal{L} = \mathcal{L}_{BCE} + \alpha \mathcal{L}_{Con} + \beta \mathcal{L}_{SARE},
\end{equation}
where $\mathcal{L}_{BCE}$ is binary cross-entropy loss function, and $\alpha$ and $\beta$ are the constant scaling factor. 
\par
In our experiments, we use ResNet-18 as the representative of convolutional neural networks (ConvNet). Besides, we also use Vision Transformer (ViT) with Adapter \citep{pfeiffer2020adapterfusion}, which has been shown to be more effective than vanilla ViT for the FAS problem \citep{fas-adapter-eccv2022}. 
Furthermore, we also first introduce ViT-Convpass in the FAS problem, which brings vision-specific inductive bias to ViT for the FAS task.
In our experiment, ViT-Adapter and ViT-Convpass \citep{convpass} use the same ViT backbone of `\textit{vit\_base\_patch16\_224}' \citep{ViT}.
 
\section{Experiment}
\subsection{Datasets and protocols.}
Our experiments involve six benchmark datasets CASIA-FASD (\textit{C}) \citep{DB-CASIAFASD}, IDIAP REPLAY ATTACK (\textit{I}) \citep{LBP-FAS-BIOSIG-2012}, MSU MFSD (\textit{M}) \citep{FAS-IDA-TIFS-2015}, OULU-NPU (\textit{O}) \citep{OULU_NPU_2017}, NTU ROSE-YOUTU (\textit{Y})\citep{FAS-UnsupervisedDA-TIFS-2018}, and SiW (\textit{S})\citep{FAS-Auxiliary-CVPR-2018}.
Following \citep{MADDG-CVPR-2019}, we utilize the leave-one-out cross-domain protocol\citep{MADDG-CVPR-2019}, which uses the four datasets \textit{M}, \textit{I}, \textit{C}, and \textit{O}. We follow \citep{metapattern} to abbreviate this protocol as MICO for short. Also, we follow \citep{wang2020cross, metapattern} to use the MICY protocol to provide more extensive cross-domain evaluation, which uses the datasets of \textit{M}, \textit{I}, \textit{C}, and \textit{Y}. 
Moreover, to explore the situation when there are more real-face examples, we propose a new protocol MICO+SY, where the real-face examples of $\textit{S}$ and $\textit{Y}$ are added into the training when doing the leave-one-out experiment based on MICO. For comparisons, we use Half-Total Error Rate (HTER, \textcolor{black}{the lower the better}) and Area Under the receiver operating characteristic Curve (AUC, the higher the better).

\subsection{Implementation details}
\textbf{Data Processing} We use the \textit{dlib} \citep{dlib09}  detector to detect and crop face images,  and \wh{neither} special preprocessing nor alignment is needed. All face images are resized to $224\times224$ for network input to models. We train models with a maximum of 200 epochs, using the Adam optimizer and the learning rate of 0.0001. 

\noindent \textbf{Augmentation operation of TI-Aug} \textcolor{black}{Following the state-of-the-art augmentation study \cite{adv_auto}, we use ShearX/Y, TranslateX/Y, Rotate, AutoContrast, Invert, Equalize, Solarize, Posterize, Contrast, Color, Brightness, Sharpness, and Cutout as for the Traditional Augmentation operations. For the magnitude range of TI-Aug, please refer to AutoAugment\citep{adv_auto}. }

\noindent \textbf{Augmentation operation of our FAS-Aug} 
\textcolor{black}{All our proposed FAS-Aug operations and the magnitude range used for sampling the parameters during training are shown in Table \ref{tab:FAS_Aug operation}. As can be seen from the table, there are six operations with a range where the parameters are sampled. The left endpoint and right endpoints of the ranges are selected empirically, which are based on common occurrences.   }

\noindent \textcolor{black}{Similar to \citep{cubuk2019autoaugment,adv_auto}, we discretize the range of magnitudes into 10 values uniformly for both types of augmentation with numerical magnitudes. The results of different discretized magnitudes of our proposed FAS-Aug are demonstrated in Fig.~\ref{fig:magnitude}.}

\begin{figure}
    \centering
    \includegraphics[width=0.5\textwidth]{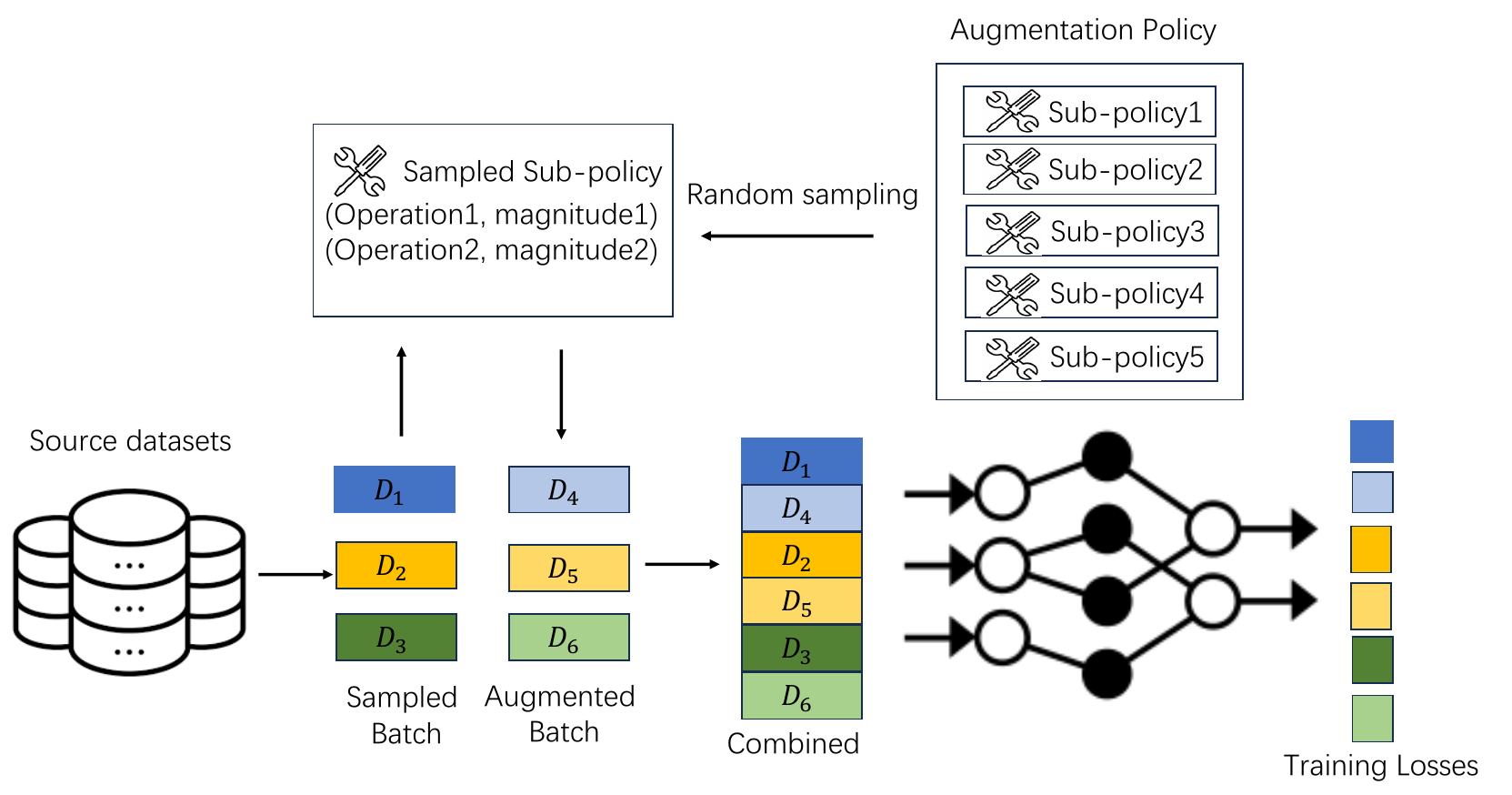}
    \caption{\textcolor{black}{Illustration of sampling augmentation policy. At the beginning of each epoch,  an augmentation policy for augmentation is sampled. The policy contains 5 sub-policies to be sampled. Each sub-policy contains two processes and one process is defined by the augmentation operation and the corresponding magnitude, which are sampled according to Table 1 of the revised manuscript.}}
    \label{fig:policy}
\end{figure}
\noindent 
\textbf{Augmentation sampling during training}
\textcolor{black}{We follow \cite{adv_auto} to set up the sampling policy of augmentation, which is depicted in Fig.\ref{fig:policy}. At each epoch, we randomly set up an augmentation policy. Each policy contains 5 sub-policies. Each sub-policy has two augmentation operations, and each operation has a corresponding magnitude controller. For each image, only a sub-policy is sampled uniformly at random, and the two corresponding operations are applied sequentially to the image.  }
\noindent \textbf{Calculation of ${R_i}$} \textcolor{black}{In the context of FAS, a ``domain" refers to a unique data distribution, typically characterized by specific environmental factors affecting data capture. Such factors can introduce variations in the data, known as domain or distribution shifts, complicating the FAS challenge. For instance, different datasets in FAS have their unique capture environments -- each with distinct characteristics that affect the data's distribution. This variability underscores the rationale for associating each dataset with a specific domain, and thus a domain is usually defined according to a dataset in the practice of FAS.} When sampling a data batch (${B_i}$) from a dataset (${i}$) during training, we consider it as originating from domain ${D_i}$. Through the application of FAS-Aug, we generate a new batch ${B_j}$ which, due to undergoing different capturing and processing conditions, is associated with a new domain (${D_j}$), distinct from (${D_i}$). \textcolor{black}{In experiments utilizing the SARE technique, we calculate the attack risk (${R_i}$) using attack examples from a dataset (${D_i}$). FAS-Aug introduces a novel aspect to this process by incorporating a recapturing simulation operation within a sub-policy. This operation labels processed samples as `Spoof', creating a new domain (${D_j}$) for computing a different attack risk (${R_j}$).}

\noindent \textbf{Computational cost} 
There is extra computational cost proportional to $m$, as the risks, $R_1$, $R_2$ ... $R_m$ are needed to be calculated. We highlight that by properly managing the data loading and inference, the cost can be properly controlled. Referencing Fig.~\ref{fig:policy}, we first sample data from three domains, yielding three data batches: $B_1$, $B_2$, and $B_3$. Subsequently, we apply a sampled sub-policy to each batch to generate three additional batches: $B_4$, $B_5$, and $B_6$, setting $m=6$. To compute $R_1$, $R_2$, $R_3$, $R_4$, $R_5$, and $R_6$, a single inference suffices, obviating the need for six separate inferences. In practice, these batches are consolidated into a larger batch $B$, which is then processed in a single network inference step with $B$ as input, producing the output logit $Y$. Leveraging the batch order, slicing operations extract $Y_1$, $Y_2$, $Y_3$, $Y_4$, $Y_5$, and $Y_6$ for $R$ value calculations. This approach significantly reduces computational overhead, as the inference can be conducted in parallel with the Graphic Processing Unit (GPU). \textcolor{black}{As only a single inference is required, the memory constraint depends on the size of $B$. When the available GPU's computational capability is limited, such as with an NVIDIA 1080Ti (11GB), it is common to use a medium batch size ($S^B$) for model training, typically 32, 64, or 128, to fit within the GPU's memory constraints. Therefore, when sampling small data batches $B_1$, $B_2$, ... $B_m$, the sampling size should be chosen such that $\lfloor S^B/m\rfloor$ is considered.}

\begin{table*}[tbp]
  \centering
  \caption{Training models based on the MICO protocol but with extra real-face examples from SiW \citep{FAS-Auxiliary-CVPR-2018} and ROSE-YOUTU \citep{FAS-UnsupervisedDA-TIFS-2018} (`+SY').}\label{tab:MICO+SY}
  \resizebox{0.95\textwidth}{!}{
    \begin{tabular}{lcccccccc}
    \hline
    \multicolumn{1}{c}{\multirow{2}[2]{*}{ ViT-Convpass}} & \multicolumn{2}{c}{C\&I\&O+SY to M} & \multicolumn{2}{c}{O\&M\&I+ SY to C} & \multicolumn{2}{c}{O\&C\&M+SY  to I} & \multicolumn{2}{c}{I\&C\&M+SY to O  }  \bigstrut[t]\\
    \cline{2-9}
          & HTER(\%) & AUC(\%) & HTER(\%) & AUC(\%) & HTER(\%) & AUC(\%) & HTER(\%) & AUC(\%) \bigstrut[b]\\
    \hline
    \&TA & 24.03 & 82.23 & 12.94 & 93.05 & 17.45 & 87.41 & 9.42  & 95.95  \bigstrut[t]\\
    \&FA & 19.95 & 86.30 & 9.00  & 95.77 & 12.61 & 87.34 & 9.17  & 96.80 \\
    \&FA\&CS & \textbf{6.92}  & \textbf{98.28} & \textbf{8.06}  & \textbf{96.98} & \textbf{12.20} & \textbf{93.87} & \textbf{8.50}  & \textbf{97.20}  \bigstrut[b]\\
    \hline
    \end{tabular}%
    }
\end{table*}%
\begin{table}[htbp]
  \centering
  \caption{\textcolor{black}{The effectiveness of our FAS-Aug on the state-of-the-art method SSDG \citep{SSDG-CVPR-2020}, SSAN \citep{SSAN}, FLIP-MCL \citep{FLIP-MCL}. `*' means that the experimental results are obtained by our implementation with their officially released code, and no supplemented dataset CelebA-Spoof is used. `+SY' means the training data includes the real-face examples from SiW and ROSE-YOUTU datasets.}}
  \begin{tabular}{c|c|c|c|c}
    \hline
    \multirow{2}[4]{*}{C \&M \&I to O} & \multicolumn{2}{c|}{SSDG-R*} & \multicolumn{2}{c}{SSDG-R*+SY} \bigstrut\\
\cline{2-5}          &  \&TA &  \&FA &  \&TA &  \&FA \bigstrut\\
    \hline
    HTER (\%)  & 16.73 & \textbf{15.25} & 15.57 & \textbf{13.13} \bigstrut\\
    \hline
    AUC (\%)    & 90.84 & \textbf{93.17} & 92.62 & \textbf{94.90} \bigstrut\\
    \hline
    \hline
    \multicolumn{1}{r|}{\multirow{2}[4]{*}{C \&M \&I to O}} & \multicolumn{2}{c|}{SSAN-R*} & \multicolumn{2}{c}{SSAN-R*+SY} \bigstrut\\
\cline{2-5}          &  \&TA &  \&FA &  \&TA &  \&FA \bigstrut\\
    \hline
    HTER (\%)   & 16.92 & \textbf{13.79} & 15.01 & \textbf{12.30} \bigstrut\\
    \hline
    AUC (\%)    & 90.72 & \textbf{93.94} & 92.30 & \textbf{94.46} \bigstrut\\
    \hline
    \hline
    \multirow{2}[4]{*}{C \&M \&I to O} & \multicolumn{2}{c}{FLIP-MCL*} & \multicolumn{2}{c}{FLIP-MCL*+SY} \bigstrut\\
\cline{2-5}          &  \&TA &  \&FA &  \&TA &  \&FA \bigstrut\\
    \hline
    HTER (\%)   & 8.69  & \textbf{7.41}  & 8.97  & \textbf{7.79} \bigstrut\\
    \hline
    AUC (\%)    & 96.95 & \textbf{97.46} & 96.38 & \textbf{97.07} \bigstrut\\
    \hline
    \end{tabular}%
  \label{tab:FA-on-sota}%
\end{table}%

\subsection{Experimental results}
In this section, we present experimental results. In the table and figures, we use the suffix after a backbone name to represent different settings: `backbone\&TA' indicates the results with Traditional Augmentation; `backbone\&FA' indicates the results with our FAS-Aug only; `backbone\&FA\&CS' indicates the results with our FAS-Aug, $\mathcal{L}_{Con}$ and $\mathcal{L}_{SARE}$.

\subsubsection{Effectiveness of the proposed FAS Augmentation}
\noindent \textbf{Comparison with TI-Aug.} 
As shown in Table~\ref{tab:compare-TA-FA},  in the experiments of `O\&M\&I to C', the HTER and AUC results of `ResNet-18\&FA' is not better than `ResNet-18\&TA'. We conjecture that the model are relying on the augmented non-invariant artifacts for classification. When the target testing lacks the augmented artifacts, errors would be increased. 
Whereas, the errors can be reduced by learning more generalized features with our method based on SARE. Moreover, if we compare our FAS-Aug to TI-Aug with different backbones in different experiments with the average HTER and AUC performance over the four sub-protocols, we can see that our FAS-Aug outperforms the TI-Aug, and fits the FAS problem better \wh{with} augmented data diversity, and our proposed method based on SARE can further help to learn more generalized features and make improvement. 
Besides, by comparing different backbones, we find ViT-Convpass \citep{convpass} is a more effective backbone than ResNet-18 and ViT-Adapter for the FAS problem. Thus, we use ViT-Convpass in \wh{the} below experiments.

\noindent \textbf{Exploration of more real-face examples}
In practical real-world scenarios, there could be more real-face examples than the spoofing ones, but this case is largely ignored by previous works.  
To study this ignored case, we extract the real-face examples from ROSE-YOUTU and SiW datasets into the MICO benchmark and conduct experiments. \wh{Counterintuitively}, we find that directly adding real-face examples may not directly benefit the performance when comparing results of `ViT-Convpass\&TA/\&FA' in Table~\ref{tab:compare-TA-FA} and Table~\ref{tab:MICO+SY}. 
We conjecture the reason that when introducing more real face \wh{examples} (+SY) in training, more real-face examples also create imbalance and bring \wh{poor} performance if only simple cross-entropy loss is used. Still, we can see from Table~\ref{tab:MICO+SY} that our FAS-Aug and SARE generally \wh{outperforms} TI-Aug by a large margin when working with more real-face examples.

\noindent\textbf{Using FAS-Aug with SOTA methods} \textcolor{black}{
We also try our FAS-Aug with a state-of-the-art method SSDG-R\footnote{https://github.com/taylover-pei/SSDG-CVPR2020} \citep{SSDG-CVPR-2020}, SSAN\footnote{https://github.com/wangzhuo2019/SSAN}\citep{SSAN}, and FLIP-MCL\footnote{https://github.com/koushiksrivats/FLIP} \citep{FLIP-MCL} with the officially released code and report the results in Table~\ref{tab:FA-on-sota}. Despite using the official implementation, there are some gaps between the reimplementation and the reported results of SSDG, SSAN, and FLIP-MCL. Nevertheless, in our experiments using TA or FA, the settings are the same, and thus the comparison is fair. In Table~\ref{tab:FA-on-sota}, our FAS-Aug can help SSDG-R(\&FA), SSAN-R(\&FA), FLIP-MCL(\&FA) achieve better AUC and HTER performance than themselves with TI-Aug (SSDG-R\&TA). Moreover, when adding real-face examples from the ROSE-YOUTU and SiW datasets (`+SY'), our FAS-Aug can bring more significant HTER and AUC performance improvement than TI-Aug to these methods with the increased data diversity.
}

Through the above discussion, we can see that our proposed Face Anti-Spoofing Augmentation can provide FAS-specific data diversity and can generally outperform traditional image augmentation and fit the FAS task better. Also, our proposed FAS-Aug can be used with other methods, such as SSDG-R \citep{SSDG-CVPR-2020}.

\begin{table}[tbp]
  \centering
 \caption{HTER and AUC performance of ViT-Convpass with ($\checkmark$) or without ($\times$) $\mathcal{L}_{Con}$ and $\mathcal{L}_{SARE}$.}
  \resizebox{0.45\textwidth}{!}{
    \begin{tabular}{ccccc}
    \hline
    \multirow{2}[2]{*}{I\&C\&M to O} & \multicolumn{1}{l}{$\times$ $\mathcal{L}_{Con}$} & \multicolumn{1}{l}{$\checkmark$$\mathcal{L}_{Con}$} & \multicolumn{1}{l}{$\times$ $\mathcal{L}_{Con}$} & \multicolumn{1}{l}{$\checkmark$ $\mathcal{L}_{Con}$} \bigstrut[t]\\
          & \multicolumn{1}{l}{$\times$ $\mathcal{L}_{SARE}$} &  \multicolumn{1}{l}{$\times\mathcal{L}_{SARE}$}&\multicolumn{1}{l}{$\checkmark$\ $\mathcal{L}_{SARE}$} & \multicolumn{1}{l}{$\checkmark$\ $\mathcal{L}_{SARE}$} \bigstrut[b]\\
    \hline
    HTER (\%) & 8.31 &7.43  &7.81  & \textbf{6.77} \bigstrut[t]\\
    \hline
    AUC (\%) & 97.41 & 98.11 &98.03 & \textbf{98.25} \bigstrut[b]\\
    \hline
    \end{tabular}
    }%
  \label{tab:ablation}%
\end{table}%
\begin{table*}[tbp]
  \centering
 
  \caption{Experimental on the leave-one-out benchmark MICO. Results are in terms of HTER (\%) and AUC (\%).}
    \resizebox{0.95\textwidth}{!}{
    \begin{tabular}{l|c|c|c|c|c|c|c|c}
    \hline
    \multicolumn{1}{c|}{\multirow{2}[4]{*}{Method}} & \multicolumn{2}{c|}{C\&I\&O to M} & \multicolumn{2}{c|}{O\&M\&I to C} & \multicolumn{2}{c|}{O\&C\&M to I} & \multicolumn{2}{c}{I\&C\&M to O} \bigstrut\\
\cline{2-9}      & HTER(\%) & AUC(\%) & HTER(\%) & AUC(\%) & HTER(\%) & AUC(\%) & HTER(\%) & AUC(\%) \bigstrut\\
    \hline
    MMD-AAE \citep{MMDAAE-CVPR-2018}   &  27.08 & 83.19 & 44.59 & 58.29 & 31.58 & 75.18 & 40.98 & 63.08\\
    \hline 
    MADDG \citep{MADDG-CVPR-2019}   & 17.69 & 88.06 & 24.5 & 84.51 & 22.19 & 84.99 & 27.98 & 80.02\\
    \hline
    RFMetaFAS \citep{RFMetaFAS-AAAI-2020}   & 13.89      &  93.98     &  20.27     &     88.16  &    17.30   &  90.48     &  16.45     &   91.16\\
    \hline
    NAS-Baesline w/ D-Meta \citep{NASFAS-TPAMI-2020} & 11.62 & 95.85 & 16.96 & 89.73 & 16.82 & 91.68 & 18.64 & 88.45 \\
    \hline
    NAS w/ D-Meta \citep{NASFAS-TPAMI-2020} & 16.85 & 90.42 & 15.21 & 92.64 & 11.63 & 96.98 & 13.16 & 94.18 \\
    \hline
    NAS-FAS \citep{NASFAS-TPAMI-2020} & 19.53 & 88.63 & 16.54 & 90.18 & 14.51 & 93.84 & 13.80 & 93.43 \\
    \hline
    SSDG-M \citep{SSDG-CVPR-2020}& 16.67 & 90.47 & 23.11 & 85.45 & 18.21 & 94.61 & 25.17 & 81.83 \\
    \hline
    SSDG-R  \citep{SSDG-CVPR-2020} & 7.38  & 97.17 & 10.44 & 95.94 & 11.71 & 96.59 & 15.61 & 91.54 \\
    \hline
    FAS-DR-BC(MT) \citep{MetaTeacher-TPAMI-2021}   &     11.67    &  93.09     &  18.44   &  89.67     &     11.93  &    94.95   &  16.23     & 91.18 \\
    \hline
    SSAN \citep{SSAN} & 6.57 & 98.78 & 10.00 & 96.67 &  \textbf{8.88} & 96.79 & 13.72 & 93.62 \\
    \hline
    PatchNet \citep{wang2022patchnet} & 7.10 & 98.46 & 11.33 & 94.58 &  13.4 & 95.67 & 11.82 & 95.07 \\ \hline
    AMEL \citep{zhou2022adaptive} & 10.23 & 96.62 & 11.88 & 94.39 &  18.60 & 88.79 & 11.31 & 93.36\\  \hline
    HFN+MP \citep{metapattern}  & 5.24  &	97.28 &	9.11 &	96.09 &	15.35 &	90.67 & 12.4&	94.26 \\
    \hline
    \hline
    ViT-Convpass\&FA\&CS (ours)& \textbf{4.62} & \textbf{98.92} & \textbf{7.28} & \textbf{97.02} & 10.89 & \textbf{97.05} & \textbf{6.77} & \textbf{98.25} \bigstrut\\
    \hline

    \end{tabular}%
    }
  \label{tab:sota-mico}%
\end{table*}%
\subsubsection{Ablation study}
\begin{figure}
     \centering
     \small
     \begin{subfigure}[c]{0.4\textwidth}
         \centering
         \includegraphics[width=\textwidth]{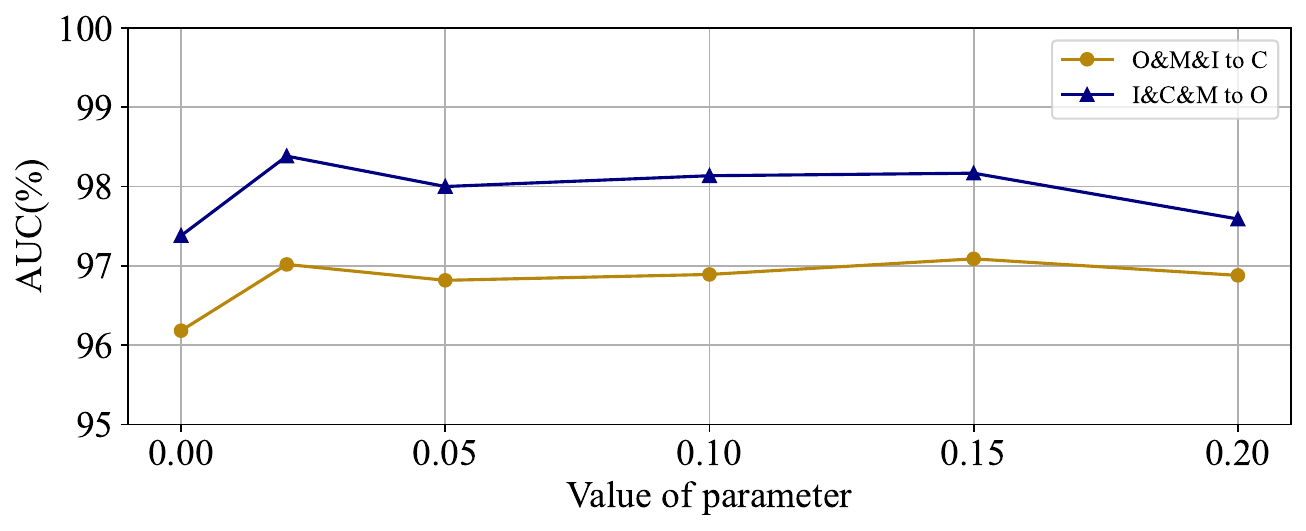}
         \subcaption{$\alpha \in [0,0.2]$ ($\beta$=10)}
         \label{fig:alpha analysis}
     \end{subfigure}
     \hspace{0.07cm}
     \begin{subfigure}[c]{0.4\textwidth}
         \centering
         \includegraphics[width=\textwidth]{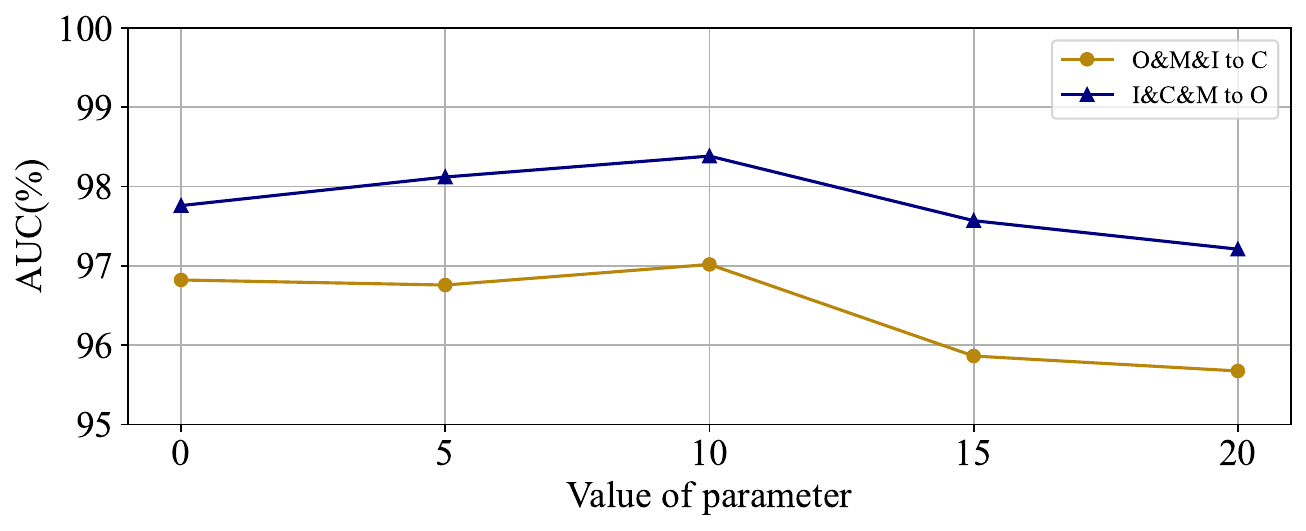}
         \subcaption{$\beta \in [0,20]$ ($\alpha$=0.02)}
         \label{fig:beta analysis}
     \end{subfigure}   
        \caption{Hyper-parameter analysis with ViT-Convpass. The sub-figure (a) shows the performance fixing $\beta=10$ and change $\alpha$ from 0.0 to 0.2. The sub-figure (b) shows the performance of fixing $\alpha=0.02$ and change $\beta$ from 0 to 20.}
        \label{fig:parameter analysis}
\end{figure}

\noindent \textbf{Effectiveness of $\mathcal{L}_{Con}$ and $\mathcal{L}_{SARE}$} 
In Table~\ref{tab:ablation}, we remove $\mathcal{L}_{Con}$ and $\mathcal{L}_{SARE}$ separately and train the ViT-Convpass model. We can see from Table~\ref{tab:ablation} that using the $\mathcal{L}_{Con}$ and $\mathcal{L}_{SARE}$ can separately benefit the model's generalization capability, and combine them together can bring better performance. 
In Fig.~\ref{fig:parameter analysis}, we study the AUC performance of different $\alpha$ and $\beta$. In two experiments, the curves show a similar pattern, indicating that $\alpha=0.02$ and $\beta=10$ are recommended for different experiments.

\begin{figure}[tbp]
     \centering
     \includegraphics[width=0.45\textwidth]{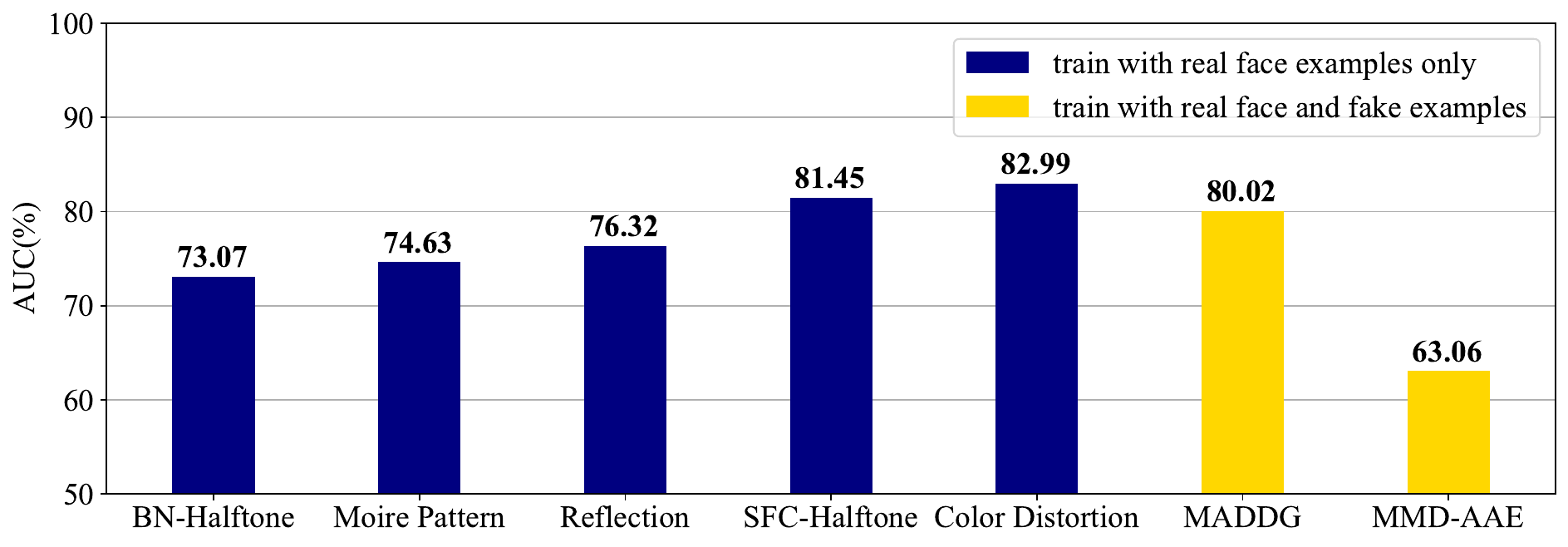}
     \caption{
     Effectiveness of using our recapturing simulation for synthesizing \textbf{spoof} examples on the `I\&C\&M to O'. Blue bars are the results of training with real-face examples \textbf{only} and the \textit{spoof} examples are synthesized by our recapturing simulation.  
     MMD-AAE \citep{MMDAAE-CVPR-2018} and MADDG \citep{MADDG-CVPR-2019} are trained with real and fake examples (yellow bars).
     }   \label{fig:Aug_Effectiveness} 
\end{figure}

\noindent \textbf{Effectiveness of recapturing simulation.} 
\textcolor{black}{In the above experiments, our FAS-Aug is operated on both real-face and fake-face examples. 
To validate using our recapturing simulation to synthesize spoof examples, we use only real face examples to conduct an experiment with `ViT-Convpass\&FA'. The experimental results are presented in Fig.\ref{fig:Aug_Effectiveness}. In each experiment, only the real face examples are used for augmentation and each time one type of recapturing operation is used. For example, in the experiment of BN-Halftone, only the operation of `BNHalftone' is used with random magnitude to synthesize spoofing attack samples from the real face examples for model training. }
From Fig.~\ref{fig:Aug_Effectiveness}, we can see that even if we only use real-face examples and the synthesized spoofing attack examples by our FAS-Aug, the AUC performance can be significantly over 50\%, indicating the effectiveness of our augmentation of recaptured examples.
Moreover, the AUC performance of each recapturing simulation can be comparable to or even better than existing methods MADDG\citep{MADDG-CVPR-2019} and MMD-AAE\citep{MMDAAE-CVPR-2018} which use both real and spoofing face examples in training. 
Therefore, using our recapturing simulation for spoof data synthesis is valid, and can benefit some scenarios when only real-face examples are available \citep{li2022one}.

\begin{table*}[tbp]
  \centering
 
  \caption{Experimental on the leave-one-out benchmark MICY. Results are in terms of HTER (\%) and AUC (\%).}
    \resizebox{0.95\textwidth}{!}{
    \begin{tabular}{l|c|c|c|c|c|c|c|c}
    \hline
    \multicolumn{1}{c|}{\multirow{2}[4]{*}{Method}} & \multicolumn{2}{c|}{M\&C\&Y to I} & \multicolumn{2}{c|}{I\&C\&Y to M} & \multicolumn{2}{c|}{I\&M\&Y to C} & \multicolumn{2}{c}{I\&C\&M to Y} \bigstrut\\
\cline{2-9}      & HTER (\%) & AUC (\%) & HTER (\%) & AUC (\%) & HTER (\%) & AUC (\%) & HTER (\%) & AUC (\%) \bigstrut\\
    \hline
    HFN+MP \citep{metapattern} & 10.42 & 95.58 & \textbf{7.31} & 96.79 & 9.44 & 96.05 & 17.24 & 89.76\bigstrut\\
    \hline
    \hline
    ViT-Convpass\&FA\&CS(ours) & \textbf{8.21} & \textbf{97.04} & 7.41 &\textbf{ 97.51} & \textbf{7.39} & \textbf{97.65} & \textbf{9.93} & \textbf{96.21} \bigstrut\\
    \hline
    \end{tabular}%
    }
  \label{tab:sota-micy}%
\end{table*}%

\begin{table}[htbp]
  \centering
  \caption{Comparing the result of using both FAS-Aug and TI-Aug (\&FA\&TA) with only using TI-Aug (\&TA) or FAS-Aug (\&FA).}
    \resizebox{0.9\linewidth}{!}{
    \begin{tabular}{l|c|c|c|c}
    \hline
    \multicolumn{1}{c|}{\multirow{2}[4]{*}{Method}} & \multicolumn{2}{c}{O\&M\&I to C} & \multicolumn{2}{c}{I\&C\&M to O} \\
\cline{2-5}      & HTER(\%) & AUC(\%) & HTER(\%) & AUC(\%) \\
    \hline
    ViT-ConvPass\&TA & 7.61 & 97.20 & 10.74 & 95.82 \\
    \hline
    ViT-ConvPass\&FA & 7.89 & 96.54 & 8.31 & 97.41 \\
    \hline
    ViT-ConvPass\&FA\&TA & 7.11 & 97.59 & 6.71 & 98.21 \\
    \hline
    \end{tabular}%
    }
  \label{tab:mix-aug}%
\end{table}%
\begin{table}[tbp]
  \centering
  \caption{Cross-dataset testing between the 3D Mask datasets}
  \resizebox{0.9\linewidth}{!}{
    \begin{tabular}{l|c|c|c|c}
    \hline
    \multicolumn{1}{c|}{\multirow{2}[4]{*}{Method}} & \multicolumn{2}{c|}{CeFA to CASIA-SURF} & \multicolumn{2}{c}{Hifi Mask to HKBU v2} \\
\cline{2-5}          & HTER(\%) & AUC(\%) & HTER(\%) & AUC(\%) \\
    \hline
    ViT-ConvPass  & 43.73 & 55.39 & 4.25  & 99.25 \\
    \hline
    ViT-ConvPass\&FA & 42.60 & 59.74 & 3.92  & 99.46 \\
    \hline
    \end{tabular}%
    }
    \vspace{-1em}
  \label{tab}%
\end{table}%

\subsubsection{Experiment of combining FAS-Aug and TI-Aug}\label{G-FASandTI}

We also conduct the experiment using the FAS-Aug and TI-Aug together. As shown in Table \ref{tab:mix-aug}, the HTER and AUC results of `ViT-ConvPass\&FA\&TA’ are better than that of `ViT-ConvPass\&TA' and `ViT-ConvPass\&FA', which means that our FAS-Aug is not exclusive and can also be used with existing data augmentations for face anti-spoofing.

\subsubsection{Using FAS-Aug with 3D Mask datasets}\label{3D-Mask}
We also consider challenges from 3D Mask attacks. We verify the effectiveness of our FAS-Aug, we conduct cross-testing experiments by using the CASIA-SURF \citep{CASIA-SURF}, CeFA\citep{cefa}, HiFi Mask \citep{hifi} and HKBU Marvs V2 \citep{hkbuv2} datasets. From the results in Table~\ref{tab}, we can see that our FAS-Aug can also be useful when working with 3D Mask datasets as our FAS-Aug also increases data diversity by simulating the capturing process.

\subsubsection{Comparison with state-of-the-art methods}

\noindent \textbf{Leave-one-out protocol MICO.}
In Table~\ref{tab:sota-mico}, we compare our method `ViT-Convpass\&FA\&CS' with latest \wh{state-of-the-arts} \citep{NASFAS-TPAMI-2020, SSDG-CVPR-2020, metapattern, wang2022patchnet,zhou2022adaptive,MADDG-CVPR-2019,RFMetaFAS-AAAI-2020}, in the leave-one-out domain generalization protocol MICO \citep{MADDG-CVPR-2019}. In `O\&C\&M to I', our method achieves comparable HTER to the lowest HTER of SSAN \citep{SSAN}. Furthermore, on the other comparisons, our method achieves the best HTER and AUC performance and \wh{provides} new state-of-the-art results.

\noindent \textbf{Leave-one-out protocol MICY.}  
The MICY protocol is first \wh{introduced} by Wang \textit{et al.} \citep{wang2020unsupervised} to study multi-source unsupervised domain adaptation FAS. Then, Cai \textit{et al.} \citep{metapattern} extend the MICY for domain generalization FAS. To make \wh{a} more extensive comparison and show the effectiveness of our method, we also conduct \wh{experiments} on MICY. As shown in Table~\ref{tab:sota-micy}. in `I\&C\&M to Y', our method \wh{achieves} much lower HTER and higher AUC than a recent work `HFN+MP' \citep{metapattern} by a large margin. In the other experiments, our method also \wh{surpasses} `HFN+MP' \citep{metapattern} in terms of AUC.
\begin{table}[tbp]
  \centering
  \centering   
  \caption{Experimental results with limited source domains. }
  \resizebox{0.45\textwidth}{!}{
    \begin{tabular}{l|c|c|c|c}
    \hline
    \multirow{2}[4]{*}{Method} & \multicolumn{2}{c|}{M\&I to C} & \multicolumn{2}{c}{M\&I to O} \bigstrut\\
\cline{2-5}      & HTER(\%) & AUC(\%) & HTER(\%) & AUC(\%) \bigstrut\\
\hline
ColorTexture \citep{FAS-ColorTexture-TIFS-2016} & 55.17 & 46.89 &  53.31 & 45.16 \\\hline
LBP-TOP \citep{LBP-TOP-EJIVP-2014} & 45.27 &  54.88 & 47.26 & 50.21 \\\hline
MADDG \citep{MADDG-CVPR-2019} & 41.02 & 64.33 & 39.35 & 65.10  \\
\hline
SSDG-M \citep{SSDG-CVPR-2020} & 31.89 & 71.29 & 36.01 & 66.88  \\
\hline
AMEL \citep{zhou2022adaptive} & 23.33& 85.17 &	19.68
& 87.01 \\
   \hline
HFN+MP \citep{metapattern} & 30.89 & 72.48 &	20.94
& 86.71 \\
    \hline
   \hline
ViT-Convpass\&FA\&CS(ours)& \textbf{16.89} & \textbf{90.06} & \textbf{15.10} & \textbf{92.69} \bigstrut\\
    \hline
    \end{tabular}%
    }
  \label{tab:sota-MI-CO}%
\end{table}%
\noindent \textbf{Limited source domains.}
We also validate our method in the situation when two source domains are available. Table~\ref{tab:sota-MI-CO} shows the results of using \textit{M} and \textit{I} for training. The results of our `ViT-Convpass\&FA\&CS' also significantly surpass recent state-of-the-art methods\citep{metapattern, zhou2022adaptive} by a large margin in terms of HTER ($>6$ \%) and AUC ($>4$\%). Our method can augment more data, and thus its effectiveness is still valid when the available source domains are limited.

\section{Conclusion and future work}
In this work, we fill in the gaps that there is no FAS-specific data augmentation and propose a bag of FAS augmentations based on physical capturing and recapturing processes. Our proposed FAS-Aug can generally outperform traditional image augmentations with the synthesized spoofing artifacts. Furthermore, we propose Spoofing Attack Risk Equalization to prevent models from relying on specific types of artifacts to learn more generalized FAS models. We validate our FAS-Aug and SARE on the cross-domain FAS protocols and our method can provide new state-of-the-art performance. Moreover, we validate the effectiveness in the scenarios, where our FAS-Aug is used with other methods, where there are more real-face examples than spoof ones, and where there are only real-face examples. Therefore, our FAS-Aug can benefit the entire FAS community.

\textcolor{black}{In the future, we could study how to utilize our FAS-Aug in developing advanced self-supervised learning or contrastive learning algorithms for FAS. Moreover, beyond FAS, our work may also be used for other security-related research, such as recaptured document detection, which is similar to face presentation attack detection. Besides, some researchers may be interested in utilizing our FAS-Aug when developing robust adversarial noise against printing in the physical world.}

\section{Data availability statement}
The ten datasets used in this paper are publically available (\textit{i.e.} SiW \citep{FAS-Auxiliary-CVPR-2018}, ROSE-YOUTU\citep{FAS-UnsupervisedDA-TIFS-2018}, OULU-NPU \citep{OULU_NPU_2017}, CASIA FASD\citep{DB-CASIAFASD}, IDIAP Replay-Attack \citep{LBP-FAS-BIOSIG-2012}, MSU MFSD \citep{FAS-IDA-TIFS-2015}, CASIA-SURF CeFA \citep{cefa}, CASIA-SURF \citep{CASIA-SURF}, CASIA-SURF HiFi Mask \citep{hifi}, HKBU Marvs V2 \citep{hkbuv2}). These datasets can be found in third-party institutions, including Idiap Research Institute, Hong Kong Baptist University, Institute of Automation of Chinese Academy of Sciences, Tencent Youtu Research, and Michigan State University.
\newpage
\section*{Acknowledgement}
This work was carried out at Rapid-Rich Object Search (ROSE) Lab, School of Electrical \& Electronic Engineering, Nanyang Technological University. This research is supported by the NTU-PKU Joint Research Institute (a collaboration between the Nanyang Technological University and Peking University that is sponsored by a donation from the Ng Teng Fong Charitable Foundation). This work is also partly by the Basic and Frontier Research Project of PCL, the Major Key Project of PCL, partly by Guangdong Basic and Applied Basic Research Foundation (Grant No. 2024A1515010454\&2023A1515140037), National Natural Science Foundation of China under Grant 62306061, and Chow Sang Sang Group Research Fund under grant DON-RMG 9229161.
{\small
\bibliography{egbib}

\begin{thebibliography}{}
\providecommand{\doi}[1]{\url{https://doi.org/#1}}
\bibcommenthead

\bibitem [\protect \citeauthoryear {%
{Boulkenafet}%
, {Komulainen}%
\BCBL {}\ \BBA {} {Hadid}%
}{%
{Boulkenafet}%
\ \protect \BOthers {.}}{%
{\protect \APACyear {2016}}%
}]{%
FAS-ColorTexture-TIFS-2016}
\APACinsertmetastar {%
FAS-ColorTexture-TIFS-2016}%
\begin{APACrefauthors}%
{Boulkenafet}, Z.%
, {Komulainen}, J.%
\BCBL {} {Hadid}, A.%
\end{APACrefauthors}%
\unskip\
\newblock
\APACrefYearMonthDay{2016}{Aug}{}.
\newblock
{\BBOQ}\APACrefatitle {{Face Spoofing Detection Using Colour Texture Analysis}} {{Face Spoofing Detection Using Colour Texture Analysis}}.{\BBCQ}
\newblock
\APACjournalVolNumPages{IEEE Transactions on Information Forensics and Security}{11}{8}{1818-1830}.
\newblock

\newblock

\newblock
\begin{APACrefDOI} \doi{10.1109/TIFS.2016.2555286} \end{APACrefDOI}
\PrintBackRefs{\CurrentBib}

\bibitem [\protect \citeauthoryear {%
Boulkenafet%
, Komulainen%
, Li%
, Feng%
\BCBL {}\ \BBA {} Hadid%
}{%
Boulkenafet%
\ \protect \BOthers {.}}{%
{\protect \APACyear {2017}}%
}]{%
OULU_NPU_2017}
\APACinsertmetastar {%
OULU_NPU_2017}%
\begin{APACrefauthors}%
Boulkenafet, Z.%
, Komulainen, J.%
, Li, L.%
, Feng, X.%
\BCBL {} Hadid, A.%
\end{APACrefauthors}%
\unskip\
\newblock
\APACrefYearMonthDay{2017}{{\APACmonth{05}}}{}.
\newblock
{\BBOQ}\APACrefatitle {{OULU-NPU}: A mobile face presentation attack database with real-world variations} {{OULU-NPU}: A mobile face presentation attack database with real-world variations}.{\BBCQ}
\newblock
 \APACrefbtitle {{IEEE International Conference on Automatic Face and Gesture Recognition}.} {{IEEE International Conference on Automatic Face and Gesture Recognition}.}
\PrintBackRefs{\CurrentBib}

\bibitem [\protect \citeauthoryear {%
Cai%
\ \BBA {} Chen%
}{%
Cai%
\ \BBA {} Chen%
}{%
{\protect \APACyear {2019}}%
}]{%
deep-forest-lbp}
\APACinsertmetastar {%
deep-forest-lbp}%
\begin{APACrefauthors}%
Cai, R.%
\BCBT {}\ \BBA {} Chen, C.%
\end{APACrefauthors}%
\unskip\
\newblock
\APACrefYearMonthDay{2019}{}{}.
\newblock
{\BBOQ}\APACrefatitle {Learning deep forest with multi-scale Local Binary Pattern features for face anti-spoofing} {Learning deep forest with multi-scale local binary pattern features for face anti-spoofing}.{\BBCQ}
\newblock
\APACjournalVolNumPages{arXiv preprint arXiv:1910.03850}{}{}{}.
\newblock

\newblock

\PrintBackRefs{\CurrentBib}

\bibitem [\protect \citeauthoryear {%
Cai%
\ \protect \BOthers {.}}{%
Cai%
\ \protect \BOthers {.}}{%
{\protect \APACyear {2023}}%
}]{%
Cai_2023_ICCV}
\APACinsertmetastar {%
Cai_2023_ICCV}%
\begin{APACrefauthors}%
Cai, R.%
, Cui, Y.%
, Li, Z.%
, Yu, Z.%
, Li, H.%
, Hu, Y.%
\BCBL {} Kot, A.%
\end{APACrefauthors}%
\unskip\
\newblock
\APACrefYearMonthDay{2023}{October}{}.
\newblock
{\BBOQ}\APACrefatitle {{Rehearsal-Free Domain Continual Face Anti-Spoofing: Generalize More and Forget Less}} {{Rehearsal-Free Domain Continual Face Anti-Spoofing: Generalize More and Forget Less}}.{\BBCQ}
\newblock
 \APACrefbtitle {{Proceedings of the IEEE/CVF International Conference on Computer Vision (ICCV)}} {{Proceedings of the IEEE/CVF International Conference on Computer Vision (ICCV)}}\ (\BPG~8037-8048).
\PrintBackRefs{\CurrentBib}

\bibitem [\protect \citeauthoryear {%
Cai%
, Li%
, Wang%
, Chen%
\BCBL {}\ \BBA {} Kot%
}{%
Cai%
\ \protect \BOthers {.}}{%
{\protect \APACyear {2020}}%
}]{%
CAI-2020-DRL}
\APACinsertmetastar {%
CAI-2020-DRL}%
\begin{APACrefauthors}%
Cai, R.%
, Li, H.%
, Wang, S.%
, Chen, C.%
\BCBL {} Kot, A.C.%
\end{APACrefauthors}%
\unskip\
\newblock
\APACrefYearMonthDay{2020}{}{}.
\newblock
{\BBOQ}\APACrefatitle {{DRL-FAS: A Novel Framework Based on Deep Reinforcement Learning for Face Anti-Spoofing}} {{DRL-FAS: A Novel Framework Based on Deep Reinforcement Learning for Face Anti-Spoofing}}.{\BBCQ}
\newblock
\APACjournalVolNumPages{IEEE Transactions on Information Forensics and Security}{16}{}{937--951}.
\newblock

\newblock

\PrintBackRefs{\CurrentBib}

\bibitem [\protect \citeauthoryear {%
Cai%
\ \protect \BOthers {.}}{%
Cai%
\ \protect \BOthers {.}}{%
{\protect \APACyear {2022}}%
}]{%
metapattern}
\APACinsertmetastar {%
metapattern}%
\begin{APACrefauthors}%
Cai, R.%
, Li, Z.%
, Wan, R.%
, Li, H.%
, Hu, Y.%
\BCBL {} Kot, A.C.%
\end{APACrefauthors}%
\unskip\
\newblock
\APACrefYearMonthDay{2022}{}{}.
\newblock
{\BBOQ}\APACrefatitle {{Learning Meta Pattern for Face Anti-Spoofing}} {{Learning Meta Pattern for Face Anti-Spoofing}}.{\BBCQ}
\newblock
\APACjournalVolNumPages{IEEE Transactions on Information Forensics and Security}{17}{}{1201-1213}.
\newblock

\newblock

\newblock
\begin{APACrefDOI} \doi{10.1109/TIFS.2022.3158551} \end{APACrefDOI}
\PrintBackRefs{\CurrentBib}

\bibitem [\protect \citeauthoryear {%
Cai%
, Song%
\BCBL {}\ \protect \BOthers {.}}{%
Cai%
, Song%
\BCBL {}\ \protect \BOthers {.}}{%
{\protect \APACyear {2024}}%
}]{%
cai2023benchlmm}
\APACinsertmetastar {%
cai2023benchlmm}%
\begin{APACrefauthors}%
Cai, R.%
, Song, Z.%
, Guan, D.%
, Chen, Z.%
, Luo, X.%
, Yi, C.%
\BCBL {} Kot, A.%
\end{APACrefauthors}%
\unskip\
\newblock
\APACrefYearMonthDay{2024}{}{}.
\newblock
{\BBOQ}\APACrefatitle {Benchlmm: Benchmarking cross-style visual capability of large multimodal models} {Benchlmm: Benchmarking cross-style visual capability of large multimodal models}.{\BBCQ}
\newblock
 \APACrefbtitle {{European Conference on Computer Vision (ECCV)}.} {{European Conference on Computer Vision (ECCV)}.}
\PrintBackRefs{\CurrentBib}

\bibitem [\protect \citeauthoryear {%
Cai%
, Yu%
\BCBL {}\ \protect \BOthers {.}}{%
Cai%
, Yu%
\BCBL {}\ \protect \BOthers {.}}{%
{\protect \APACyear {2024}}%
}]{%
s-adapter}
\APACinsertmetastar {%
s-adapter}%
\begin{APACrefauthors}%
Cai, R.%
, Yu, Z.%
, Kong, C.%
, Li, H.%
, Chen, C.%
, Hu, Y.H.%
\BCBL {} Kot, A.%
\end{APACrefauthors}%
\unskip\
\newblock
\APACrefYearMonthDay{2024}{}{}.
\newblock
{\BBOQ}\APACrefatitle {S-Adapter: Generalizing Vision Transformer for Face Anti-Spoofing with Statistical Tokens} {S-adapter: Generalizing vision transformer for face anti-spoofing with statistical tokens}.{\BBCQ}
\newblock
\APACjournalVolNumPages{IEEE Transactions on Information Forensics Security}{}{}{}.
\newblock

\newblock

\PrintBackRefs{\CurrentBib}

\bibitem [\protect \citeauthoryear {%
Chae%
, Yoo%
, Sun%
, Kang%
\BCBL {}\ \BBA {} Ko%
}{%
Chae%
\ \protect \BOthers {.}}{%
{\protect \APACyear {2017}}%
}]{%
chae2017subpixel}
\APACinsertmetastar {%
chae2017subpixel}%
\begin{APACrefauthors}%
Chae, S\BHBI H.%
, Yoo, C\BHBI H.%
, Sun, J\BHBI Y.%
, Kang, M\BHBI C.%
\BCBL {} Ko, S\BHBI J.%
\end{APACrefauthors}%
\unskip\
\newblock
\APACrefYearMonthDay{2017}{}{}.
\newblock
{\BBOQ}\APACrefatitle {Subpixel rendering for the pentile display based on the human visual system} {Subpixel rendering for the pentile display based on the human visual system}.{\BBCQ}
\newblock
\APACjournalVolNumPages{IEEE Transactions on Consumer Electronics}{63}{4}{401--409}.
\newblock

\newblock

\PrintBackRefs{\CurrentBib}

\bibitem [\protect \citeauthoryear {%
Chen%
, Li%
, Cai%
, Zeng%
\BCBL {}\ \BBA {} Huang%
}{%
Chen%
\ \protect \BOthers {.}}{%
{\protect \APACyear {2023}}%
}]{%
chen2023distortion}
\APACinsertmetastar {%
chen2023distortion}%
\begin{APACrefauthors}%
Chen, C.%
, Li, B.%
, Cai, R.%
, Zeng, J.%
\BCBL {} Huang, J.%
\end{APACrefauthors}%
\unskip\
\newblock
\APACrefYearMonthDay{2023}{}{}.
\newblock
{\BBOQ}\APACrefatitle {Distortion model-based spectral augmentation for generalized recaptured document detection} {Distortion model-based spectral augmentation for generalized recaptured document detection}.{\BBCQ}
\newblock
\APACjournalVolNumPages{IEEE Transactions on Information Forensics and Security}{19}{}{1283--1298}.
\newblock

\newblock

\PrintBackRefs{\CurrentBib}

\bibitem [\protect \citeauthoryear {%
{Chingovska}%
, {Anjos}%
\BCBL {}\ \BBA {} {Marcel}%
}{%
{Chingovska}%
\ \protect \BOthers {.}}{%
{\protect \APACyear {2012}}%
}]{%
LBP-FAS-BIOSIG-2012}
\APACinsertmetastar {%
LBP-FAS-BIOSIG-2012}%
\begin{APACrefauthors}%
{Chingovska}, I.%
, {Anjos}, A.%
\BCBL {} {Marcel}, S.%
\end{APACrefauthors}%
\unskip\
\newblock
\APACrefYearMonthDay{2012}{Sep.}{}.
\newblock
{\BBOQ}\APACrefatitle {On the effectiveness of local binary patterns in face anti-spoofing} {On the effectiveness of local binary patterns in face anti-spoofing}.{\BBCQ}
\newblock
 \APACrefbtitle {{2012 BIOSIG - Proceedings of the International Conference of Biometrics Special Interest Group (BIOSIG)}} {{2012 BIOSIG - Proceedings of the International Conference of Biometrics Special Interest Group (BIOSIG)}}\ (\BPG~1-7).
\PrintBackRefs{\CurrentBib}

\bibitem [\protect \citeauthoryear {%
ChromaSoft%
}{%
ChromaSoft%
}{%
{\protect \APACyear {{\protect \bibnodate {}}}}%
}]{%
profiles_c}
\APACinsertmetastar {%
profiles_c}%
\begin{APACrefauthors}%
ChromaSoft%
\end{APACrefauthors}%
\unskip\
\newblock
\APACrefYearMonthDay{{\protect \bibnodate {}}}{}{}.
\newblock
\APACrefbtitle {{ICM Profiles}.} {{ICM Profiles}.}
\newblock
\APAChowpublished {\url{https://sites.google.com/site/chromasoft/icmprofiles}}.
\PrintBackRefs{\CurrentBib}

\bibitem [\protect \citeauthoryear {%
Consortium%
\ \protect \BOthers {.}}{%
Consortium%
\ \protect \BOthers {.}}{%
{\protect \APACyear {2004}}%
}]{%
international2004role}
\APACinsertmetastar {%
international2004role}%
\begin{APACrefauthors}%
Consortium, I.C.%
\BCBT {}\ \BOthersPeriod {.}
\end{APACrefauthors}%
\unskip\
\newblock
\APACrefYearMonthDay{2004}{}{}.
\newblock
{\BBOQ}\APACrefatitle {The role of ICC profiles in a colour reproduction system} {The role of icc profiles in a colour reproduction system}.{\BBCQ}
\newblock
\APACjournalVolNumPages{ICC White Paper: http://www. color. org/ICC\_white\_paper\_7\_role\_of\_ICC\_profiles. pdf}{}{}{}.
\newblock

\newblock

\PrintBackRefs{\CurrentBib}

\bibitem [\protect \citeauthoryear {%
Cubuk%
, Zoph%
, Mane%
, Vasudevan%
\BCBL {}\ \BBA {} Le%
}{%
Cubuk%
\ \protect \BOthers {.}}{%
{\protect \APACyear {2019}}%
}]{%
cubuk2019autoaugment}
\APACinsertmetastar {%
cubuk2019autoaugment}%
\begin{APACrefauthors}%
Cubuk, E.D.%
, Zoph, B.%
, Mane, D.%
, Vasudevan, V.%
\BCBL {} Le, Q.V.%
\end{APACrefauthors}%
\unskip\
\newblock
\APACrefYearMonthDay{2019}{}{}.
\newblock
{\BBOQ}\APACrefatitle {{AutoAugment: Learning augmentation strategies from data}} {{AutoAugment: Learning augmentation strategies from data}}.{\BBCQ}
\newblock
 \APACrefbtitle {{Proceedings of the IEEE/CVF Conference on Computer Vision and Pattern Recognition}} {{Proceedings of the IEEE/CVF Conference on Computer Vision and Pattern Recognition}}\ (\BPGS\ 113--123).
\PrintBackRefs{\CurrentBib}

\bibitem [\protect \citeauthoryear {%
Cubuk%
, Zoph%
, Shlens%
\BCBL {}\ \BBA {} Le%
}{%
Cubuk%
\ \protect \BOthers {.}}{%
{\protect \APACyear {2020}}%
}]{%
cubuk2020randaugment}
\APACinsertmetastar {%
cubuk2020randaugment}%
\begin{APACrefauthors}%
Cubuk, E.D.%
, Zoph, B.%
, Shlens, J.%
\BCBL {} Le, Q.V.%
\end{APACrefauthors}%
\unskip\
\newblock
\APACrefYearMonthDay{2020}{}{}.
\newblock
{\BBOQ}\APACrefatitle {RandAugment: Practical automated data augmentation with a reduced search space} {Randaugment: Practical automated data augmentation with a reduced search space}.{\BBCQ}
\newblock
 \APACrefbtitle {Proceedings of the IEEE/CVF Conference on Computer Vision and Pattern Recognition Workshops} {Proceedings of the ieee/cvf conference on computer vision and pattern recognition workshops}\ (\BPGS\ 702--703).
\PrintBackRefs{\CurrentBib}

\bibitem [\protect \citeauthoryear {%
de Freitas~Pereira%
\ \protect \BOthers {.}}{%
de Freitas~Pereira%
\ \protect \BOthers {.}}{%
{\protect \APACyear {2014}}%
}]{%
LBP-TOP-EJIVP-2014}
\APACinsertmetastar {%
LBP-TOP-EJIVP-2014}%
\begin{APACrefauthors}%
de Freitas~Pereira, T.%
, Komulainen, J.%
, Anjos, A.%
, De~Martino, J.M.%
, Hadid, A.%
, Pietik{\"a}inen, M.%
\BCBL {} Marcel, S.%
\end{APACrefauthors}%
\unskip\
\newblock
\APACrefYearMonthDay{2014}{}{}.
\newblock
{\BBOQ}\APACrefatitle {Face liveness detection using dynamic texture} {Face liveness detection using dynamic texture}.{\BBCQ}
\newblock
\APACjournalVolNumPages{EURASIP Journal on Image and Video Processing}{2014}{1}{2}.
\newblock

\newblock

\PrintBackRefs{\CurrentBib}

\bibitem [\protect \citeauthoryear {%
Dosovitskiy%
\ \protect \BOthers {.}}{%
Dosovitskiy%
\ \protect \BOthers {.}}{%
{\protect \APACyear {2020}}%
}]{%
ViT}
\APACinsertmetastar {%
ViT}%
\begin{APACrefauthors}%
Dosovitskiy, A.%
, Beyer, L.%
, Kolesnikov, A.%
, Weissenborn, D.%
, Zhai, X.%
, Unterthiner, T.%
\BDBL {}others%
\end{APACrefauthors}%
\unskip\
\newblock
\APACrefYearMonthDay{2020}{}{}.
\newblock
{\BBOQ}\APACrefatitle {An image is worth 16x16 words: Transformers for image recognition at scale} {An image is worth 16x16 words: Transformers for image recognition at scale}.{\BBCQ}
\newblock
 \APACrefbtitle {{International Conference on Learning Representations (ICLR)}.} {{International Conference on Learning Representations (ICLR)}.}
\PrintBackRefs{\CurrentBib}

\bibitem [\protect \citeauthoryear {%
Eck%
}{%
Eck%
}{%
{\protect \APACyear {2021}}%
}]{%
eck2021introduction}
\APACinsertmetastar {%
eck2021introduction}%
\begin{APACrefauthors}%
Eck, D.J.%
\end{APACrefauthors}%
\unskip\
\newblock
\APACrefYearMonthDay{2021}{}{}.
\newblock
{\BBOQ}\APACrefatitle {Introduction to Computer Graphics} {Introduction to computer graphics}.{\BBCQ}
\newblock
\BIn{} (\BPGS\ 212--215).
\newblock
\APACaddressPublisher{}{Hobart and William Smith College}.
\PrintBackRefs{\CurrentBib}

\bibitem [\protect \citeauthoryear {%
Elliott%
\ \protect \BOthers {.}}{%
Elliott%
\ \protect \BOthers {.}}{%
{\protect \APACyear {2002}}%
}]{%
elliott200213}
\APACinsertmetastar {%
elliott200213}%
\begin{APACrefauthors}%
Elliott, C.H.B.%
, Han, S.%
, Im, M.H.%
, Higgins, M.%
, Higgins, P.%
, Hong, M.%
\BDBL {}Chung, K.%
\end{APACrefauthors}%
\unskip\
\newblock
\APACrefYearMonthDay{2002}{}{}.
\newblock
{\BBOQ}\APACrefatitle {13.3 Co-Optimization of Color AMLCD Subpixel Architecture and Rendering Algorithms} {13.3 co-optimization of color amlcd subpixel architecture and rendering algorithms}.{\BBCQ}
\newblock
 \APACrefbtitle {{SID Symposium Digest of Technical Papers}} {{SID Symposium Digest of Technical Papers}}\ (\BVOL~33, \BPGS\ 172--175).
\PrintBackRefs{\CurrentBib}

\bibitem [\protect \citeauthoryear {%
Fang%
, Au%
\BCBL {}\ \BBA {} Cheung%
}{%
Fang%
\ \protect \BOthers {.}}{%
{\protect \APACyear {2013}}%
}]{%
fang2013subpixel}
\APACinsertmetastar {%
fang2013subpixel}%
\begin{APACrefauthors}%
Fang, L.%
, Au, O.C.%
\BCBL {} Cheung, N\BHBI M.%
\end{APACrefauthors}%
\unskip\
\newblock
\APACrefYearMonthDay{2013}{}{}.
\newblock
{\BBOQ}\APACrefatitle {Subpixel rendering: from font rendering to image subsampling [Applications Corner]} {Subpixel rendering: from font rendering to image subsampling [applications corner]}.{\BBCQ}
\newblock
\APACjournalVolNumPages{IEEE Signal Processing Magazine}{30}{3}{177--189}.
\newblock

\newblock

\PrintBackRefs{\CurrentBib}

\bibitem [\protect \citeauthoryear {%
Ford%
\ \BBA {} Roberts%
}{%
Ford%
\ \BBA {} Roberts%
}{%
{\protect \APACyear {1998}}%
}]{%
cmyk_conversion1998colour}
\APACinsertmetastar {%
cmyk_conversion1998colour}%
\begin{APACrefauthors}%
Ford, A.%
\BCBT {}\ \BBA {} Roberts, A.%
\end{APACrefauthors}%
\unskip\
\newblock
\APACrefYearMonthDay{1998}{}{}.
\newblock
{\BBOQ}\APACrefatitle {Colour space conversions} {Colour space conversions}.{\BBCQ}
\newblock
\APACjournalVolNumPages{Westminster University, London}{1998}{}{1--31}.
\newblock

\newblock

\PrintBackRefs{\CurrentBib}

\bibitem [\protect \citeauthoryear {%
{Galbally}%
\ \BBA {} {Marcel}%
}{%
{Galbally}%
\ \BBA {} {Marcel}%
}{%
{\protect \APACyear {2014}}%
}]{%
IQA-ICPR-2014}
\APACinsertmetastar {%
IQA-ICPR-2014}%
\begin{APACrefauthors}%
{Galbally}, J.%
\BCBT {}\ \BBA {} {Marcel}, S.%
\end{APACrefauthors}%
\unskip\
\newblock
\APACrefYearMonthDay{2014}{Aug}{}.
\newblock
{\BBOQ}\APACrefatitle {{Face Anti-spoofing Based on General Image Quality Assessment}} {{Face Anti-spoofing Based on General Image Quality Assessment}}.{\BBCQ}
\newblock
 \APACrefbtitle {{2014 22nd International Conference on Pattern Recognition}} {{2014 22nd International Conference on Pattern Recognition}}\ (\BPG~1173-1178).
\newblock
\begin{APACrefDOI} \doi{10.1109/ICPR.2014.211} \end{APACrefDOI}
\PrintBackRefs{\CurrentBib}

\bibitem [\protect \citeauthoryear {%
Garcia%
\ \BBA {} de Queiroz%
}{%
Garcia%
\ \BBA {} de Queiroz%
}{%
{\protect \APACyear {2015}}%
{\protect \APACexlab {{\protect \BCnt {1}}}}}]{%
garcia2015face}
\APACinsertmetastar {%
garcia2015face}%
\begin{APACrefauthors}%
Garcia, D.C.%
\BCBT {}\ \BBA {} de Queiroz, R.L.%
\end{APACrefauthors}%
\unskip\
\newblock
\APACrefYearMonthDay{2015{\protect \BCnt {1}}}{}{}.
\newblock
{\BBOQ}\APACrefatitle {Face-spoofing 2D-detection based on Moir{\'e}-pattern analysis} {Face-spoofing 2d-detection based on moir{\'e}-pattern analysis}.{\BBCQ}
\newblock
\APACjournalVolNumPages{IEEE Transactions on Information Forensics and Security}{10}{4}{778--786}.
\newblock

\newblock

\PrintBackRefs{\CurrentBib}

\bibitem [\protect \citeauthoryear {%
Garcia%
\ \BBA {} de Queiroz%
}{%
Garcia%
\ \BBA {} de Queiroz%
}{%
{\protect \APACyear {2015}}%
{\protect \APACexlab {{\protect \BCnt {2}}}}}]{%
moire-analysis-TIFS-2015}
\APACinsertmetastar {%
moire-analysis-TIFS-2015}%
\begin{APACrefauthors}%
Garcia, D.C.%
\BCBT {}\ \BBA {} de Queiroz, R.L.%
\end{APACrefauthors}%
\unskip\
\newblock
\APACrefYearMonthDay{2015{\protect \BCnt {2}}}{}{}.
\newblock
{\BBOQ}\APACrefatitle {{Face-Spoofing 2D-Detection Based on Moiré-Pattern Analysis}} {{Face-Spoofing 2D-Detection Based on Moiré-Pattern Analysis}}.{\BBCQ}
\newblock
\APACjournalVolNumPages{IEEE Transactions on Information Forensics and Security}{10}{4}{778--786}.
\newblock

\newblock

\PrintBackRefs{\CurrentBib}

\bibitem [\protect \citeauthoryear {%
Green%
}{%
Green%
}{%
{\protect \APACyear {2013}}%
}]{%
green2013gamut}
\APACinsertmetastar {%
green2013gamut}%
\begin{APACrefauthors}%
Green, P.%
\end{APACrefauthors}%
\unskip\
\newblock
\APACrefYearMonthDay{2013}{}{}.
\newblock
{\BBOQ}\APACrefatitle {Gamut mapping for the perceptual reference medium gamut} {Gamut mapping for the perceptual reference medium gamut}.{\BBCQ}
\newblock
 \APACrefbtitle {{2013 Colour and Visual Computing Symposium (CVCS)}} {{2013 Colour and Visual Computing Symposium (CVCS)}}\ (\BPGS\ 1--6).
\PrintBackRefs{\CurrentBib}

\bibitem [\protect \citeauthoryear {%
He%
, Fan%
, Wu%
, Xie%
\BCBL {}\ \BBA {} Girshick%
}{%
He%
\ \protect \BOthers {.}}{%
{\protect \APACyear {2020}}%
}]{%
he2020momentum}
\APACinsertmetastar {%
he2020momentum}%
\begin{APACrefauthors}%
He, K.%
, Fan, H.%
, Wu, Y.%
, Xie, S.%
\BCBL {} Girshick, R.%
\end{APACrefauthors}%
\unskip\
\newblock
\APACrefYearMonthDay{2020}{}{}.
\newblock
{\BBOQ}\APACrefatitle {Momentum contrast for unsupervised visual representation learning} {Momentum contrast for unsupervised visual representation learning}.{\BBCQ}
\newblock
 \APACrefbtitle {{Proceedings of the IEEE/CVF Conference on Computer Vision and Pattern Recognition}} {{Proceedings of the IEEE/CVF Conference on Computer Vision and Pattern Recognition}}\ (\BPGS\ 9729--9738).
\PrintBackRefs{\CurrentBib}

\bibitem [\protect \citeauthoryear {%
Huang%
\ \protect \BOthers {.}}{%
Huang%
\ \protect \BOthers {.}}{%
{\protect \APACyear {2022}}%
}]{%
fas-adapter-eccv2022}
\APACinsertmetastar {%
fas-adapter-eccv2022}%
\begin{APACrefauthors}%
Huang, H\BHBI P.%
, Sun, D.%
, Liu, Y.%
, Chu, W\BHBI S.%
, Xiao, T.%
, Yuan, J.%
\BDBL {}Yang, M\BHBI H.%
\end{APACrefauthors}%
\unskip\
\newblock
\APACrefYearMonthDay{2022}{}{}.
\newblock
{\BBOQ}\APACrefatitle {Adaptive Transformers for Robust Few-shot Cross-domain Face Anti-spoofing} {Adaptive transformers for robust few-shot cross-domain face anti-spoofing}.{\BBCQ}
\newblock
 \APACrefbtitle {European Conference on Computer Vision.} {European conference on computer vision.}
\PrintBackRefs{\CurrentBib}

\bibitem [\protect \citeauthoryear {%
HutchColor%
}{%
HutchColor%
}{%
{\protect \APACyear {{\protect \bibnodate {}}}}%
}]{%
profiles_b}
\APACinsertmetastar {%
profiles_b}%
\begin{APACrefauthors}%
HutchColor, L.%
\end{APACrefauthors}%
\unskip\
\newblock
\APACrefYearMonthDay{{\protect \bibnodate {}}}{}{}.
\newblock
\APACrefbtitle {{RGB color space profiles}.} {{RGB color space profiles}.}
\newblock
\APAChowpublished {\url{http://www.hutchcolor.com/profiles.html}}.
\PrintBackRefs{\CurrentBib}

\bibitem [\protect \citeauthoryear {%
Incorporated%
}{%
Incorporated%
}{%
{\protect \APACyear {{\protect \bibnodate {}}}}%
}]{%
profile_a}
\APACinsertmetastar {%
profile_a}%
\begin{APACrefauthors}%
Incorporated, A.S.%
\end{APACrefauthors}%
\unskip\
\newblock
\APACrefYearMonthDay{{\protect \bibnodate {}}}{}{}.
\newblock
\APACrefbtitle {{ICC profile downloads - Windows - Adobe Inc..}} {{ICC profile downloads - Windows - Adobe Inc..}}
\newblock
\APAChowpublished {\url{https://www.adobe.com/support/downloads/iccprofiles/iccprofiles_win.html}}.
\PrintBackRefs{\CurrentBib}

\bibitem [\protect \citeauthoryear {%
Jia%
, Zhang%
, Shan%
\BCBL {}\ \BBA {} Chen%
}{%
Jia%
\ \protect \BOthers {.}}{%
{\protect \APACyear {2020}}%
}]{%
SSDG-CVPR-2020}
\APACinsertmetastar {%
SSDG-CVPR-2020}%
\begin{APACrefauthors}%
Jia, Y.%
, Zhang, J.%
, Shan, S.%
\BCBL {} Chen, X.%
\end{APACrefauthors}%
\unskip\
\newblock
\APACrefYearMonthDay{2020}{}{}.
\newblock
{\BBOQ}\APACrefatitle {{Single-Side Domain Generalization for Face Anti-Spoofing}} {{Single-Side Domain Generalization for Face Anti-Spoofing}}.{\BBCQ}
\newblock
 \APACrefbtitle {{2020 IEEE/CVF Conference on Computer Vision and Pattern Recognition (CVPR)}} {{2020 IEEE/CVF Conference on Computer Vision and Pattern Recognition (CVPR)}}\ (\BPG~8481-8490).
\newblock
\begin{APACrefDOI} \doi{10.1109/CVPR42600.2020.00851} \end{APACrefDOI}
\PrintBackRefs{\CurrentBib}

\bibitem [\protect \citeauthoryear {%
Jie%
\ \BBA {} Deng%
}{%
Jie%
\ \BBA {} Deng%
}{%
{\protect \APACyear {2022}}%
}]{%
convpass}
\APACinsertmetastar {%
convpass}%
\begin{APACrefauthors}%
Jie, S.%
\BCBT {}\ \BBA {} Deng, Z\BHBI H.%
\end{APACrefauthors}%
\unskip\
\newblock
\APACrefYearMonthDay{2022}{}{}.
\newblock
{\BBOQ}\APACrefatitle {Convolutional bypasses are better vision transformer adapters} {Convolutional bypasses are better vision transformer adapters}.{\BBCQ}
\newblock
\APACjournalVolNumPages{arXiv preprint arXiv:2207.07039}{}{}{}.
\newblock

\newblock

\PrintBackRefs{\CurrentBib}

\bibitem [\protect \citeauthoryear {%
Joshi%
}{%
Joshi%
}{%
{\protect \APACyear {2015}}%
}]{%
joshi2015opencv}
\APACinsertmetastar {%
joshi2015opencv}%
\begin{APACrefauthors}%
Joshi, P.%
\end{APACrefauthors}%
\unskip\
\newblock
\APACrefYearMonthDay{2015}{}{}.
\newblock
{\BBOQ}\APACrefatitle {OpenCV with Python by example} {Opencv with python by example}.{\BBCQ}
\newblock
\BIn{} (\BPG~39-40).
\newblock
\APACaddressPublisher{}{Packt Publishing Ltd}.
\PrintBackRefs{\CurrentBib}

\bibitem [\protect \citeauthoryear {%
Khalid%
}{%
Khalid%
}{%
{\protect \APACyear {2021}}%
}]{%
gabriel_carvalho_2021}
\APACinsertmetastar {%
gabriel_carvalho_2021}%
\begin{APACrefauthors}%
Khalid, M.J.%
\end{APACrefauthors}%
\unskip\
\newblock
\APACrefYearMonthDay{2021}{Apr}{}.
\newblock
\APACrefbtitle {Introduction to image processing using Python.} {Introduction to image processing using python.}
\newblock
\APAChowpublished {\url{https://ggcarvalho.dev/posts/imageproc/}}.
\PrintBackRefs{\CurrentBib}

\bibitem [\protect \citeauthoryear {%
Khosla%
\ \protect \BOthers {.}}{%
Khosla%
\ \protect \BOthers {.}}{%
{\protect \APACyear {2020}}%
}]{%
SupCon}
\APACinsertmetastar {%
SupCon}%
\begin{APACrefauthors}%
Khosla, P.%
, Teterwak, P.%
, Wang, C.%
, Sarna, A.%
, Tian, Y.%
, Isola, P.%
\BDBL {}Krishnan, D.%
\end{APACrefauthors}%
\unskip\
\newblock
\APACrefYearMonthDay{2020}{}{}.
\newblock
{\BBOQ}\APACrefatitle {Supervised contrastive learning} {Supervised contrastive learning}.{\BBCQ}
\newblock
\APACjournalVolNumPages{Advances in Neural Information Processing Systems}{33}{}{18661--18673}.
\newblock

\newblock

\PrintBackRefs{\CurrentBib}

\bibitem [\protect \citeauthoryear {%
King%
}{%
King%
}{%
{\protect \APACyear {2009}}%
}]{%
dlib09}
\APACinsertmetastar {%
dlib09}%
\begin{APACrefauthors}%
King, D.E.%
\end{APACrefauthors}%
\unskip\
\newblock
\APACrefYearMonthDay{2009}{}{}.
\newblock
{\BBOQ}\APACrefatitle {Dlib-ml: A Machine Learning Toolkit} {Dlib-ml: A machine learning toolkit}.{\BBCQ}
\newblock
\APACjournalVolNumPages{Journal of Machine Learning Research}{10}{}{1755-1758}.
\newblock

\newblock

\PrintBackRefs{\CurrentBib}

\bibitem [\protect \citeauthoryear {%
{Komulainen}%
, {Hadid}%
, {Pietikäinen}%
, {Anjos}%
\BCBL {}\ \BBA {} {Marcel}%
}{%
{Komulainen}%
\ \protect \BOthers {.}}{%
{\protect \APACyear {2013}}%
}]{%
MotionLBP-ICB-2013}
\APACinsertmetastar {%
MotionLBP-ICB-2013}%
\begin{APACrefauthors}%
{Komulainen}, J.%
, {Hadid}, A.%
, {Pietikäinen}, M.%
, {Anjos}, A.%
\BCBL {} {Marcel}, S.%
\end{APACrefauthors}%
\unskip\
\newblock
\APACrefYearMonthDay{2013}{June}{}.
\newblock
{\BBOQ}\APACrefatitle {Complementary countermeasures for detecting scenic face spoofing attacks} {Complementary countermeasures for detecting scenic face spoofing attacks}.{\BBCQ}
\newblock
 \APACrefbtitle {{2013 International Conference on Biometrics (ICB)}} {{2013 International Conference on Biometrics (ICB)}}\ (\BPG~1-7).
\newblock
\begin{APACrefDOI} \doi{10.1109/ICB.2013.6612968} \end{APACrefDOI}
\PrintBackRefs{\CurrentBib}

\bibitem [\protect \citeauthoryear {%
Lau%
\ \BBA {} Arce%
}{%
Lau%
\ \BBA {} Arce%
}{%
{\protect \APACyear {2018}}%
}]{%
dht_2018modern}
\APACinsertmetastar {%
dht_2018modern}%
\begin{APACrefauthors}%
Lau, D.L.%
\BCBT {}\ \BBA {} Arce, G.R.%
\end{APACrefauthors}%
\unskip\
\newblock
\APACrefYear{2018}.
\newblock
\APACrefbtitle {Modern digital halftoning} {Modern digital halftoning}.
\newblock
\APACaddressPublisher{}{CRC Press}.
\PrintBackRefs{\CurrentBib}

\bibitem [\protect \citeauthoryear {%
{Li}%
, {He}%
\BCBL {}\ \protect \BOthers {.}}{%
{Li}%
, {He}%
\BCBL {}\ \protect \BOthers {.}}{%
{\protect \APACyear {2018}}%
}]{%
FAS-3DCNN-TIFS-2018}
\APACinsertmetastar {%
FAS-3DCNN-TIFS-2018}%
\begin{APACrefauthors}%
{Li}, H.%
, {He}, P.%
, {Wang}, S.%
, {Rocha}, A.%
, {Jiang}, X.%
\BCBL {} {Kot}, A.C.%
\end{APACrefauthors}%
\unskip\
\newblock
\APACrefYearMonthDay{2018}{Oct}{}.
\newblock
{\BBOQ}\APACrefatitle {{Learning Generalized Deep Feature Representation for Face Anti-Spoofing}} {{Learning Generalized Deep Feature Representation for Face Anti-Spoofing}}.{\BBCQ}
\newblock
\APACjournalVolNumPages{IEEE Transactions on Information Forensics and Security}{13}{10}{2639-2652}.
\newblock

\newblock

\newblock
\begin{APACrefDOI} \doi{10.1109/TIFS.2018.2825949} \end{APACrefDOI}
\PrintBackRefs{\CurrentBib}

\bibitem [\protect \citeauthoryear {%
{Li}%
, {Li}%
\BCBL {}\ \protect \BOthers {.}}{%
{Li}%
, {Li}%
\BCBL {}\ \protect \BOthers {.}}{%
{\protect \APACyear {2018}}%
}]{%
FAS-UnsupervisedDA-TIFS-2018}
\APACinsertmetastar {%
FAS-UnsupervisedDA-TIFS-2018}%
\begin{APACrefauthors}%
{Li}, H.%
, {Li}, W.%
, {Cao}, H.%
, {Wang}, S.%
, {Huang}, F.%
\BCBL {} {Kot}, A.C.%
\end{APACrefauthors}%
\unskip\
\newblock
\APACrefYearMonthDay{2018}{July}{}.
\newblock
{\BBOQ}\APACrefatitle {{Unsupervised Domain Adaptation for Face Anti-Spoofing}} {{Unsupervised Domain Adaptation for Face Anti-Spoofing}}.{\BBCQ}
\newblock
\APACjournalVolNumPages{IEEE Transactions on Information Forensics and Security}{13}{7}{1794-1809}.
\newblock

\newblock

\newblock
\begin{APACrefDOI} \doi{10.1109/TIFS.2018.2801312} \end{APACrefDOI}
\PrintBackRefs{\CurrentBib}

\bibitem [\protect \citeauthoryear {%
H.~Li%
, Pan%
, Wang%
\BCBL {}\ \BBA {} Kot%
}{%
H.~Li%
\ \protect \BOthers {.}}{%
{\protect \APACyear {2018}}%
}]{%
MMDAAE-CVPR-2018}
\APACinsertmetastar {%
MMDAAE-CVPR-2018}%
\begin{APACrefauthors}%
Li, H.%
, Pan, S.J.%
, Wang, S.%
\BCBL {} Kot, A.C.%
\end{APACrefauthors}%
\unskip\
\newblock
\APACrefYearMonthDay{2018}{}{}.
\newblock
{\BBOQ}\APACrefatitle {{Domain Generalization with Adversarial Feature Learning}} {{Domain Generalization with Adversarial Feature Learning}}.{\BBCQ}
\newblock
 \APACrefbtitle {Proceedings of the IEEE Conference on Computer Vision and Pattern Recognition} {Proceedings of the ieee conference on computer vision and pattern recognition}\ (\BPGS\ 5400--5409).
\PrintBackRefs{\CurrentBib}

\bibitem [\protect \citeauthoryear {%
{Li}%
, {Wang}%
\BCBL {}\ \BBA {} {Kot}%
}{%
{Li}%
\ \protect \BOthers {.}}{%
{\protect \APACyear {2016}}%
}]{%
LI-QUALITY}
\APACinsertmetastar {%
LI-QUALITY}%
\begin{APACrefauthors}%
{Li}, H.%
, {Wang}, S.%
\BCBL {} {Kot}, A.C.%
\end{APACrefauthors}%
\unskip\
\newblock
\APACrefYearMonthDay{2016}{}{}.
\newblock
{\BBOQ}\APACrefatitle {Face spoofing detection with image quality regression} {Face spoofing detection with image quality regression}.{\BBCQ}
\newblock
 \APACrefbtitle {{2016 Sixth International Conference on Image Processing Theory, Tools and Applications (IPTA)}} {{2016 Sixth International Conference on Image Processing Theory, Tools and Applications (IPTA)}}\ (\BPG~1-6).
\PrintBackRefs{\CurrentBib}

\bibitem [\protect \citeauthoryear {%
L.~Li%
\ \protect \BOthers {.}}{%
L.~Li%
\ \protect \BOthers {.}}{%
{\protect \APACyear {2019}}%
}]{%
TIFS-2019-MotionBlur}
\APACinsertmetastar {%
TIFS-2019-MotionBlur}%
\begin{APACrefauthors}%
Li, L.%
, Xia, Z.%
, Hadid, A.%
, Jiang, X.%
, Zhang, H.%
\BCBL {} Feng, X.%
\end{APACrefauthors}%
\unskip\
\newblock
\APACrefYearMonthDay{2019}{}{}.
\newblock
{\BBOQ}\APACrefatitle {{Replayed Video Attack Detection Based on Motion Blur Analysis}} {{Replayed Video Attack Detection Based on Motion Blur Analysis}}.{\BBCQ}
\newblock
\APACjournalVolNumPages{IEEE Transactions on Information Forensics and Security}{14}{9}{2246-2261}.
\newblock

\newblock

\newblock
\begin{APACrefDOI} \doi{10.1109/TIFS.2019.2895212} \end{APACrefDOI}
\PrintBackRefs{\CurrentBib}

\bibitem [\protect \citeauthoryear {%
Y.~Li%
\ \protect \BOthers {.}}{%
Y.~Li%
\ \protect \BOthers {.}}{%
{\protect \APACyear {2020}}%
}]{%
li2020differentiable}
\APACinsertmetastar {%
li2020differentiable}%
\begin{APACrefauthors}%
Li, Y.%
, Hu, G.%
, Wang, Y.%
, Hospedales, T.%
, Robertson, N.M.%
\BCBL {} Yang, Y.%
\end{APACrefauthors}%
\unskip\
\newblock
\APACrefYearMonthDay{2020}{}{}.
\newblock
{\BBOQ}\APACrefatitle {Differentiable automatic data augmentation} {Differentiable automatic data augmentation}.{\BBCQ}
\newblock
 \APACrefbtitle {{European Conference on Computer Vision}} {{European Conference on Computer Vision}}\ (\BPGS\ 580--595).
\PrintBackRefs{\CurrentBib}

\bibitem [\protect \citeauthoryear {%
Z.~Li%
\ \protect \BOthers {.}}{%
Z.~Li%
\ \protect \BOthers {.}}{%
{\protect \APACyear {2022}}%
}]{%
li2022one}
\APACinsertmetastar {%
li2022one}%
\begin{APACrefauthors}%
Li, Z.%
, Cai, R.%
, Li, H.%
, Lam, K\BHBI Y.%
, Hu, Y.%
\BCBL {} Kot, A.C.%
\end{APACrefauthors}%
\unskip\
\newblock
\APACrefYearMonthDay{2022}{}{}.
\newblock
{\BBOQ}\APACrefatitle {One-Class Knowledge Distillation for Face Presentation Attack Detection} {One-class knowledge distillation for face presentation attack detection}.{\BBCQ}
\newblock
\APACjournalVolNumPages{IEEE Transactions on Information Forensics and Security}{}{}{}.
\newblock

\newblock

\PrintBackRefs{\CurrentBib}

\bibitem [\protect \citeauthoryear {%
Lin%
\ \protect \BOthers {.}}{%
Lin%
\ \protect \BOthers {.}}{%
{\protect \APACyear {2024}}%
}]{%
Lin_2024_CVPR}
\APACinsertmetastar {%
Lin_2024_CVPR}%
\begin{APACrefauthors}%
Lin, X.%
, Wang, S.%
, Cai, R.%
, Liu, Y.%
, Fu, Y.%
, Tang, W.%
\BDBL {}Kot, A.%
\end{APACrefauthors}%
\unskip\
\newblock
\APACrefYearMonthDay{2024}{June}{}.
\newblock
{\BBOQ}\APACrefatitle {Suppress and Rebalance: Towards Generalized Multi-Modal Face Anti-Spoofing} {Suppress and rebalance: Towards generalized multi-modal face anti-spoofing}.{\BBCQ}
\newblock
 \APACrefbtitle {{Proceedings of the IEEE/CVF Conference on Computer Vision and Pattern Recognition (CVPR)}} {{Proceedings of the IEEE/CVF Conference on Computer Vision and Pattern Recognition (CVPR)}}\ (\BPG~211-221).
\PrintBackRefs{\CurrentBib}

\bibitem [\protect \citeauthoryear {%
LingChen%
\ \protect \BOthers {.}}{%
LingChen%
\ \protect \BOthers {.}}{%
{\protect \APACyear {2020}}%
}]{%
lingchen2020uniformaugment}
\APACinsertmetastar {%
lingchen2020uniformaugment}%
\begin{APACrefauthors}%
LingChen, T.C.%
, Khonsari, A.%
, Lashkari, A.%
, Nazari, M.R.%
, Sambee, J.S.%
\BCBL {} Nascimento, M.A.%
\end{APACrefauthors}%
\unskip\
\newblock
\APACrefYearMonthDay{2020}{}{}.
\newblock
{\BBOQ}\APACrefatitle {{UniformAugment: A search-free probabilistic data augmentation approach}} {{UniformAugment: A search-free probabilistic data augmentation approach}}.{\BBCQ}
\newblock
\APACjournalVolNumPages{arXiv preprint arXiv:2003.14348}{}{}{}.
\newblock

\newblock

\PrintBackRefs{\CurrentBib}

\bibitem [\protect \citeauthoryear {%
A.~Liu%
\ \protect \BOthers {.}}{%
A.~Liu%
\ \protect \BOthers {.}}{%
{\protect \APACyear {2021}}%
}]{%
cefa}
\APACinsertmetastar {%
cefa}%
\begin{APACrefauthors}%
Liu, A.%
, Tan, Z.%
, Wan, J.%
, Escalera, S.%
, Guo, G.%
\BCBL {} Li, S.Z.%
\end{APACrefauthors}%
\unskip\
\newblock
\APACrefYearMonthDay{2021}{}{}.
\newblock
{\BBOQ}\APACrefatitle {{CASIA-SURF CeFA: A benchmark for multi-modal cross-ethnicity face anti-spoofing}} {{CASIA-SURF CeFA: A benchmark for multi-modal cross-ethnicity face anti-spoofing}}.{\BBCQ}
\newblock
 \APACrefbtitle {{Proceedings of the IEEE/CVF Winter Conference on Applications of Computer Vision}} {{Proceedings of the IEEE/CVF Winter Conference on Applications of Computer Vision}}\ (\BPGS\ 1179--1187).
\PrintBackRefs{\CurrentBib}

\bibitem [\protect \citeauthoryear {%
A.~Liu%
\ \protect \BOthers {.}}{%
A.~Liu%
\ \protect \BOthers {.}}{%
{\protect \APACyear {2022}}%
}]{%
hifi}
\APACinsertmetastar {%
hifi}%
\begin{APACrefauthors}%
Liu, A.%
, Zhao, C.%
, Yu, Z.%
, Wan, J.%
, Su, A.%
, Liu, X.%
\BDBL {}others%
\end{APACrefauthors}%
\unskip\
\newblock
\APACrefYearMonthDay{2022}{}{}.
\newblock
{\BBOQ}\APACrefatitle {Contrastive context-aware learning for 3d high-fidelity mask face presentation attack detection} {Contrastive context-aware learning for 3d high-fidelity mask face presentation attack detection}.{\BBCQ}
\newblock
\APACjournalVolNumPages{IEEE Transactions on Information Forensics and Security}{17}{}{2497--2507}.
\newblock

\newblock

\PrintBackRefs{\CurrentBib}

\bibitem [\protect \citeauthoryear {%
S.~Liu%
, Yang%
, Yuen%
\BCBL {}\ \BBA {} Zhao%
}{%
S.~Liu%
\ \protect \BOthers {.}}{%
{\protect \APACyear {2016}}%
}]{%
hkbuv2}
\APACinsertmetastar {%
hkbuv2}%
\begin{APACrefauthors}%
Liu, S.%
, Yang, B.%
, Yuen, P.C.%
\BCBL {} Zhao, G.%
\end{APACrefauthors}%
\unskip\
\newblock
\APACrefYearMonthDay{2016}{}{}.
\newblock
{\BBOQ}\APACrefatitle {{A 3D mask face anti-spoofing database with real world variations}} {{A 3D mask face anti-spoofing database with real world variations}}.{\BBCQ}
\newblock
 \APACrefbtitle {{Proceedings of the IEEE Conference on Computer Vision and Pattern Recognition Workshops}} {{Proceedings of the IEEE Conference on Computer Vision and Pattern Recognition Workshops}}\ (\BPGS\ 100--106).
\PrintBackRefs{\CurrentBib}

\bibitem [\protect \citeauthoryear {%
Y.~Liu%
\ \protect \BOthers {.}}{%
Y.~Liu%
\ \protect \BOthers {.}}{%
{\protect \APACyear {2022}}%
}]{%
liu2022source}
\APACinsertmetastar {%
liu2022source}%
\begin{APACrefauthors}%
Liu, Y.%
, Chen, Y.%
, Dai, W.%
, Gou, M.%
, Huang, C\BHBI T.%
\BCBL {} Xiong, H.%
\end{APACrefauthors}%
\unskip\
\newblock
\APACrefYearMonthDay{2022}{}{}.
\newblock
{\BBOQ}\APACrefatitle {Source-Free Domain Adaptation with Contrastive Domain Alignment and Self-supervised Exploration for Face Anti-spoofing} {Source-free domain adaptation with contrastive domain alignment and self-supervised exploration for face anti-spoofing}.{\BBCQ}
\newblock
 \APACrefbtitle {{European Conference on Computer Vision}} {{European Conference on Computer Vision}}\ (\BPGS\ 511--528).
\PrintBackRefs{\CurrentBib}

\bibitem [\protect \citeauthoryear {%
Y.~Liu%
, Jourabloo%
\BCBL {}\ \BBA {} Liu%
}{%
Y.~Liu%
\ \protect \BOthers {.}}{%
{\protect \APACyear {2018}}%
}]{%
FAS-Auxiliary-CVPR-2018}
\APACinsertmetastar {%
FAS-Auxiliary-CVPR-2018}%
\begin{APACrefauthors}%
Liu, Y.%
, Jourabloo, A.%
\BCBL {} Liu, X.%
\end{APACrefauthors}%
\unskip\
\newblock
\APACrefYearMonthDay{2018}{}{}.
\newblock
{\BBOQ}\APACrefatitle {{Learning Deep Models for Face Anti-Spoofing: Binary or Auxiliary Supervision}} {{Learning Deep Models for Face Anti-Spoofing: Binary or Auxiliary Supervision}}.{\BBCQ}
\newblock
 \APACrefbtitle {{Proceedings of the IEEE Conference on Computer Vision and Pattern Recognition}} {{Proceedings of the IEEE Conference on Computer Vision and Pattern Recognition}}\ (\BPGS\ 389--398).
\newblock
\APACaddressPublisher{Salt Lake City, UT}{}.
\PrintBackRefs{\CurrentBib}

\bibitem [\protect \citeauthoryear {%
Y.~Liu%
\ \BBA {} Liu%
}{%
Y.~Liu%
\ \BBA {} Liu%
}{%
{\protect \APACyear {2022}}%
}]{%
9779478}
\APACinsertmetastar {%
9779478}%
\begin{APACrefauthors}%
Liu, Y.%
\BCBT {}\ \BBA {} Liu, X.%
\end{APACrefauthors}%
\unskip\
\newblock
\APACrefYearMonthDay{2022}{}{}.
\newblock
{\BBOQ}\APACrefatitle {Spoof Trace Disentanglement for Generic Face Anti-Spoofing} {Spoof trace disentanglement for generic face anti-spoofing}.{\BBCQ}
\newblock
\APACjournalVolNumPages{IEEE Transactions on Pattern Analysis and Machine Intelligence}{}{}{1-1}.
\newblock

\newblock

\newblock
\begin{APACrefDOI} \doi{10.1109/TPAMI.2022.3176387} \end{APACrefDOI}
\PrintBackRefs{\CurrentBib}

\bibitem [\protect \citeauthoryear {%
Morris%
}{%
Morris%
}{%
{\protect \APACyear {2005}}%
}]{%
morris2005colour}
\APACinsertmetastar {%
morris2005colour}%
\begin{APACrefauthors}%
Morris, R.A.%
\end{APACrefauthors}%
\unskip\
\newblock
\APACrefYearMonthDay{2005}{}{}.
\newblock
{\BBOQ}\APACrefatitle {Colour management} {Colour management}.{\BBCQ}
\newblock
\APACjournalVolNumPages{H{\"a}user, CL, Steiner, A, Holstein, J, and Scoble M J. Digital imaging of biological type specimens. A manual for best practice. Stuttgart: European Network for Biodiversity Information}{}{}{31--36}.
\newblock

\newblock

\PrintBackRefs{\CurrentBib}

\bibitem [\protect \citeauthoryear {%
M{\"u}ller%
\ \BBA {} Hutter%
}{%
M{\"u}ller%
\ \BBA {} Hutter%
}{%
{\protect \APACyear {2021}}%
}]{%
muller2021trivialaugment}
\APACinsertmetastar {%
muller2021trivialaugment}%
\begin{APACrefauthors}%
M{\"u}ller, S.G.%
\BCBT {}\ \BBA {} Hutter, F.%
\end{APACrefauthors}%
\unskip\
\newblock
\APACrefYearMonthDay{2021}{}{}.
\newblock
{\BBOQ}\APACrefatitle {TrivialAugment: Tuning-free yet state-of-the-art data augmentation} {Trivialaugment: Tuning-free yet state-of-the-art data augmentation}.{\BBCQ}
\newblock
 \APACrefbtitle {{Proceedings of the IEEE/CVF International Conference on Computer Vision}} {{Proceedings of the IEEE/CVF International Conference on Computer Vision}}\ (\BPGS\ 774--782).
\PrintBackRefs{\CurrentBib}

\bibitem [\protect \citeauthoryear {%
Pappas%
}{%
Pappas%
}{%
{\protect \APACyear {1994}}%
}]{%
pappas1994digital}
\APACinsertmetastar {%
pappas1994digital}%
\begin{APACrefauthors}%
Pappas, T.N.%
\end{APACrefauthors}%
\unskip\
\newblock
\APACrefYearMonthDay{1994}{}{}.
\newblock
{\BBOQ}\APACrefatitle {Digital Halftoning Techniques for Printing} {Digital halftoning techniques for printing}.{\BBCQ}
\newblock
 \APACrefbtitle {Icps} {Icps}\ (\BVOL~94, \BPG~47th).
\PrintBackRefs{\CurrentBib}

\bibitem [\protect \citeauthoryear {%
P{\'e}rez-Cabo%
, Jim{\'e}nez-Cabello%
, Costa-Pazo%
\BCBL {}\ \BBA {} L{\'o}pez-Sastre%
}{%
P{\'e}rez-Cabo%
\ \protect \BOthers {.}}{%
{\protect \APACyear {2020}}%
}]{%
perez2020learning}
\APACinsertmetastar {%
perez2020learning}%
\begin{APACrefauthors}%
P{\'e}rez-Cabo, D.%
, Jim{\'e}nez-Cabello, D.%
, Costa-Pazo, A.%
\BCBL {} L{\'o}pez-Sastre, R.J.%
\end{APACrefauthors}%
\unskip\
\newblock
\APACrefYearMonthDay{2020}{}{}.
\newblock
{\BBOQ}\APACrefatitle {{Learning to Learn Face-PAD: a lifelong learning approach}} {{Learning to Learn Face-PAD: a lifelong learning approach}}.{\BBCQ}
\newblock
 \APACrefbtitle {{2020 IEEE International Joint Conference on Biometrics (IJCB)}} {{2020 IEEE International Joint Conference on Biometrics (IJCB)}}\ (\BPGS\ 1--9).
\PrintBackRefs{\CurrentBib}

\bibitem [\protect \citeauthoryear {%
Peters%
}{%
Peters%
}{%
{\protect \APACyear {2016}}%
}]{%
peters_2016}
\APACinsertmetastar {%
peters_2016}%
\begin{APACrefauthors}%
Peters, C.%
\end{APACrefauthors}%
\unskip\
\newblock
\APACrefYearMonthDay{2016}{Dec}{}.
\newblock
\APACrefbtitle {Free Blue Noise textures.} {Free blue noise textures.}
\newblock
\APAChowpublished {\url{http://momentsingraphics.de/BlueNoise.html}}.
\newblock
\APACaddressPublisher{}{Moments in Graphics}.
\PrintBackRefs{\CurrentBib}

\bibitem [\protect \citeauthoryear {%
Pfeiffer%
, Kamath%
, R{\"u}ckl{\'e}%
, Cho%
\BCBL {}\ \BBA {} Gurevych%
}{%
Pfeiffer%
\ \protect \BOthers {.}}{%
{\protect \APACyear {2021}}%
}]{%
pfeiffer2020adapterfusion}
\APACinsertmetastar {%
pfeiffer2020adapterfusion}%
\begin{APACrefauthors}%
Pfeiffer, J.%
, Kamath, A.%
, R{\"u}ckl{\'e}, A.%
, Cho, K.%
\BCBL {} Gurevych, I.%
\end{APACrefauthors}%
\unskip\
\newblock
\APACrefYearMonthDay{2021}{}{}.
\newblock
{\BBOQ}\APACrefatitle {AdapterFusion: Non-destructive task composition for transfer learning} {Adapterfusion: Non-destructive task composition for transfer learning}.{\BBCQ}
\newblock
 \APACrefbtitle {{Proceedings of the 16th Conference of the European Chapter of the Association for Computational Linguistics}.} {{Proceedings of the 16th Conference of the European Chapter of the Association for Computational Linguistics}.}
\PrintBackRefs{\CurrentBib}

\bibitem [\protect \citeauthoryear {%
Pinto%
\ \protect \BOthers {.}}{%
Pinto%
\ \protect \BOthers {.}}{%
{\protect \APACyear {2020}}%
}]{%
Pinto}
\APACinsertmetastar {%
Pinto}%
\begin{APACrefauthors}%
Pinto, A.%
, Goldenstein, S.%
, Ferreira, A.%
, Carvalho, T.%
, Pedrini, H.%
\BCBL {} Rocha, A.%
\end{APACrefauthors}%
\unskip\
\newblock
\APACrefYearMonthDay{2020}{}{}.
\newblock
{\BBOQ}\APACrefatitle {{Leveraging Shape, Reflectance and Albedo From Shading for Face Presentation Attack Detection}} {{Leveraging Shape, Reflectance and Albedo From Shading for Face Presentation Attack Detection}}.{\BBCQ}
\newblock
\APACjournalVolNumPages{IEEE Transactions on Information Forensics and Security}{15}{}{3347-3358}.
\newblock

\newblock

\newblock
\begin{APACrefDOI} \doi{10.1109/TIFS.2020.2988168} \end{APACrefDOI}
\PrintBackRefs{\CurrentBib}

\bibitem [\protect \citeauthoryear {%
Qin%
\ \protect \BOthers {.}}{%
Qin%
\ \protect \BOthers {.}}{%
{\protect \APACyear {2021}}%
}]{%
MetaTeacher-TPAMI-2021}
\APACinsertmetastar {%
MetaTeacher-TPAMI-2021}%
\begin{APACrefauthors}%
Qin, Y.%
, Yu, Z.%
, Yan, L.%
, Wang, Z.%
, Zhao, C.%
\BCBL {} Lei, Z.%
\end{APACrefauthors}%
\unskip\
\newblock
\APACrefYearMonthDay{2021}{}{}.
\newblock
{\BBOQ}\APACrefatitle {{Meta-teacher for Face Anti-Spoofing}} {{Meta-teacher for Face Anti-Spoofing}}.{\BBCQ}
\newblock
\APACjournalVolNumPages{IEEE Transactions on Pattern Analysis and Machine Intelligence}{Early Access}{}{1-1}.
\newblock

\newblock

\newblock
\begin{APACrefDOI} \doi{10.1109/TPAMI.2021.3091167} \end{APACrefDOI}
\PrintBackRefs{\CurrentBib}

\bibitem [\protect \citeauthoryear {%
Shao%
, Lan%
, Li%
\BCBL {}\ \BBA {} Yuen%
}{%
Shao%
\ \protect \BOthers {.}}{%
{\protect \APACyear {2019}}%
}]{%
MADDG-CVPR-2019}
\APACinsertmetastar {%
MADDG-CVPR-2019}%
\begin{APACrefauthors}%
Shao, R.%
, Lan, X.%
, Li, J.%
\BCBL {} Yuen, P.C.%
\end{APACrefauthors}%
\unskip\
\newblock
\APACrefYearMonthDay{2019}{}{}.
\newblock
{\BBOQ}\APACrefatitle {{Multi-Adversarial Discriminative Deep Domain Generalization for Face Presentation Attack Detection}} {{Multi-Adversarial Discriminative Deep Domain Generalization for Face Presentation Attack Detection}}.{\BBCQ}
\newblock
 \APACrefbtitle {{2019 IEEE/CVF Conference on Computer Vision and Pattern Recognition (CVPR)}} {{2019 IEEE/CVF Conference on Computer Vision and Pattern Recognition (CVPR)}}\ (\BPG~10015-10023).
\newblock
\begin{APACrefDOI} \doi{10.1109/CVPR.2019.01026} \end{APACrefDOI}
\PrintBackRefs{\CurrentBib}

\bibitem [\protect \citeauthoryear {%
Shao%
, Lan%
\BCBL {}\ \BBA {} Yuen%
}{%
Shao%
\ \protect \BOthers {.}}{%
{\protect \APACyear {2020}}%
}]{%
RFMetaFAS-AAAI-2020}
\APACinsertmetastar {%
RFMetaFAS-AAAI-2020}%
\begin{APACrefauthors}%
Shao, R.%
, Lan, X.%
\BCBL {} Yuen, P.C.%
\end{APACrefauthors}%
\unskip\
\newblock
\APACrefYearMonthDay{2020}{Apr.}{}.
\newblock
{\BBOQ}\APACrefatitle {{Regularized Fine-Grained Meta Face Anti-Spoofing}} {{Regularized Fine-Grained Meta Face Anti-Spoofing}}.{\BBCQ}
\newblock
\APACjournalVolNumPages{Proceedings of the AAAI Conference on Artificial Intelligence}{34}{07}{11974-11981}.
\newblock

\newblock

\newblock
\begin{APACrefDOI} \doi{10.1609/aaai.v34i07.6873} \end{APACrefDOI}
\PrintBackRefs{\CurrentBib}

\bibitem [\protect \citeauthoryear {%
Shorten%
\ \BBA {} Khoshgoftaar%
}{%
Shorten%
\ \BBA {} Khoshgoftaar%
}{%
{\protect \APACyear {2019}}%
}]{%
shorten2019survey}
\APACinsertmetastar {%
shorten2019survey}%
\begin{APACrefauthors}%
Shorten, C.%
\BCBT {}\ \BBA {} Khoshgoftaar, T.M.%
\end{APACrefauthors}%
\unskip\
\newblock
\APACrefYearMonthDay{2019}{}{}.
\newblock
{\BBOQ}\APACrefatitle {A survey on image data augmentation for deep learning} {A survey on image data augmentation for deep learning}.{\BBCQ}
\newblock
\APACjournalVolNumPages{Journal of Big Data}{6}{1}{1--48}.
\newblock

\newblock

\PrintBackRefs{\CurrentBib}

\bibitem [\protect \citeauthoryear {%
Spindler%
\ \protect \BOthers {.}}{%
Spindler%
\ \protect \BOthers {.}}{%
{\protect \APACyear {2006}}%
}]{%
spindler2006system}
\APACinsertmetastar {%
spindler2006system}%
\begin{APACrefauthors}%
Spindler, J.P.%
, Hatwar, T.K.%
, Miller, M.E.%
, Arnold, A.D.%
, Murdoch, M.J.%
, Kane, P.J.%
\BDBL {}Van~Slyke, S.A.%
\end{APACrefauthors}%
\unskip\
\newblock
\APACrefYearMonthDay{2006}{}{}.
\newblock
{\BBOQ}\APACrefatitle {System considerations for RGBW OLED displays} {System considerations for rgbw oled displays}.{\BBCQ}
\newblock
\APACjournalVolNumPages{Journal of the Society for Information Display}{14}{1}{37--48}.
\newblock

\newblock

\PrintBackRefs{\CurrentBib}

\bibitem [\protect \citeauthoryear {%
Srivatsan%
, Naseer%
\BCBL {}\ \BBA {} Nandakumar%
}{%
Srivatsan%
\ \protect \BOthers {.}}{%
{\protect \APACyear {2023}}%
}]{%
FLIP-MCL}
\APACinsertmetastar {%
FLIP-MCL}%
\begin{APACrefauthors}%
Srivatsan, K.%
, Naseer, M.%
\BCBL {} Nandakumar, K.%
\end{APACrefauthors}%
\unskip\
\newblock
\APACrefYearMonthDay{2023}{}{}.
\newblock
{\BBOQ}\APACrefatitle {{FLIP: Cross-domain face anti-spoofing with language guidance}} {{FLIP: Cross-domain face anti-spoofing with language guidance}}.{\BBCQ}
\newblock
 \APACrefbtitle {{Proceedings of the IEEE/CVF International Conference on Computer Vision}} {{Proceedings of the IEEE/CVF International Conference on Computer Vision}}\ (\BPGS\ 19685--19696).
\PrintBackRefs{\CurrentBib}

\bibitem [\protect \citeauthoryear {%
Sun%
, Song%
, Chen%
, Huang%
\BCBL {}\ \BBA {} Kot%
}{%
Sun%
\ \protect \BOthers {.}}{%
{\protect \APACyear {2020}}%
}]{%
Ternary-TIFS-2018}
\APACinsertmetastar {%
Ternary-TIFS-2018}%
\begin{APACrefauthors}%
Sun, W.%
, Song, Y.%
, Chen, C.%
, Huang, J.%
\BCBL {} Kot, A.C.%
\end{APACrefauthors}%
\unskip\
\newblock
\APACrefYearMonthDay{2020}{}{}.
\newblock
{\BBOQ}\APACrefatitle {{Face Spoofing Detection Based on Local Ternary Label Supervision in Fully Convolutional Networks}} {{Face Spoofing Detection Based on Local Ternary Label Supervision in Fully Convolutional Networks}}.{\BBCQ}
\newblock
\APACjournalVolNumPages{IEEE Transactions on Information Forensics and Security}{15}{}{3181-3196}.
\newblock

\newblock

\newblock
\begin{APACrefDOI} \doi{10.1109/TIFS.2020.2985530} \end{APACrefDOI}
\PrintBackRefs{\CurrentBib}

\bibitem [\protect \citeauthoryear {%
Troscianko%
\ \BBA {} Stevens%
}{%
Troscianko%
\ \BBA {} Stevens%
}{%
{\protect \APACyear {2015}}%
}]{%
troscianko2015image}
\APACinsertmetastar {%
troscianko2015image}%
\begin{APACrefauthors}%
Troscianko, J.%
\BCBT {}\ \BBA {} Stevens, M.%
\end{APACrefauthors}%
\unskip\
\newblock
\APACrefYearMonthDay{2015}{}{}.
\newblock
{\BBOQ}\APACrefatitle {Image calibration and analysis toolbox--a free software suite for objectively measuring reflectance, colour and pattern} {Image calibration and analysis toolbox--a free software suite for objectively measuring reflectance, colour and pattern}.{\BBCQ}
\newblock
\APACjournalVolNumPages{Methods in Ecology and Evolution}{6}{11}{1320--1331}.
\newblock

\newblock

\PrintBackRefs{\CurrentBib}

\bibitem [\protect \citeauthoryear {%
Ulichney%
}{%
Ulichney%
}{%
{\protect \APACyear {1988}}%
}]{%
ulichney1988dithering}
\APACinsertmetastar {%
ulichney1988dithering}%
\begin{APACrefauthors}%
Ulichney, R.A.%
\end{APACrefauthors}%
\unskip\
\newblock
\APACrefYearMonthDay{1988}{}{}.
\newblock
{\BBOQ}\APACrefatitle {Dithering with blue noise} {Dithering with blue noise}.{\BBCQ}
\newblock
\APACjournalVolNumPages{Proceedings of the IEEE}{76}{1}{56--79}.
\newblock

\newblock

\PrintBackRefs{\CurrentBib}

\bibitem [\protect \citeauthoryear {%
Ulichney%
}{%
Ulichney%
}{%
{\protect \APACyear {1993}}%
}]{%
ulichney1993void}
\APACinsertmetastar {%
ulichney1993void}%
\begin{APACrefauthors}%
Ulichney, R.A.%
\end{APACrefauthors}%
\unskip\
\newblock
\APACrefYearMonthDay{1993}{}{}.
\newblock
{\BBOQ}\APACrefatitle {Void-and-cluster method for dither array generation} {Void-and-cluster method for dither array generation}.{\BBCQ}
\newblock
 \APACrefbtitle {Human Vision, Visual Processing, and Digital Display IV} {Human vision, visual processing, and digital display iv}\ (\BVOL\ 1913, \BPGS\ 332--343).
\PrintBackRefs{\CurrentBib}

\bibitem [\protect \citeauthoryear {%
Wan%
, Shi%
, Duan%
, Tan%
\BCBL {}\ \BBA {} Kot%
}{%
Wan%
\ \protect \BOthers {.}}{%
{\protect \APACyear {2017}}%
}]{%
wan2017benchmarking_iccv}
\APACinsertmetastar {%
wan2017benchmarking_iccv}%
\begin{APACrefauthors}%
Wan, R.%
, Shi, B.%
, Duan, L\BHBI Y.%
, Tan, A\BHBI H.%
\BCBL {} Kot, A.C.%
\end{APACrefauthors}%
\unskip\
\newblock
\APACrefYearMonthDay{2017}{}{}.
\newblock
{\BBOQ}\APACrefatitle {Benchmarking single-image reflection removal algorithms} {Benchmarking single-image reflection removal algorithms}.{\BBCQ}
\newblock
 \APACrefbtitle {{Proceedings of the IEEE International Conference on Computer Vision}} {{Proceedings of the IEEE International Conference on Computer Vision}}\ (\BPGS\ 3922--3930).
\PrintBackRefs{\CurrentBib}

\bibitem [\protect \citeauthoryear {%
Wan%
\ \protect \BOthers {.}}{%
Wan%
\ \protect \BOthers {.}}{%
{\protect \APACyear {2022}}%
}]{%
wan2017benchmarking_pami}
\APACinsertmetastar {%
wan2017benchmarking_pami}%
\begin{APACrefauthors}%
Wan, R.%
, Shi, B.%
, Li, H.%
, Hong, Y.%
, Duan, L.%
\BCBL {} Kot~Chichung, A.%
\end{APACrefauthors}%
\unskip\
\newblock
\APACrefYearMonthDay{2022}{}{}.
\newblock
{\BBOQ}\APACrefatitle {Benchmarking single-image reflection removal algorithms} {Benchmarking single-image reflection removal algorithms}.{\BBCQ}
\newblock
\APACjournalVolNumPages{IEEE Transactions on Pattern Analysis and Machine Intelligence}{}{}{1-1}.
\newblock

\newblock

\newblock
\begin{APACrefDOI} \doi{10.1109/TPAMI.2022.3168560} \end{APACrefDOI}
\PrintBackRefs{\CurrentBib}

\bibitem [\protect \citeauthoryear {%
C\BHBI Y.~Wang%
, Lu%
, Yang%
\BCBL {}\ \BBA {} Lai%
}{%
C\BHBI Y.~Wang%
\ \protect \BOthers {.}}{%
{\protect \APACyear {2022}}%
}]{%
wang2022patchnet}
\APACinsertmetastar {%
wang2022patchnet}%
\begin{APACrefauthors}%
Wang, C\BHBI Y.%
, Lu, Y\BHBI D.%
, Yang, S\BHBI T.%
\BCBL {} Lai, S\BHBI H.%
\end{APACrefauthors}%
\unskip\
\newblock
\APACrefYearMonthDay{2022}{}{}.
\newblock
{\BBOQ}\APACrefatitle {{PatchNet: A Simple Face Anti-Spoofing Framework via Fine-Grained Patch Recognition}} {{PatchNet: A Simple Face Anti-Spoofing Framework via Fine-Grained Patch Recognition}}.{\BBCQ}
\newblock
 \APACrefbtitle {Proceedings of the IEEE/CVF Conference on Computer Vision and Pattern Recognition} {Proceedings of the ieee/cvf conference on computer vision and pattern recognition}\ (\BPGS\ 20281--20290).
\PrintBackRefs{\CurrentBib}

\bibitem [\protect \citeauthoryear {%
G.~Wang%
, Han%
, Shan%
\BCBL {}\ \BBA {} Chen%
}{%
G.~Wang%
\ \protect \BOthers {.}}{%
{\protect \APACyear {2020}}%
{\protect \APACexlab {{\protect \BCnt {1}}}}}]{%
wang2020cross}
\APACinsertmetastar {%
wang2020cross}%
\begin{APACrefauthors}%
Wang, G.%
, Han, H.%
, Shan, S.%
\BCBL {} Chen, X.%
\end{APACrefauthors}%
\unskip\
\newblock
\APACrefYearMonthDay{2020{\protect \BCnt {1}}}{}{}.
\newblock
{\BBOQ}\APACrefatitle {{Cross-Domain Face Presentation Attack Detection via Multi-Domain Disentangled Representation Learning}} {{Cross-Domain Face Presentation Attack Detection via Multi-Domain Disentangled Representation Learning}}.{\BBCQ}
\newblock
 \APACrefbtitle {{Proceedings of the IEEE/CVF Conference on Computer Vision and Pattern Recognition}} {{Proceedings of the IEEE/CVF Conference on Computer Vision and Pattern Recognition}}\ (\BPGS\ 6678--6687).
\PrintBackRefs{\CurrentBib}

\bibitem [\protect \citeauthoryear {%
G.~Wang%
, Han%
, Shan%
\BCBL {}\ \BBA {} Chen%
}{%
G.~Wang%
\ \protect \BOthers {.}}{%
{\protect \APACyear {2020}}%
{\protect \APACexlab {{\protect \BCnt {2}}}}}]{%
wang2020unsupervised}
\APACinsertmetastar {%
wang2020unsupervised}%
\begin{APACrefauthors}%
Wang, G.%
, Han, H.%
, Shan, S.%
\BCBL {} Chen, X.%
\end{APACrefauthors}%
\unskip\
\newblock
\APACrefYearMonthDay{2020{\protect \BCnt {2}}}{}{}.
\newblock
{\BBOQ}\APACrefatitle {Unsupervised adversarial domain adaptation for cross-domain face presentation attack detection} {Unsupervised adversarial domain adaptation for cross-domain face presentation attack detection}.{\BBCQ}
\newblock
\APACjournalVolNumPages{IEEE Transactions on Information Forensics and Security}{16}{}{56--69}.
\newblock

\newblock

\PrintBackRefs{\CurrentBib}

\bibitem [\protect \citeauthoryear {%
J.~Wang%
\ \protect \BOthers {.}}{%
J.~Wang%
\ \protect \BOthers {.}}{%
{\protect \APACyear {2021}}%
}]{%
wang2021self}
\APACinsertmetastar {%
wang2021self}%
\begin{APACrefauthors}%
Wang, J.%
, Zhang, J.%
, Bian, Y.%
, Cai, Y.%
, Wang, C.%
\BCBL {} Pu, S.%
\end{APACrefauthors}%
\unskip\
\newblock
\APACrefYearMonthDay{2021}{}{}.
\newblock
{\BBOQ}\APACrefatitle {Self-domain adaptation for face anti-spoofing} {Self-domain adaptation for face anti-spoofing}.{\BBCQ}
\newblock
 \APACrefbtitle {{Proceedings of the AAAI Conference on Artificial Intelligence}} {{Proceedings of the AAAI Conference on Artificial Intelligence}}\ (\BVOL~35, \BPGS\ 2746--2754).
\PrintBackRefs{\CurrentBib}

\bibitem [\protect \citeauthoryear {%
W.~Wang%
, Liu%
, Zheng%
, Ying%
\BCBL {}\ \BBA {} Wen%
}{%
W.~Wang%
\ \protect \BOthers {.}}{%
{\protect \APACyear {2023}}%
}]{%
TIFS-negative}
\APACinsertmetastar {%
TIFS-negative}%
\begin{APACrefauthors}%
Wang, W.%
, Liu, P.%
, Zheng, H.%
, Ying, R.%
\BCBL {} Wen, F.%
\end{APACrefauthors}%
\unskip\
\newblock
\APACrefYearMonthDay{2023}{}{}.
\newblock
{\BBOQ}\APACrefatitle {Domain Generalization for Face Anti-Spoofing via Negative Data Augmentation} {Domain generalization for face anti-spoofing via negative data augmentation}.{\BBCQ}
\newblock
\APACjournalVolNumPages{IEEE Transactions on Information Forensics and Security}{18}{}{2333-2344}.
\newblock

\newblock

\newblock
\begin{APACrefDOI} \doi{10.1109/TIFS.2023.3266138} \end{APACrefDOI}
\PrintBackRefs{\CurrentBib}

\bibitem [\protect \citeauthoryear {%
W.~Wang%
\ \protect \BOthers {.}}{%
W.~Wang%
\ \protect \BOthers {.}}{%
{\protect \APACyear {2019}}%
}]{%
FAS-CVPR2019-STASN}
\APACinsertmetastar {%
FAS-CVPR2019-STASN}%
\begin{APACrefauthors}%
Wang, W.%
, Luo, W.%
, Bao, L.%
, Gao, Y.%
, Gong, D.%
, Zheng, S.%
\BDBL {}Overwijk, A.%
\end{APACrefauthors}%
\unskip\
\newblock
\APACrefYearMonthDay{2019}{}{}.
\newblock
{\BBOQ}\APACrefatitle {{Face Anti-Spoofing: Model Matters, so Does Data}} {{Face Anti-Spoofing: Model Matters, so Does Data}}.{\BBCQ}
\newblock
 \APACrefbtitle {{Proceedings of the IEEE Conference on Computer Vision and Pattern Recognition}.} {{Proceedings of the IEEE Conference on Computer Vision and Pattern Recognition}.}
\PrintBackRefs{\CurrentBib}

\bibitem [\protect \citeauthoryear {%
Z.~Wang%
\ \protect \BOthers {.}}{%
Z.~Wang%
\ \protect \BOthers {.}}{%
{\protect \APACyear {2022}}%
}]{%
SSAN}
\APACinsertmetastar {%
SSAN}%
\begin{APACrefauthors}%
Wang, Z.%
, Wang, Z.%
, Yu, Z.%
, Deng, W.%
, Li, J.%
, Gao, T.%
\BCBL {} Wang, Z.%
\end{APACrefauthors}%
\unskip\
\newblock
\APACrefYearMonthDay{2022}{}{}.
\newblock
{\BBOQ}\APACrefatitle {{Domain Generalization via Shuffled Style Assembly for Face Anti-Spoofing}} {{Domain Generalization via Shuffled Style Assembly for Face Anti-Spoofing}}.{\BBCQ}
\newblock
 \APACrefbtitle {Proceedings of the IEEE/CVF Conference on Computer Vision and Pattern Recognition} {Proceedings of the ieee/cvf conference on computer vision and pattern recognition}\ (\BPGS\ 4123--4133).
\PrintBackRefs{\CurrentBib}

\bibitem [\protect \citeauthoryear {%
{Wen}%
, {Han}%
\BCBL {}\ \BBA {} {Jain}%
}{%
{Wen}%
\ \protect \BOthers {.}}{%
{\protect \APACyear {2015}}%
}]{%
FAS-IDA-TIFS-2015}
\APACinsertmetastar {%
FAS-IDA-TIFS-2015}%
\begin{APACrefauthors}%
{Wen}, D.%
, {Han}, H.%
\BCBL {} {Jain}, A.K.%
\end{APACrefauthors}%
\unskip\
\newblock
\APACrefYearMonthDay{2015}{April}{}.
\newblock
{\BBOQ}\APACrefatitle {{Face Spoof Detection With Image Distortion Analysis}} {{Face Spoof Detection With Image Distortion Analysis}}.{\BBCQ}
\newblock
\APACjournalVolNumPages{IEEE Transactions on Information Forensics and Security}{10}{4}{746-761}.
\newblock

\newblock

\newblock
\begin{APACrefDOI} \doi{10.1109/TIFS.2015.2400395} \end{APACrefDOI}
\PrintBackRefs{\CurrentBib}

\bibitem [\protect \citeauthoryear {%
Wu%
, Zeng%
, Hu%
, Shi%
\BCBL {}\ \BBA {} Mei%
}{%
Wu%
\ \protect \BOthers {.}}{%
{\protect \APACyear {2021}}%
}]{%
wu2021dual}
\APACinsertmetastar {%
wu2021dual}%
\begin{APACrefauthors}%
Wu, H.%
, Zeng, D.%
, Hu, Y.%
, Shi, H.%
\BCBL {} Mei, T.%
\end{APACrefauthors}%
\unskip\
\newblock
\APACrefYearMonthDay{2021}{}{}.
\newblock
{\BBOQ}\APACrefatitle {Dual Spoof Disentanglement Generation for Face Anti-spoofing with Depth Uncertainty Learning} {Dual spoof disentanglement generation for face anti-spoofing with depth uncertainty learning}.{\BBCQ}
\newblock
\APACjournalVolNumPages{IEEE Transactions on Circuits and Systems for Video Technology}{}{}{}.
\newblock

\newblock

\PrintBackRefs{\CurrentBib}

\bibitem [\protect \citeauthoryear {%
Xiao%
, Farrell%
, Catrysse%
\BCBL {}\ \BBA {} Wandell%
}{%
Xiao%
\ \protect \BOthers {.}}{%
{\protect \APACyear {2009}}%
}]{%
xiao2009mobile}
\APACinsertmetastar {%
xiao2009mobile}%
\begin{APACrefauthors}%
Xiao, F.%
, Farrell, J.E.%
, Catrysse, P.B.%
\BCBL {} Wandell, B.%
\end{APACrefauthors}%
\unskip\
\newblock
\APACrefYearMonthDay{2009}{}{}.
\newblock
{\BBOQ}\APACrefatitle {Mobile imaging: the big challenge of the small pixel} {Mobile imaging: the big challenge of the small pixel}.{\BBCQ}
\newblock
 \APACrefbtitle {Digital Photography V} {Digital photography v}\ (\BVOL\ 7250, \BPGS\ 173--181).
\PrintBackRefs{\CurrentBib}

\bibitem [\protect \citeauthoryear {%
Xiong%
\ \protect \BOthers {.}}{%
Xiong%
\ \protect \BOthers {.}}{%
{\protect \APACyear {2009}}%
}]{%
xiong2009performance}
\APACinsertmetastar {%
xiong2009performance}%
\begin{APACrefauthors}%
Xiong, Y.%
, Wang, L.%
, Xu, W.%
, Zou, J.%
, Wu, H.%
, Xu, Y.%
\BDBL {}Yu, G.%
\end{APACrefauthors}%
\unskip\
\newblock
\APACrefYearMonthDay{2009}{}{}.
\newblock
{\BBOQ}\APACrefatitle {Performance analysis of PLED based flat panel display with RGBW sub-pixel layout} {Performance analysis of pled based flat panel display with rgbw sub-pixel layout}.{\BBCQ}
\newblock
\APACjournalVolNumPages{Organic Electronics}{10}{5}{857--862}.
\newblock

\newblock

\PrintBackRefs{\CurrentBib}

\bibitem [\protect \citeauthoryear {%
Yang%
, Cai%
\BCBL {}\ \BBA {} Kot%
}{%
Yang%
\ \protect \BOthers {.}}{%
{\protect \APACyear {2022}}%
}]{%
AMCMM2022}
\APACinsertmetastar {%
AMCMM2022}%
\begin{APACrefauthors}%
Yang, W.%
, Cai, R.%
\BCBL {} Kot, A.%
\end{APACrefauthors}%
\unskip\
\newblock
\APACrefYearMonthDay{2022}{}{}.
\newblock
{\BBOQ}\APACrefatitle {Image Inpainting Detection via Enriched Attentive Pattern with Near Original Image Augmentation} {Image inpainting detection via enriched attentive pattern with near original image augmentation}.{\BBCQ}
\newblock
 \APACrefbtitle {Proceedings of the 30th ACM International Conference on Multimedia} {Proceedings of the 30th acm international conference on multimedia}\ (\BPG~2816–2824).
\newblock
\APACaddressPublisher{New York, NY, USA}{Association for Computing Machinery}.
\newblock
\begin{APACrefURL} {https://doi.org/10.1145/3503161.3547921} \end{APACrefURL}
\newblock
\begin{APACrefDOI} \doi{10.1145/3503161.3547921} \end{APACrefDOI}
\PrintBackRefs{\CurrentBib}

\bibitem [\protect \citeauthoryear {%
Yu%
\ \protect \BOthers {.}}{%
Yu%
\ \protect \BOthers {.}}{%
{\protect \APACyear {2024}}%
}]{%
yu2024benchmarking}
\APACinsertmetastar {%
yu2024benchmarking}%
\begin{APACrefauthors}%
Yu, Z.%
, Cai, R.%
, Li, Z.%
, Yang, W.%
, Shi, J.%
\BCBL {} Kot, A.C.%
\end{APACrefauthors}%
\unskip\
\newblock
\APACrefYearMonthDay{2024}{}{}.
\newblock
{\BBOQ}\APACrefatitle {{Benchmarking Joint Face Spoofing and Forgery Detection with Visual and Physiological Cues}} {{Benchmarking Joint Face Spoofing and Forgery Detection with Visual and Physiological Cues}}.{\BBCQ}
\newblock
\APACjournalVolNumPages{IEEE Transactions on Dependable and Secure Computing (TDSC)}{}{}{}.
\newblock

\newblock

\PrintBackRefs{\CurrentBib}

\bibitem [\protect \citeauthoryear {%
Yu%
, Li%
, Shi%
, Xia%
\BCBL {}\ \BBA {} Zhao%
}{%
Yu%
, Li%
\BCBL {}\ \protect \BOthers {.}}{%
{\protect \APACyear {2021}}%
}]{%
yu2021revisiting}
\APACinsertmetastar {%
yu2021revisiting}%
\begin{APACrefauthors}%
Yu, Z.%
, Li, X.%
, Shi, J.%
, Xia, Z.%
\BCBL {} Zhao, G.%
\end{APACrefauthors}%
\unskip\
\newblock
\APACrefYearMonthDay{2021}{}{}.
\newblock
{\BBOQ}\APACrefatitle {{Revisiting Pixel-Wise Supervision for Face Anti-Spoofing}} {{Revisiting Pixel-Wise Supervision for Face Anti-Spoofing}}.{\BBCQ}
\newblock
\APACjournalVolNumPages{IEEE Transactions on Biometrics, Behavior, and Identity Science}{3}{3}{285-295}.
\newblock

\newblock

\newblock
\begin{APACrefDOI} \doi{10.1109/TBIOM.2021.3065526} \end{APACrefDOI}
\PrintBackRefs{\CurrentBib}

\bibitem [\protect \citeauthoryear {%
Yu%
\ \protect \BOthers {.}}{%
Yu%
\ \protect \BOthers {.}}{%
{\protect \APACyear {2022}}%
}]{%
yu2022deep}
\APACinsertmetastar {%
yu2022deep}%
\begin{APACrefauthors}%
Yu, Z.%
, Qin, Y.%
, Li, X.%
, Zhao, C.%
, Lei, Z.%
\BCBL {} Zhao, G.%
\end{APACrefauthors}%
\unskip\
\newblock
\APACrefYearMonthDay{2022}{}{}.
\newblock
{\BBOQ}\APACrefatitle {Deep learning for face anti-spoofing: A survey} {Deep learning for face anti-spoofing: A survey}.{\BBCQ}
\newblock
\APACjournalVolNumPages{IEEE Transactions on Pattern Analysis and Machine Intelligence}{}{}{}.
\newblock

\newblock

\PrintBackRefs{\CurrentBib}

\bibitem [\protect \citeauthoryear {%
Yu%
, Qin%
, Zhao%
, Li%
\BCBL {}\ \BBA {} Zhao%
}{%
Yu%
, Qin%
\BCBL {}\ \protect \BOthers {.}}{%
{\protect \APACyear {2021}}%
}]{%
DCCDN_IJCAI_2021}
\APACinsertmetastar {%
DCCDN_IJCAI_2021}%
\begin{APACrefauthors}%
Yu, Z.%
, Qin, Y.%
, Zhao, H.%
, Li, X.%
\BCBL {} Zhao, G.%
\end{APACrefauthors}%
\unskip\
\newblock
\APACrefYearMonthDay{2021}{}{}.
\newblock
{\BBOQ}\APACrefatitle {{Dual-Cross Central Difference Network for Face Anti-Spoofing}} {{Dual-Cross Central Difference Network for Face Anti-Spoofing}}.{\BBCQ}
\newblock
 \APACrefbtitle {{International Joint Conference on Artificial Intelligence (IJCAI)}.} {{International Joint Conference on Artificial Intelligence (IJCAI)}.}
\PrintBackRefs{\CurrentBib}

\bibitem [\protect \citeauthoryear {%
Yu%
, Wan%
\BCBL {}\ \protect \BOthers {.}}{%
Yu%
, Wan%
\BCBL {}\ \protect \BOthers {.}}{%
{\protect \APACyear {2021}}%
}]{%
NASFAS-TPAMI-2020}
\APACinsertmetastar {%
NASFAS-TPAMI-2020}%
\begin{APACrefauthors}%
Yu, Z.%
, Wan, J.%
, Qin, Y.%
, Li, X.%
, Li, S.Z.%
\BCBL {} Zhao, G.%
\end{APACrefauthors}%
\unskip\
\newblock
\APACrefYearMonthDay{2021}{}{}.
\newblock
{\BBOQ}\APACrefatitle {{NAS-FAS: Static-Dynamic Central Difference Network Search for Face Anti-Spoofing}} {{NAS-FAS: Static-Dynamic Central Difference Network Search for Face Anti-Spoofing}}.{\BBCQ}
\newblock
\APACjournalVolNumPages{IEEE Transactions on Pattern Analysis and Machine Intelligence}{43}{9}{3005-3023}.
\newblock

\newblock

\newblock
\begin{APACrefDOI} \doi{10.1109/TPAMI.2020.3036338} \end{APACrefDOI}
\PrintBackRefs{\CurrentBib}

\bibitem [\protect \citeauthoryear {%
Yu%
\ \protect \BOthers {.}}{%
Yu%
\ \protect \BOthers {.}}{%
{\protect \APACyear {2020}}%
}]{%
CDCN-CVPR-2020}
\APACinsertmetastar {%
CDCN-CVPR-2020}%
\begin{APACrefauthors}%
Yu, Z.%
, Zhao, C.%
, Wang, Z.%
, Qin, Y.%
, Su, Z.%
, Li, X.%
\BDBL {}Zhao, G.%
\end{APACrefauthors}%
\unskip\
\newblock
\APACrefYearMonthDay{2020}{}{}.
\newblock
{\BBOQ}\APACrefatitle {{Searching Central Difference Convolutional Networks for Face Anti-Spoofing}} {{Searching Central Difference Convolutional Networks for Face Anti-Spoofing}}.{\BBCQ}
\newblock
 \APACrefbtitle {{2020 IEEE/CVF Conference on Computer Vision and Pattern Recognition (CVPR)}} {{2020 IEEE/CVF Conference on Computer Vision and Pattern Recognition (CVPR)}}\ (\BPG~5294-5304).
\newblock
\begin{APACrefDOI} \doi{10.1109/CVPR42600.2020.00534} \end{APACrefDOI}
\PrintBackRefs{\CurrentBib}

\bibitem [\protect \citeauthoryear {%
Yuan%
, Timofte%
, Slabaugh%
\BCBL {}\ \BBA {} Leonardis%
}{%
Yuan%
\ \protect \BOthers {.}}{%
{\protect \APACyear {2019}}%
}]{%
yuan2019aim}
\APACinsertmetastar {%
yuan2019aim}%
\begin{APACrefauthors}%
Yuan, S.%
, Timofte, R.%
, Slabaugh, G.%
\BCBL {} Leonardis, A.%
\end{APACrefauthors}%
\unskip\
\newblock
\APACrefYearMonthDay{2019}{}{}.
\newblock
{\BBOQ}\APACrefatitle {Aim 2019 challenge on image demoireing: Dataset and study} {Aim 2019 challenge on image demoireing: Dataset and study}.{\BBCQ}
\newblock
 \APACrefbtitle {{2019 IEEE/CVF International Conference on Computer Vision Workshop (ICCVW)}} {{2019 IEEE/CVF International Conference on Computer Vision Workshop (ICCVW)}}\ (\BPGS\ 3526--3533).
\PrintBackRefs{\CurrentBib}

\bibitem [\protect \citeauthoryear {%
Zha%
\ \protect \BOthers {.}}{%
Zha%
\ \protect \BOthers {.}}{%
{\protect \APACyear {2023}}%
}]{%
data-ai}
\APACinsertmetastar {%
data-ai}%
\begin{APACrefauthors}%
Zha, D.%
, Bhat, Z.P.%
, Lai, K\BHBI H.%
, Yang, F.%
, Jiang, Z.%
, Zhong, S.%
\BCBL {} Hu, X.%
\end{APACrefauthors}%
\unskip\
\newblock
\APACrefYearMonthDay{2023}{}{}.
\newblock
{\BBOQ}\APACrefatitle {Data-centric artificial intelligence: A survey} {Data-centric artificial intelligence: A survey}.{\BBCQ}
\newblock
\APACjournalVolNumPages{arXiv preprint arXiv:2303.10158}{}{}{}.
\newblock

\newblock

\PrintBackRefs{\CurrentBib}

\bibitem [\protect \citeauthoryear {%
K\BHBI Y.~Zhang%
\ \protect \BOthers {.}}{%
K\BHBI Y.~Zhang%
\ \protect \BOthers {.}}{%
{\protect \APACyear {2021}}%
}]{%
zhang2021structure}
\APACinsertmetastar {%
zhang2021structure}%
\begin{APACrefauthors}%
Zhang, K\BHBI Y.%
, Yao, T.%
, Zhang, J.%
, Liu, S.%
, Yin, B.%
, Ding, S.%
\BCBL {} Li, J.%
\end{APACrefauthors}%
\unskip\
\newblock
\APACrefYearMonthDay{2021}{}{}.
\newblock
{\BBOQ}\APACrefatitle {Structure destruction and content combination for face anti-spoofing} {Structure destruction and content combination for face anti-spoofing}.{\BBCQ}
\newblock
 \APACrefbtitle {{2021 IEEE International Joint Conference on Biometrics (IJCB)}} {{2021 IEEE International Joint Conference on Biometrics (IJCB)}}\ (\BPGS\ 1--6).
\PrintBackRefs{\CurrentBib}

\bibitem [\protect \citeauthoryear {%
S.~Zhang%
\ \protect \BOthers {.}}{%
S.~Zhang%
\ \protect \BOthers {.}}{%
{\protect \APACyear {2020}}%
}]{%
CASIA-SURF}
\APACinsertmetastar {%
CASIA-SURF}%
\begin{APACrefauthors}%
Zhang, S.%
, Liu, A.%
, Wan, J.%
, Liang, Y.%
, Guo, G.%
, Escalera, S.%
\BDBL {}Li, S.Z.%
\end{APACrefauthors}%
\unskip\
\newblock
\APACrefYearMonthDay{2020}{}{}.
\newblock
{\BBOQ}\APACrefatitle {Casia-surf: A large-scale multi-modal benchmark for face anti-spoofing} {Casia-surf: A large-scale multi-modal benchmark for face anti-spoofing}.{\BBCQ}
\newblock
\APACjournalVolNumPages{IEEE Transactions on Biometrics, Behavior, and Identity Science}{2}{2}{182--193}.
\newblock

\newblock

\PrintBackRefs{\CurrentBib}

\bibitem [\protect \citeauthoryear {%
X.~Zhang%
, Wang%
, Zhang%
\BCBL {}\ \BBA {} Zhong%
}{%
X.~Zhang%
\ \protect \BOthers {.}}{%
{\protect \APACyear {2020}}%
}]{%
adv_auto}
\APACinsertmetastar {%
adv_auto}%
\begin{APACrefauthors}%
Zhang, X.%
, Wang, Q.%
, Zhang, J.%
\BCBL {} Zhong, Z.%
\end{APACrefauthors}%
\unskip\
\newblock
\APACrefYearMonthDay{2020}{}{}.
\newblock
{\BBOQ}\APACrefatitle {{Adversarial AutoAugment}} {{Adversarial AutoAugment}}.{\BBCQ}
\newblock
 \APACrefbtitle {{2020 International Conference on Learning Representations, {ICLR}}.} {{2020 International Conference on Learning Representations, {ICLR}}.}
\PrintBackRefs{\CurrentBib}

\bibitem [\protect \citeauthoryear {%
Y.~Zhang%
}{%
Y.~Zhang%
}{%
{\protect \APACyear {1998}}%
}]{%
ht_1998space_filling}
\APACinsertmetastar {%
ht_1998space_filling}%
\begin{APACrefauthors}%
Zhang, Y.%
\end{APACrefauthors}%
\unskip\
\newblock
\APACrefYearMonthDay{1998}{}{}.
\newblock
{\BBOQ}\APACrefatitle {Space-filling curve ordered dither} {Space-filling curve ordered dither}.{\BBCQ}
\newblock
\APACjournalVolNumPages{Computers \& Graphics}{22}{4}{559--563}.
\newblock

\newblock

\PrintBackRefs{\CurrentBib}

\bibitem [\protect \citeauthoryear {%
Z.~Zhang%
\ \protect \BOthers {.}}{%
Z.~Zhang%
\ \protect \BOthers {.}}{%
{\protect \APACyear {2012}}%
}]{%
DB-CASIAFASD}
\APACinsertmetastar {%
DB-CASIAFASD}%
\begin{APACrefauthors}%
Zhang, Z.%
, Yan, J.%
, Liu, S.%
, Lei, Z.%
, Yi, D.%
\BCBL {} Li, S.Z.%
\end{APACrefauthors}%
\unskip\
\newblock
\APACrefYearMonthDay{2012}{}{}.
\newblock
{\BBOQ}\APACrefatitle {A face anti-spoofing database with diverse attacks} {A face anti-spoofing database with diverse attacks}.{\BBCQ}
\newblock
 \APACrefbtitle {{IAPR International Conference on Biometrics}} {{IAPR International Conference on Biometrics}}\ (\BPG~26-31).
\PrintBackRefs{\CurrentBib}

\bibitem [\protect \citeauthoryear {%
Zhou%
\ \protect \BOthers {.}}{%
Zhou%
\ \protect \BOthers {.}}{%
{\protect \APACyear {2022}}%
}]{%
zhou2022adaptive}
\APACinsertmetastar {%
zhou2022adaptive}%
\begin{APACrefauthors}%
Zhou, Q.%
, Zhang, K\BHBI Y.%
, Yao, T.%
, Yi, R.%
, Ding, S.%
\BCBL {} Ma, L.%
\end{APACrefauthors}%
\unskip\
\newblock
\APACrefYearMonthDay{2022}{}{}.
\newblock
{\BBOQ}\APACrefatitle {Adaptive Mixture of Experts Learning for Generalizable Face Anti-Spoofing} {Adaptive mixture of experts learning for generalizable face anti-spoofing}.{\BBCQ}
\newblock
 \APACrefbtitle {{Proceedings of the 30th ACM International Conference on Multimedia}} {{Proceedings of the 30th ACM International Conference on Multimedia}}\ (\BPGS\ 6009--6018).
\PrintBackRefs{\CurrentBib}

\end{thebibliography}
}

\end{document}


\title{FAS-Aug: Bag of Augmentations for Generalized Face Anti-Spoofing \\(Appendix)}
\maketitle
\section*{A.FAS Augmentation}
\subsection*{ICC Color Profile}
An ICC color profile incorporates the data describing the color attribute of an input or output device. We can perform the ICC transformation to change the color gamut through the mapping between the ICC color profiles. 
To develop our augmentations, we use the function, \textit{profiletoprofile()}, from the PIL.ImageCms module, which provides the ICC transformation with the aid of an input and an output profile.

For the augmentation of color diversity simulation, we uniformly sample an input and an output profile from 11 RGB ICC color profiles and pass them to the function, as the code illustrated in Figure \ref{fig:code_rgb}.  
Figure \ref{fig:rgb} shows examples of using the 11 different RGB ICC color profiles as the input profile, given a constant output profile, `sRGB.icc', and an input image.
Similarly, for the augmentation of color distortion simulation, we uniformly sample a CMYK ICC color profile as the input profile and an RGB ICC color profile as the output profile, and pass them to the function, as the code illustrated in Figure \ref{fig:code_cmyk}.
Figure \ref{fig:cmyk} shows the examples of using the 7 different CMYK ICC color profiles as the input profile, given a constant output profile, `sRGB.icc', and an input image.
\begin{figure*}
     \centering
     \includegraphics[width=\textwidth]{Appendix/rgb_profile.pdf}
     \caption{RGB Color Profile}
     \label{fig:rgb}
\end{figure*}
\begin{figure*}
     \centering
     \includegraphics[width=\textwidth]{Appendix/cmyk_profile.pdf}
     \caption{CMYK Color Profile}
     \label{fig:cmyk}
\end{figure*}
\begin{figure*}
     \centering
     \includegraphics[width=\textwidth]{Appendix/h_dot_conf.pdf}
     \caption{SFC-Halftone dot configuration}
     \label{fig:halftone}
\end{figure*}
\begin{figure*}
     \centering
     \includegraphics[width=\textwidth]{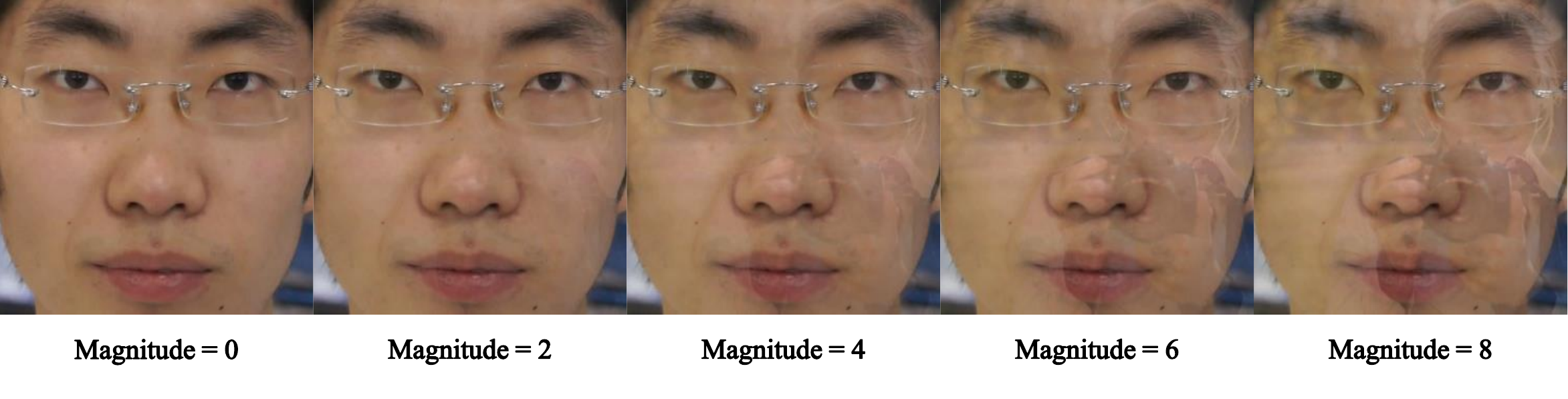}
     \caption{Reflection magnitude}
     \label{fig:reflection_mag}
\end{figure*}
\begin{figure*}
     \centering
     \includegraphics[width=\textwidth]{Appendix/r_magnitude.pdf}
     \caption{Reflection magnitude}
     \label{fig:reflection_exp}
\end{figure*}

\begin{figure}
     \centering
     \small
     \hline
     \includegraphics[width=0.45\textwidth]{Appendix/code_rgb.png}
     \hline
     \caption{code RGB}
     \label{fig:code_rgb}
\end{figure}
\begin{figure}
     \centering
     \small
     \hline
     \includegraphics[width=0.45\textwidth]{Appendix/code_cmyk.png}
     \hline
     \caption{code CMYK}
     \label{fig:code_cmyk}
\end{figure}

\subsection{Reflection Background Image}
We collect 90 background images from the internet, including pictures of indoor and outdoor scenes and pictures of people carrying a capture device. We use those picture in the augmentation of reflection artifacts simulation. Some examples of applying the augmentation are shown in Figure \ref{fig:reflection_exp}, and the examples of applying the augmentation with different magnitude $\gamma$ are shown in Figure \ref{fig:reflection_mag}.

\subsection{Subpixel}

\subsection{Space-filling curve algorithm}
Inspired by the space-filling curve ordered dither method \cite{TODO}, we design the augmentation of SFC-Halftone artifacts by quantizing a gray-scale image into 10 levels, which follows a simple algorithm suggested in \cite{TODO}. 
To generate the SFC-Halftone artifacts, we use the algorithm to generate a 3$\times$3 dot cluster for each pixel based on the gray value of the pixel. The dot configuration of different levels is shown in Figure \ref{fig:halftone}.

\subsection{Blue Noise texture}

\subsection{Augmentation policy for each experiment}